\documentclass{article}

\usepackage[final]{neurips_2020}

\usepackage{algorithm}
\usepackage{algorithmic}
\usepackage{graphicx}
\usepackage{subcaption}
\usepackage[utf8]{inputenc}
\usepackage[T1]{fontenc}
\usepackage{hyperref}
\usepackage{url}
\usepackage{booktabs}
\usepackage{amsfonts}
\usepackage{nicefrac}
\usepackage{microtype}
\usepackage{xcolor}
\usepackage{xspace}
\usepackage{amsmath,amssymb,mathtools,amsthm}
\usepackage{multirow}
\usepackage{pifont}
\usepackage{xparse}
\usepackage{pythonhighlight}
\usepackage{array}
\newcolumntype{H}{>{\setbox0=\hbox\bgroup}c<{\egroup}@{}}
\newcommand{\cmark}{\ding{51}}%
\newcommand{\xmark}{\ding{55}}

\DeclareCaptionLabelFormat{subfigure}{Figure #1~\thefigure#2:}
\DeclarePairedDelimiterX{\infdivx}[2]{(}{)}{%
  #1\;\delimsize|\delimsize|\;#2%
}
\newcommand{\kld}[2]{\ensuremath{D_{KL}\infdivx{#1}{#2}}\xspace}
\newcommand{\shortcite}[1]{[\citenum{#1}]}
\newcommand{\namecite}[1]{\citeauthor{#1} [\citenum{#1}]}
\newcommand{\remark}[1]{}
\newcommand{\ablaname}{ADM}
\newcommand{\guidedname}{ADM-G}
\setcitestyle{citesep={,}}

\NewDocumentCommand{\grad}{e{_^}}{%
  \mathop{}\!
  \nabla
  \IfValueT{#1}{_{\!#1}}
  \IfValueT{#2}{^{#2}}
}

\title{Diffusion Models Beat GANs on Image Synthesis}

\author{
  Prafulla Dhariwal\thanks{Equal contribution} \\
  OpenAI \\
  \texttt{prafulla@openai.com} \\
  \And
  Alex Nichol\footnotemark[1]\\
  OpenAI \\
  \texttt{alex@openai.com}
}

\begin{document}

\maketitle

\begin{abstract}
We show that diffusion models can achieve image sample quality superior to the current state-of-the-art generative models. We achieve this on unconditional image synthesis by finding a better architecture through a series of ablations. For conditional image synthesis, we further improve sample quality with classifier guidance: a simple, compute-efficient method for trading off diversity for fidelity using gradients from a classifier. We achieve an FID of 2.97 on ImageNet 128$\times$128, 4.59 on ImageNet 256$\times$256, and 7.72 on ImageNet 512$\times$512, and we match BigGAN-deep even with as few as 25 forward passes per sample, all while maintaining better coverage of the distribution. Finally, we find that classifier guidance combines well with upsampling diffusion models, further improving FID to 3.94 on ImageNet 256$\times$256 and 3.85 on ImageNet 512$\times$512. We release our code at \url{https://github.com/openai/guided-diffusion}. 
\end{abstract}

\section{Introduction}
\begin{figure}[h!]
    \centerline{\includegraphics[width=\textwidth]{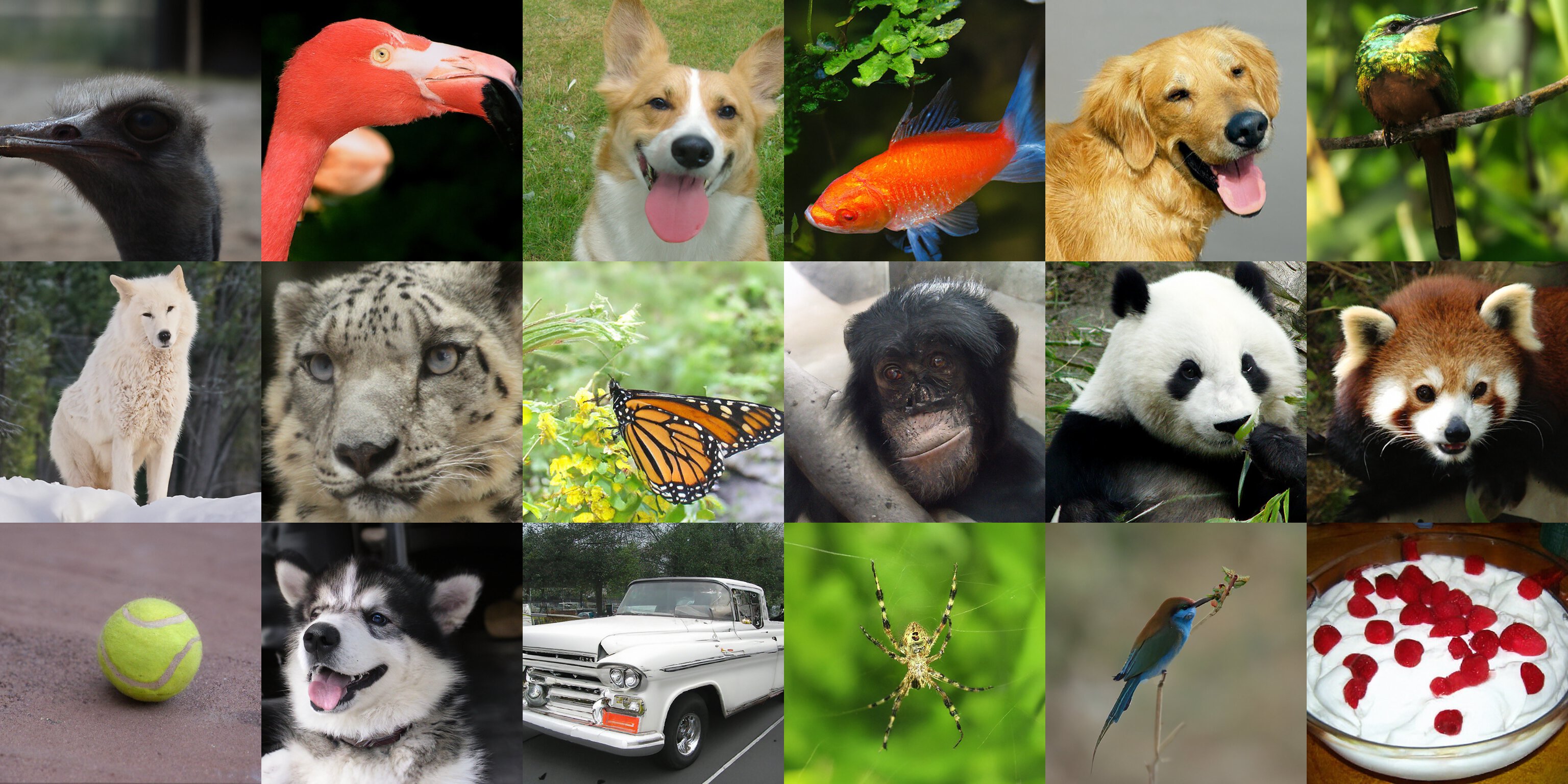}}
    \caption{Selected samples from our best ImageNet 512$\times$512 model (FID 3.85)}
    \label{fig:img512-cg-ts}
\end{figure}

Over the past few years, generative models have gained the ability to generate human-like natural language \shortcite{gpt3}, infinite high-quality synthetic images \shortcite{biggan,stylegan2,vqvae2} and highly diverse human speech and music \shortcite{wavenet,jukebox}. These models can be used in a variety of ways, such as generating images from text prompts \shortcite{stackgan,dalle} or learning useful feature representations \shortcite{bigbigan,igpt}. While these models are already capable of producing realistic images and sound, there is still much room for improvement beyond the current state-of-the-art, and better generative models could have wide-ranging impacts on graphic design, games, music production, and countless other fields. 

GANs \shortcite{gan} currently hold the state-of-the-art on most image generation tasks \shortcite{biggan,logan,stylegan2} as measured by sample quality metrics such as FID \shortcite{fid}, Inception Score \shortcite{inceptionscore} and Precision \shortcite{precrecall}. However, some of these metrics do not fully capture diversity, and it has been shown that GANs capture less diversity than state-of-the-art likelihood-based models \shortcite{vqvae2,improved,dctransformer}. Furthermore, GANs are often difficult to train, collapsing without carefully selected hyperparameters and regularizers \shortcite{biggan,sngan,orthoreg}.

While GANs hold the state-of-the-art, their drawbacks make them difficult to scale and apply to new domains. As a result, much work has been done to achieve GAN-like sample quality with likelihood-based models \shortcite{vqvae2,ddpm,dctransformer,vdvae}. While these models capture more diversity and are typically easier to scale and train than GANs, they still fall short in terms of visual sample quality. Furthermore, except for VAEs, sampling from these models is slower than GANs in terms of wall-clock time.

Diffusion models are a class of likelihood-based models which have recently been shown to produce high-quality images \shortcite{dickstein,scorematching,ddpm} while offering desirable properties such as distribution coverage, a stationary training objective, and easy scalability. These models generate samples by gradually removing noise from a signal, and their training objective can be expressed as a reweighted variational lower-bound \shortcite{ddpm}. This class of models already holds the state-of-the-art \shortcite{sde} on CIFAR-10 \shortcite{cifar10}, but still lags behind GANs on difficult generation datasets like LSUN and ImageNet. \namecite{improved} found that these models improve reliably with increased compute, and can produce high-quality samples even on the difficult ImageNet 256$\times$256 dataset using an upsampling stack. However, the FID of this model is still not competitive with BigGAN-deep \shortcite{biggan}, the current state-of-the-art on this dataset.

We hypothesize that the gap between diffusion models and GANs stems from at least two factors: first, that the model architectures used by recent GAN literature have been heavily explored and refined; second, that GANs are able to trade off diversity for fidelity, producing high quality samples but not covering the whole distribution. We aim to bring these benefits to diffusion models, first by improving model architecture and then by devising a scheme for trading off diversity for fidelity. With these improvements, we achieve a new state-of-the-art, surpassing GANs on several different metrics and datasets.

The rest of the paper is organized as follows. In Section \ref{sec:background}, we give a brief background of diffusion models based on \namecite{ddpm} and the improvements from \namecite{improved} and \namecite{ddim}, and we describe our evaluation setup. In Section \ref{sec:archablation}, we introduce simple architecture improvements that give a substantial boost to FID. In Section \ref{sec:guidance}, we describe a method for using gradients from a classifier to guide a diffusion model during sampling. We find that a single hyperparameter, the scale of the classifier gradients, can be tuned to trade off diversity for fidelity, and we can increase this gradient scale factor by an order of magnitude without obtaining adversarial examples \shortcite{adversarialexamples}.
Finally, in Section \ref{sec:results} we show that models with our improved architecture achieve state-of-the-art on unconditional image synthesis tasks, and with classifier guidance achieve state-of-the-art on conditional image synthesis. When using classifier guidance, we find that we can sample with as few as 25 forward passes while maintaining FIDs comparable to BigGAN. We also compare our improved models to upsampling stacks, finding that the two approaches give complementary improvements and that combining them gives the best results on ImageNet 256$\times$256 and 512$\times$512.

\section{Background}
\label{sec:background}
In this section, we provide a brief overview of diffusion models. For a more detailed mathematical description, we refer the reader to Appendix \ref{app:detailedbackground}.

On a high level, diffusion models sample from a distribution by reversing a gradual noising process. In particular, sampling starts with noise $x_T$ and produces gradually less-noisy samples $x_{T-1},x_{T-2},...$ until reaching a final sample $x_0$. Each timestep $t$ corresponds to a certain noise level, and $x_t$ can be thought of as a mixture of a signal $x_0$ with some noise $\epsilon$ where the signal to noise ratio is determined by the timestep $t$. For the remainder of this paper, we assume that the noise $\epsilon$ is drawn from a diagonal Gaussian distribution, which works well for natural images and simplifies various derivations.

A diffusion model learns to produce a slightly more ``denoised'' $x_{t-1}$ from $x_{t}$. \namecite{ddpm} parameterize this model as a function $\epsilon_{\theta}(x_t,t)$ which predicts the noise component of a noisy sample $x_t$. To train these models, each sample in a minibatch is produced by randomly drawing a data sample $x_0$, a timestep $t$, and noise $\epsilon$, which together give rise to a noised sample $x_t$ (Equation \ref{eq:jumpnoise}). 
The training objective is then $||\epsilon_{\theta}(x_t,t) - \epsilon||^2$, i.e. a simple mean-squared error loss between the true noise and the predicted noise (Equation \ref{eq:lsimple}). 

It is not immediately obvious how to sample from a noise predictor $\epsilon_{\theta}(x_t,t)$. Recall that diffusion sampling proceeds by repeatedly predicting $x_{t-1}$ from $x_t$, starting from $x_T$. \namecite{ddpm} show that, under reasonable assumptions, we can model the distribution $p_{\theta}(x_{t-1}|x_t)$ of $x_{t-1}$ given $x_t$ as a diagonal Gaussian $\mathcal{N}(x_{t-1};\mu_{\theta}(x_t, t), \Sigma_{\theta}(x_t, t))$,
where the mean $\mu_{\theta}(x_t, t)$ can be calculated as a function of $\epsilon_{\theta}(x_t,t)$ (Equation \ref{eq:mufromeps}). 
The variance $\Sigma_{\theta}(x_t,t)$ of this Gaussian distribution can be fixed to a known constant \shortcite{ddpm} or learned with a separate neural network head \shortcite{improved}, 
and both approaches yield high-quality samples when the total number of diffusion steps $T$ is large enough.

\namecite{ddpm} observe that the simple mean-sqaured error objective,  $L_{\text{simple}}$, works better in practice than the actual variational lower bound $L_{\text{vlb}}$ that can be derived from interpreting the denoising diffusion model as a VAE. They also note that training with this objective and using their corresponding sampling procedure is equivalent to the denoising score matching model from \namecite{improvedscore}, who use Langevin dynamics to sample from a denoising model trained with multiple noise levels to produce high quality image samples. We often use ``diffusion models'' as shorthand to refer to both classes of models.

\subsection{Improvements}
Following the breakthrough work of \namecite{improvedscore} and \namecite{ddpm}, several recent papers have proposed improvements to diffusion models. Here we describe a few of these improvements, which we employ for our models.

\namecite{improved} find that fixing the variance $\Sigma_{\theta}(x_t,t)$ to a constant as done in \namecite{ddpm} is sub-optimal for sampling with fewer diffusion steps, and propose to parameterize $\Sigma_{\theta}(x_t,t)$ as a neural network whose output $v$ is interpolated as:
\begin{alignat}{2}
    \Sigma_{\theta}(x_t,t) &= \exp(v \log \beta_t + (1-v) \log \tilde{\beta}_t) \label{eq:learnedsigma}
\end{alignat}

Here, $\beta_t$ and $\tilde{\beta}_t$ (Equation \ref{eq:betatilde}) are the variances in \namecite{ddpm} corresponding to upper and lower bounds for the reverse process variances. Additionally, \namecite{improved} propose a hybrid objective for training both $\epsilon_{\theta}(x_t,t)$ and $\Sigma_{\theta}(x_t,t)$ using the weighted sum $L_{\text{simple}} + \lambda L_{\text{vlb}}$. Learning the reverse process variances with their hybrid objective allows sampling with fewer steps without much drop in sample quality. We adopt this objective and parameterization, and use it throughout our experiments.

\namecite{ddim} propose DDIM, which formulates an alternative non-Markovian noising process that has the same forward marginals as DDPM, but allows producing different reverse samplers by changing the variance of the reverse noise. By setting this noise to 0, they provide a way to turn any model $\epsilon_{\theta}(x_t,t)$ into a deterministic mapping from latents to images, and find that this provides an alternative way to sample with fewer steps. We adopt this sampling approach when using fewer than 50 sampling steps, since \namecite{improved} found it to be beneficial in this regime.

\subsection{Sample Quality Metrics}

For comparing sample quality across models, we perform quantitative evaluations using the following metrics. While these metrics are often used in practice and correspond well with human judgement, they are not a perfect proxy, and finding better metrics for sample quality evaluation is still an open problem.

Inception Score (IS) was proposed by \namecite{inceptionscore}, and it measures how well a model captures the full ImageNet class distribution while still producing individual samples that are convincing examples of a single class. One drawback of this metric is that it does not reward covering the whole distribution or capturing diversity within a class, and models which memorize a small subset of the full dataset will still have high IS \shortcite{noteoninceptionscore}. To better capture diversity than IS, Fréchet Inception Distance (FID) was proposed by \namecite{fid}, who argued that it is more consistent with human judgement than Inception Score. FID provides a symmetric measure of the distance between two image distributions in the Inception-V3 \shortcite{inceptionv3} latent space. Recently, sFID was proposed by \namecite{dctransformer} as a version of FID that uses spatial features rather than the standard pooled features. They find that this metric better captures spatial relationships, rewarding image distributions with coherent high-level structure. Finally, \namecite{precrecall} proposed Improved Precision and Recall metrics to separately measure sample fidelity as the fraction of model samples which fall into the data manifold (precision), and diversity as the fraction of data samples which fall into the sample manifold (recall).





We use FID as our default metric for overall sample quality comparisons as it captures both diversity and fidelity and has been the de facto standard metric for state-of-the-art generative modeling work \shortcite{stylegan,stylegan2,biggan,ddpm}. We use Precision or IS to measure fidelity, and Recall to measure diversity or distribution coverage. When comparing against other methods, we re-compute these metrics using public samples or models whenever possible. This is for two reasons: first, some papers \shortcite{stylegan,stylegan2,ddpm} compare against arbitrary subsets of the training set which are not readily available; and second, subtle implementation differences can affect the resulting FID values \shortcite{cleanfid}. To ensure consistent comparisons, we use the entire training set as the reference batch \shortcite{fid,biggan}, and evaluate metrics for all models using the same codebase.

\section{Architecture Improvements}
\label{sec:archablation}
In this section we conduct several architecture ablations to find the model architecture that provides the best sample quality for diffusion models.

\namecite{ddpm} introduced the UNet architecture for diffusion models, which \namecite{adversarial} found to substantially improve sample quality over the previous architectures \shortcite{improvedscore,refinenet} used for denoising score matching. The UNet model uses a stack of residual layers and downsampling convolutions, followed by a stack of residual layers with upsampling colvolutions, with skip connections connecting the layers with the same spatial size. In addition, they use a global attention layer at the 16$\times$16 resolution with a single head, and add a projection of the timestep embedding into each residual block. \namecite{sde} found that further changes to the UNet architecture improved performance on the CIFAR-10 \shortcite{cifar10} and CelebA-64 \shortcite{celeba} datasets. We show the same result on ImageNet 128$\times$128, finding that architecture can indeed give a substantial boost to sample quality on much larger and more diverse datasets at a higher resolution.

We explore the following architectural changes:

\begin{itemize}
  \item Increasing depth versus width, holding model size relatively constant.
  \item Increasing the number of attention heads.
  \item Using attention at 32$\times$32, 16$\times$16, and 8$\times$8 resolutions rather than only at 16$\times$16.
  \item Using the BigGAN \shortcite{biggan} residual block for upsampling and downsampling the activations, following \shortcite{sde}.
  \item Rescaling residual connections with $\frac{1}{\sqrt{2}}$, following \shortcite{sde,stylegan,stylegan2}.
\end{itemize}

\begin{table}[t]
    \begin{center}
    \begin{small}
    \begin{tabular}{cccccccc}
    \toprule
    \multirow{2}*{Channels} & \multirow{2}*{Depth} & \multirow{2}*{Heads} & Attention  & BigGAN & Rescale & FID & FID \\
    & & & resolutions & up/downsample & resblock & 700K & 1200K \\
    \midrule
    160 & 2 & 1 & 16      & \xmark & \xmark &  15.33 & 13.21  \\
    \midrule
    \midrule
    128 & 4 &   &         &   &   &  -0.21 & -0.48  \\
        &   & 4 &         &   &   &  -0.54 & -0.82  \\
        &   &   & 32,16,8 &   &   &  -0.72 & -0.66  \\
        &   &   &         & \cmark &   &  -1.20 & -1.21  \\
        &   &   &         &   & \cmark &   0.16 & 0.25  \\
    160 & 2 & 4 & 32,16,8 & \cmark & \xmark &  \bf{-3.14} & \bf{-3.00} \\
    \bottomrule
    \end{tabular}
    \end{small}
    \end{center}
    \caption{Ablation of various architecture changes, evaluated at 700K and 1200K iterations}
    \label{tab:arch}
    \vskip -0.2 in
\end{table}

\begin{table}[t!]
    \begin{center}
    \begin{small}
    \begin{tabular}{ccc}
    \toprule
    Number of heads & Channels per head & FID \\
    \midrule
    1 & & 14.08 \\
    \midrule
    \midrule
    2 & & -0.50   \\
    4 & & -0.97  \\
    8 & & -1.17   \\
      & 32 & -1.36 \\
      & 64 & -1.03 \\
      & 128 & -1.08 \\
    \bottomrule
    \end{tabular}
    \end{small}
    \end{center}
    \caption{Ablation of various attention configurations. More heads or lower channels per heads both lead to improved FID.}
    \label{tab:heads}
    \vskip -0.2in
\end{table}

\begin{figure}[t]
    \begin{center}
    \begin{subfigure}{0.45\textwidth}
        \centering
        \includegraphics[width=\textwidth]{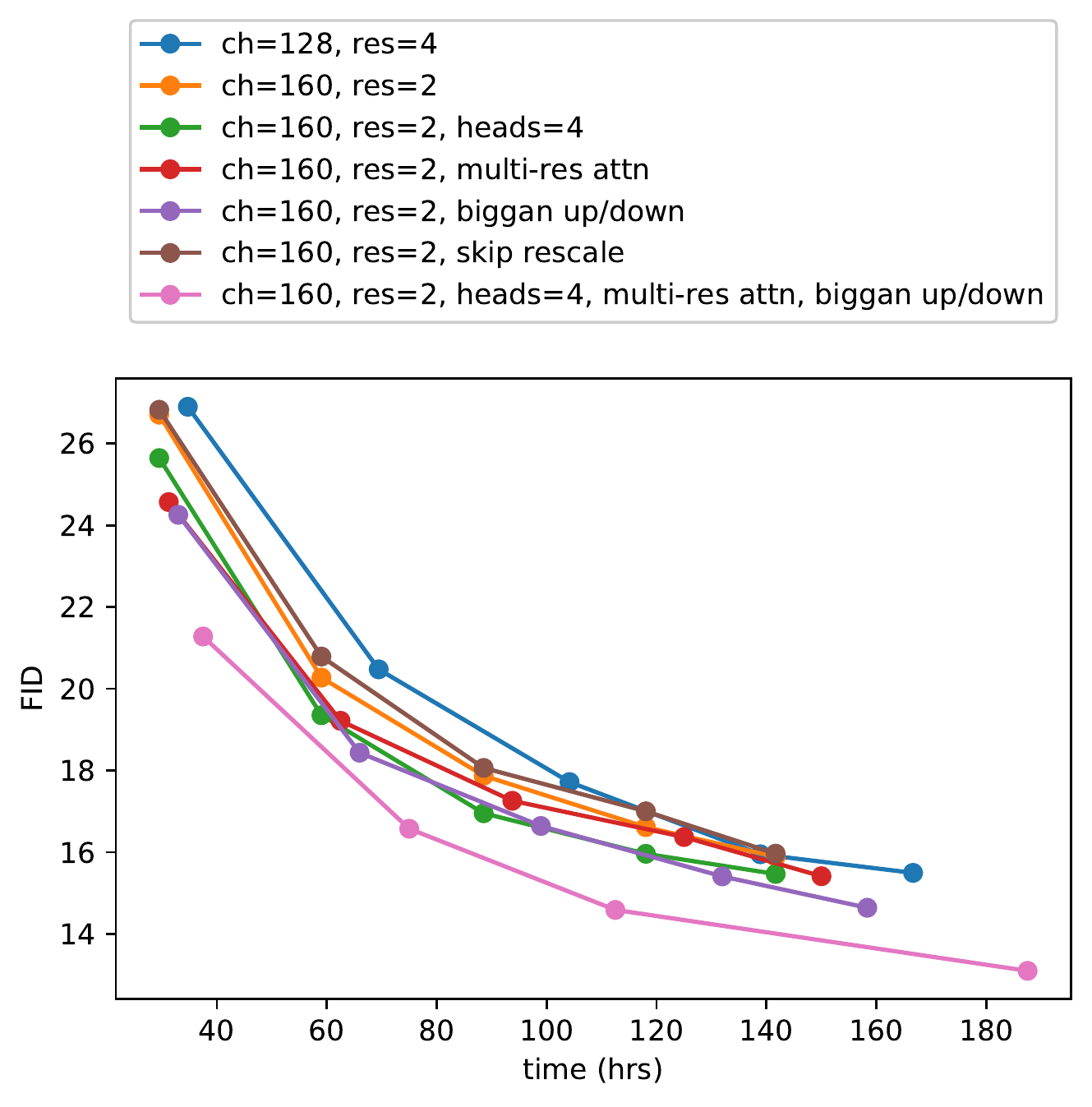}
    \end{subfigure}
    \hspace{0.025\textwidth}
    \begin{subfigure}{0.45\textwidth}
        \centering
        \includegraphics[width=\textwidth]{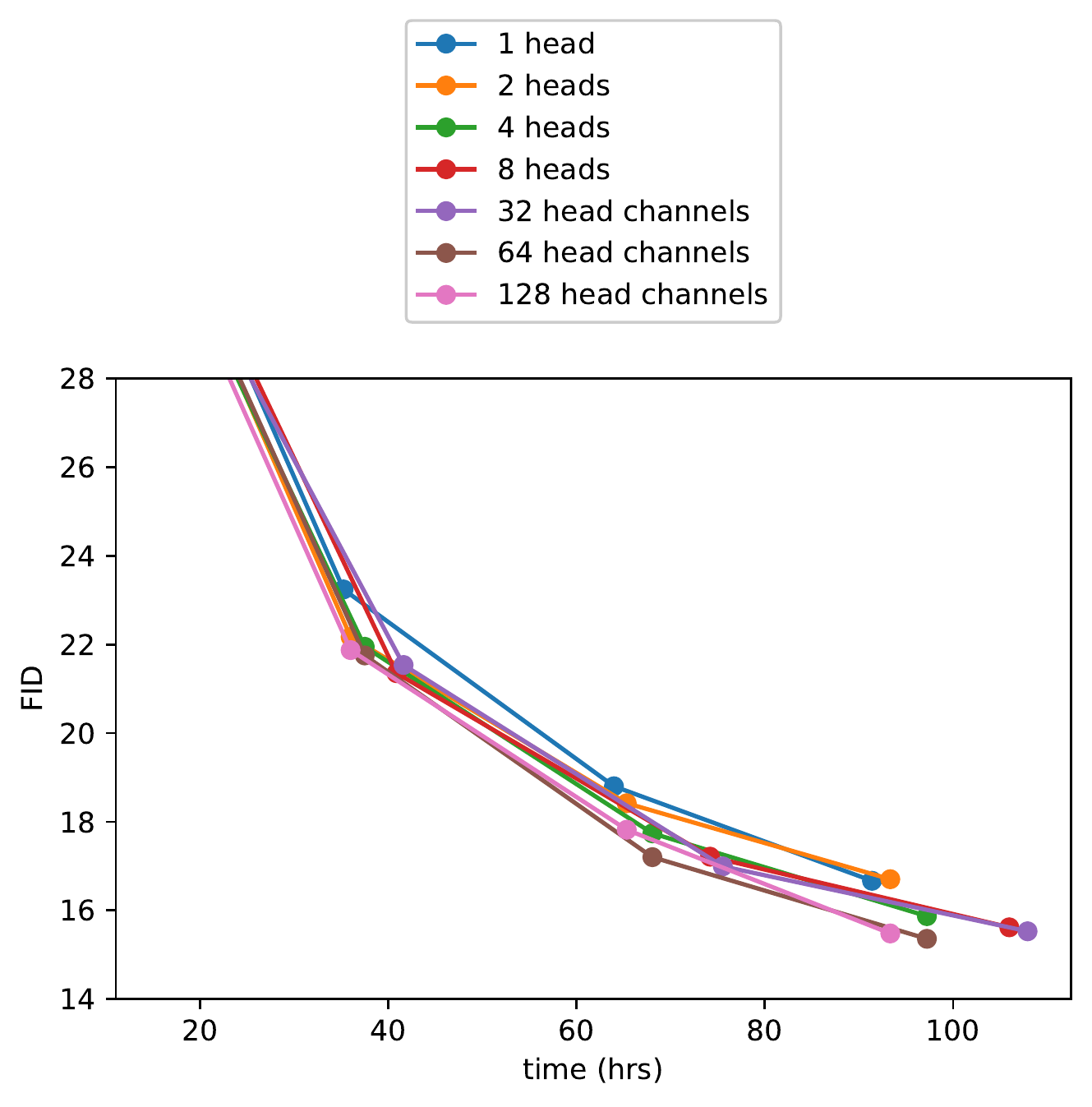}
    \end{subfigure}
    \end{center}
    \caption{\label{fig:arch} Ablation of various architecture changes, showing FID as a function of wall-clock time. FID evaluated over 10k samples instead of 50k for efficiency.}
\end{figure}

For all comparisons in this section, we train models on ImageNet 128$\times$128 with batch size 256, and sample using 250 sampling steps. We train models with the above architecture changes and compare them on FID, evaluated at two different points of training, in Table \ref{tab:arch}. Aside from rescaling residual connections, all of the other modifications improve performance and have a positive compounding effect. We observe in Figure \ref{fig:arch} that while increased depth helps performance, it increases training time and takes longer to reach the same performance as a wider model, so we opt not to use this change in further experiments. 

We also study other attention configurations that better match the Transformer architecture \shortcite{transformer}. To this end, we experimented with either fixing attention heads to a constant, or fixing the number of channels per head. For the rest of the architecture, we use 128 base channels, 2 residual blocks per resolution, multi-resolution attention, and BigGAN up/downsampling, and we train the models for 700K iterations. Table \ref{tab:heads} shows our results, indicating that more heads or fewer channels per head improves FID. In Figure \ref{fig:arch}, we see 64 channels is best for wall-clock time, so we opt to use 64 channels per head as our default. We note that this choice also better matches modern transformer architectures, and is on par with our other configurations in terms of final FID.

\subsection{Adaptive Group Normalization}
\label{sec:adapgn}
\begin{table}[t!]
    \begin{center}
    \begin{small}
    \begin{tabular}{ccc}
    \toprule
    Operation & FID \\
    \midrule
    AdaGN & 13.06 \\
    Addition + GroupNorm & 15.08 \\
    \bottomrule
    \end{tabular}
    \end{small}
    \end{center}
    \caption{Ablating the element-wise operation used when projecting timestep and class embeddings into each residual block. Replacing AdaGN with the Addition + GroupNorm layer from \namecite{ddpm} makes FID worse.}
    \label{tab:adagn}
    \vskip -0.2in
\end{table}

We also experiment with a layer \shortcite{improved} that we refer to as adaptive group normalization (AdaGN), which incorporates the timestep and class embedding into each residual block after a group normalization operation \shortcite{groupnorm}, similar to adaptive instance norm \shortcite{stylegan} and FiLM \shortcite{film}. We define this layer as $\text{AdaGN}(h, y) = y_s  \text{ GroupNorm}(h) + y_b$, where $h$ is the intermediate activations of the residual block following the first convolution, and $y = [y_s, y_b]$ is obtained from a linear projection of the timestep and class embedding. 

We had already seen AdaGN improve our earliest diffusion models, and so had it included by default in all our runs. In Table \ref{tab:adagn}, we explicitly ablate this choice, and find that the adaptive group normalization layer indeed improved FID. Both models use 128 base channels and 2 residual blocks per resolution, multi-resolution attention with 64 channels per head, and BigGAN up/downsampling, and were trained for 700K iterations.

In the rest of the paper, we use this final improved model architecture as our default: variable width with 2 residual blocks per resolution, multiple heads with 64 channels per head, attention at 32, 16 and 8 resolutions, BigGAN residual blocks for up and downsampling, and adaptive group normalization for injecting timestep and class embeddings into residual blocks.

\section{Classifier Guidance}
\label{sec:guidance}

In addition to employing well designed architectures, GANs for conditional image synthesis \shortcite{cgan,biggan} make heavy use of class labels. This often takes the form of class-conditional normalization statistics \shortcite{cganbn,cganmodulat} as well as discriminators with heads that are explicitly designed to behave like classifiers $p(y|x)$ \shortcite{cganproj}. As further evidence that class information is crucial to the success of these models, \namecite{unlabeledgan} find that it is helpful to generate synthetic labels when working in a label-limited regime.

Given this observation for GANs, it makes sense to explore different ways to condition diffusion models on class labels. We already incorporate class information into normalization layers (Section \ref{sec:adapgn}). Here, we explore a different approach: exploiting a classifier $p(y|x)$ to improve a diffusion generator. \namecite{dickstein} and \namecite{sde} show one way to achieve this, wherein a pre-trained diffusion model can be conditioned using the gradients of a classifier. In particular, we can train a classifier $p_{\phi}(y|x_t,t)$ on noisy images $x_t$, and then use gradients $\grad_{x_t} \log p_{\phi}(y|x_t,t)$ to guide the diffusion sampling process towards an arbitrary class label $y$.

In this section, we first review two ways of deriving conditional sampling processes using classifiers. We then describe how we use such classifiers in practice to improve sample quality. We choose the notation $p_{\phi}(y|x_t,t) = p_{\phi}(y|x_t)$ and $\epsilon_\theta(x_t,t) = \epsilon_\theta(x_t)$ for brevity, noting that they refer to separate functions for each timestep $t$ and at training time the models must be conditioned on the input $t$. 

\subsection{Conditional Reverse Noising Process}

We start with a diffusion model with an unconditional reverse noising process $p_{\theta}(x_t|x_{t+1})$. To condition this on a label $y$, it suffices to sample each transition\footnote{We must also sample $x_T$ conditioned on $y$, but a noisy enough diffusion process causes $x_T$ to be nearly Gaussian even in the conditional case.} according to
\begin{alignat}{1}
p_{\theta,\phi}(x_t|x_{t+1},y) = Z p_{\theta}(x_t|x_{t+1})p_{\phi}(y|x_t) \label{eq:normalizedtransition}
\end{alignat}

where $Z$ is a normalizing constant (proof in Appendix \ref{app:conditional}). It is typically intractable to sample from this distribution exactly, but \namecite{dickstein} show that it can be approximated as a perturbed Gaussian distribution. Here, we review this derivation.

Recall that our diffusion model predicts the previous timestep $x_t$ from timestep $x_{t+1}$ using a Gaussian distribution:
\begin{alignat}{2}
p_{\theta}(x_t|x_{t+1}) &= \mathcal{N}(\mu, \Sigma) \\
\log p_{\theta}(x_t|x_{t+1}) &= -\frac{1}{2} (x_t - \mu)^T \Sigma^{-1} (x_t - \mu) + C
\end{alignat}

We can assume that $\log_{\phi} p(y|x_t)$ has low curvature compared to $\Sigma^{-1}$. This assumption is reasonable in the limit of infinite diffusion steps, where $||\Sigma|| \to 0$. In this case, we can approximate $\log p_{\phi}(y|x_t)$ using a Taylor expansion around $x_t = \mu$ as 
\begin{alignat}{2}
\log p_{\phi}(y|x_t) &\approx \log p_{\phi}(y|x_t)|_{x_t = \mu} + (x_t - \mu)\grad_{x_t} \log p_{\phi}(y|x_t)|_{x_t = \mu} \\
&= (x_t - \mu)g + C_1 
\end{alignat}

Here, $g = \grad_{x_t} \log p_{\phi}(y|x_t)|_{x_t = \mu}$, and $C_1$ is a constant. This gives
\begin{alignat}{2}
    \log(p_{\theta}(x_t|x_{t+1}) p_{\phi}(y|x_t)) &\approx -\frac{1}{2} (x_t - \mu)^T \Sigma^{-1} (x_t - \mu) + (x_t - \mu)g + C_2 \\
    &= - \frac{1}{2} (x_t - \mu - \Sigma g)^T \Sigma^{-1} (x_t - \mu - \Sigma g) + \frac{1}{2} g^T \Sigma g + C_2 \\
    &= - \frac{1}{2} (x_t - \mu - \Sigma g)^T \Sigma^{-1} (x_t - \mu - \Sigma g) + C_3 \\
    &= \log p(z) + C_4, z \sim \mathcal{N}(\mu + \Sigma g, \Sigma)
    \label{eq:condsamp}
\end{alignat}

We can safely ignore the constant term $C_4$, since it corresponds to the normalizing coefficient $Z$ in Equation \ref{eq:normalizedtransition}. We have thus found that the conditional transition operator can be approximated by a Gaussian similar to the unconditional transition operator, but with its mean shifted by $\Sigma g$. Algorithm \ref{alg:guiding} summaries the corresponding sampling algorithm. We include an optional scale factor $s$ for the gradients, which we describe in more detail in Section \ref{sec:scalingcg}.  

\begin{algorithm}[t]
    \caption{Classifier guided diffusion sampling, given a diffusion model $(\mu_{\theta}(x_t), \Sigma_{\theta}(x_t))$, classifier $p_{\phi}(y|x_t)$, and gradient scale $s$.}
    \label{alg:guiding}
    \begin{algorithmic}
        \STATE Input: class label $y$, gradient scale $s$
        \STATE $x_T \gets \text{sample from } \mathcal{N}(0, \mathbf{I})$
        \FORALL{$t$ from $T$ to 1}
            \STATE $\mu, \Sigma \gets \mu_{\theta}(x_t), \Sigma_{\theta}(x_t)$
            \STATE $x_{t-1} \gets \text{sample from } \mathcal{N}(\mu + s \Sigma \grad_{x_t} \log p_{\phi}(y|x_t), \Sigma)$
        \ENDFOR
        \RETURN $x_0$
    \end{algorithmic}
\end{algorithm}

\begin{algorithm}[t]
    \caption{Classifier guided DDIM sampling, given a diffusion model $\epsilon_{\theta}(x_t)$, classifier $p_{\phi}(y|x_t)$, and gradient scale $s$.}
    \label{alg:guidingddim}
    \begin{algorithmic}
        \STATE Input: class label $y$, gradient scale $s$
        \STATE $x_T \gets \text{sample from } \mathcal{N}(0, \mathbf{I})$
        \FORALL{$t$ from $T$ to 1}
            \STATE $\hat \epsilon \gets \epsilon_{\theta}(x_t) - \sqrt{1-\bar{\alpha}_t} \grad_{x_t} \log p_{\phi}(y|x_t)$
            \STATE $x_{t-1} \gets \sqrt{\bar{\alpha}_{t-1}} \left( \frac{x_t - \sqrt{1-\bar{\alpha}_t} \hat{\epsilon}}{\sqrt{\bar{\alpha}_t}} \right) + \sqrt{1-\bar{\alpha}_{t-1}} \hat{\epsilon}$
        \ENDFOR
        \RETURN $x_0$
    \end{algorithmic}
\end{algorithm}

\subsection{Conditional Sampling for DDIM}
\label{sec:ddimguide}

The above derivation for conditional sampling is only valid for the stochastic diffusion sampling process, and cannot be applied to deterministic sampling methods like DDIM \shortcite{ddim}. To this end, we use a score-based conditioning trick adapted from \namecite{sde}, which leverages the connection between diffusion models and score matching \shortcite{scorematching}. In particular, if we have a model $\epsilon_{\theta}(x_t)$ that predicts the noise added to a sample, then this can be used to derive a score function:
\begin{alignat}{1}\grad_{x_t} \log p_{\theta}(x_t) = -\frac{1}{\sqrt{1-\bar{\alpha}_t}} \epsilon_{\theta}(x_t)
\end{alignat}

We can now substitute this into the score function for $p(x_t)p(y|x_t)$:
\begin{alignat}{2}
    \grad_{x_t} \log(p_{\theta}(x_t) p_{\phi}(y|x_t)) &= \grad_{x_t} \log p_{\theta}(x_t) + \grad_{x_t} \log p_{\phi}(y|x_t) \\
    &= -\frac{1}{\sqrt{1-\bar{\alpha}_t}} \epsilon_{\theta}(x_t) + \grad_{x_t} \log p_{\phi}(y|x_t)
\end{alignat}

Finally, we can define a new epsilon prediction $\hat{\epsilon}(x_t)$ which corresponds to the score of the joint distribution:
\begin{alignat}{1}\hat{\epsilon}(x_t) \coloneqq \epsilon_{\theta}(x_t) - \sqrt{1-\bar{\alpha}_t} \grad_{x_t} \log p_{\phi}(y|x_t)
\end{alignat}

We can then use the exact same sampling procedure as used for regular DDIM, but with the modified noise predictions $\hat{\epsilon}_{\theta}(x_t)$ instead of $\epsilon_{\theta}(x_t)$. Algorithm \ref{alg:guidingddim} summaries the corresponding sampling algorithm. 

\subsection{Scaling Classifier Gradients}
\label{sec:scalingcg}
\begin{figure}[t!]
    \begin{center}
    \begin{subfigure}{0.475\textwidth}
        \centering
        \includegraphics[width=\textwidth]{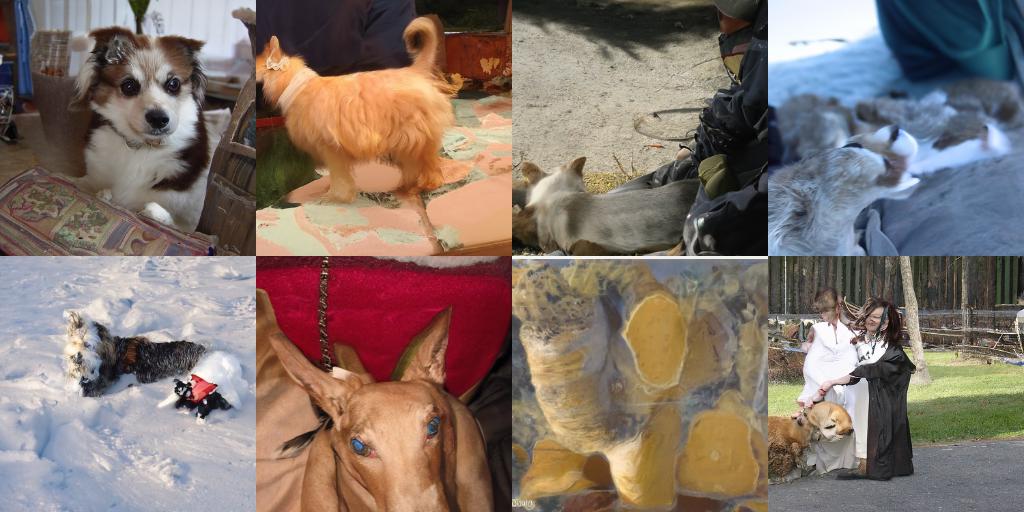}
    \end{subfigure}
    \hspace{0.025\textwidth}
    \begin{subfigure}{0.475\textwidth}
        \centering
        \includegraphics[width=\textwidth]{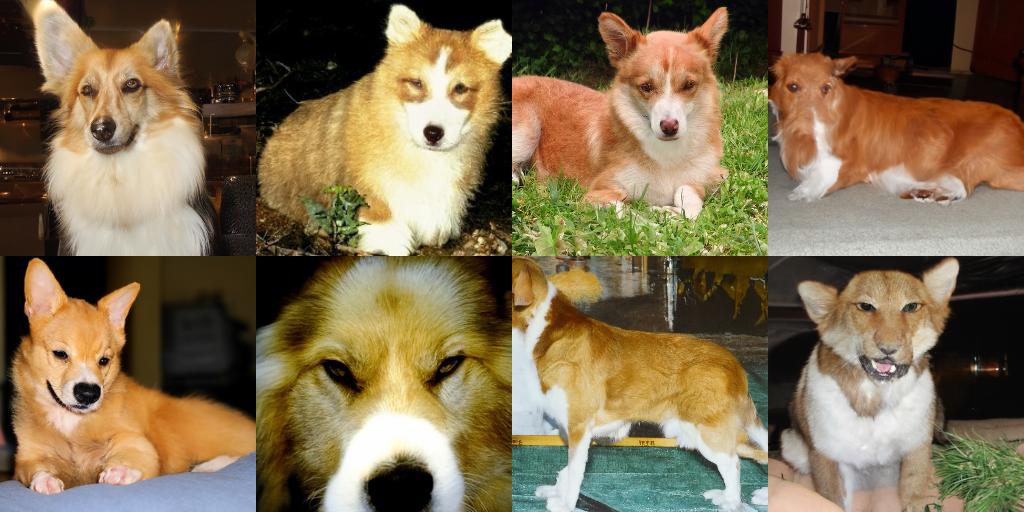}
    \end{subfigure}
    \caption{\label{fig:uncond_scales} Samples from an unconditional diffusion model with classifier guidance to condition on the class "Pembroke Welsh corgi". Using classifier scale 1.0 (left; FID: 33.0) does not produce convincing samples in this class, whereas classifier scale 10.0 (right; FID: 12.0) produces much more class-consistent images.}
    \end{center}
\end{figure}

To apply classifier guidance to a large scale generative task, we train classification models on ImageNet. Our classifier architecture is simply the downsampling trunk of the UNet model with an attention pool \shortcite{clip} at the 8x8 layer to produce the final output. We train these classifiers on the same noising distribution as the corresponding diffusion model, and also add random crops to reduce overfitting. After training, we incorporate the classifier into the sampling process of the diffusion model using Equation \ref{eq:condsamp}, as outlined by Algorithm \ref{alg:guiding}.

In initial experiments with unconditional ImageNet models, we found it necessary to scale the classifier gradients by a constant factor larger than 1. When using a scale of 1, we observed that the classifier assigned reasonable probabilities (around 50\%) to the desired classes for the final samples, but these samples did not match the intended classes upon visual inspection.
Scaling up the classifier gradients remedied this problem, and the class probabilities from the classifier increased to nearly 100\%. Figure \ref{fig:uncond_scales} shows an example of this effect.

To understand the effect of scaling classifier gradients, note that $s \cdot \grad_x \log p(y|x) = \grad_x \log \frac{1}{Z} p(y|x)^s$, where $Z$ is an arbitrary constant. As a result, the conditioning process is still theoretically grounded in a re-normalized classifier distribution proportional to $p(y|x)^s$. When $s > 1$, this distribution becomes sharper than $p(y|x)$, since larger values are amplified by the exponent. In other words, using a larger gradient scale focuses more on the modes of the classifier, which is potentially desirable for producing higher fidelity (but less diverse) samples.

\begin{table}[t]
    \begin{center}
    \begin{small}
    \begin{tabular}{cccccccc}
    \toprule
    Conditional & Guidance & Scale & FID   & sFID  & IS     & Precision & Recall  \\
    \midrule
    \xmark      & \xmark   &       & 26.21 & \bf 6.35  & 39.70  & 0.61      & 0.63 \\
    \xmark      & \cmark   & 1.0   & 33.03 & 6.99  & 32.92  & 0.56      & \bf 0.65 \\
    \xmark      & \cmark   & 10.0  & \bf 12.00 & 10.40 & \bf 95.41  & \bf 0.76      & 0.44 \\
    \hline
    \cmark      & \xmark   &       & 10.94 & 6.02  & 100.98 & 0.69      & \bf 0.63 \\ 
    \cmark      & \cmark   & 1.0   & \bf 4.59  & \bf 5.25  & 186.70 & 0.82      & 0.52 \\
    \cmark      & \cmark   & 10.0  & 9.11 &	10.93  & \bf 283.92  & \bf 0.88      & 0.32 \\
    \bottomrule
    \end{tabular}
    \end{small}
    \end{center}
    \caption{Effect of classifier guidance on sample quality. Both conditional and unconditional models were trained for 2M iterations on ImageNet 256$\times$256 with batch size 256.}
    \label{tab:guide}
    \vskip -0.2in
\end{table}

\begin{figure}[t]
    \begin{center}
    \begin{subfigure}{0.32\textwidth}
        \centerline{\includegraphics[width=\textwidth]{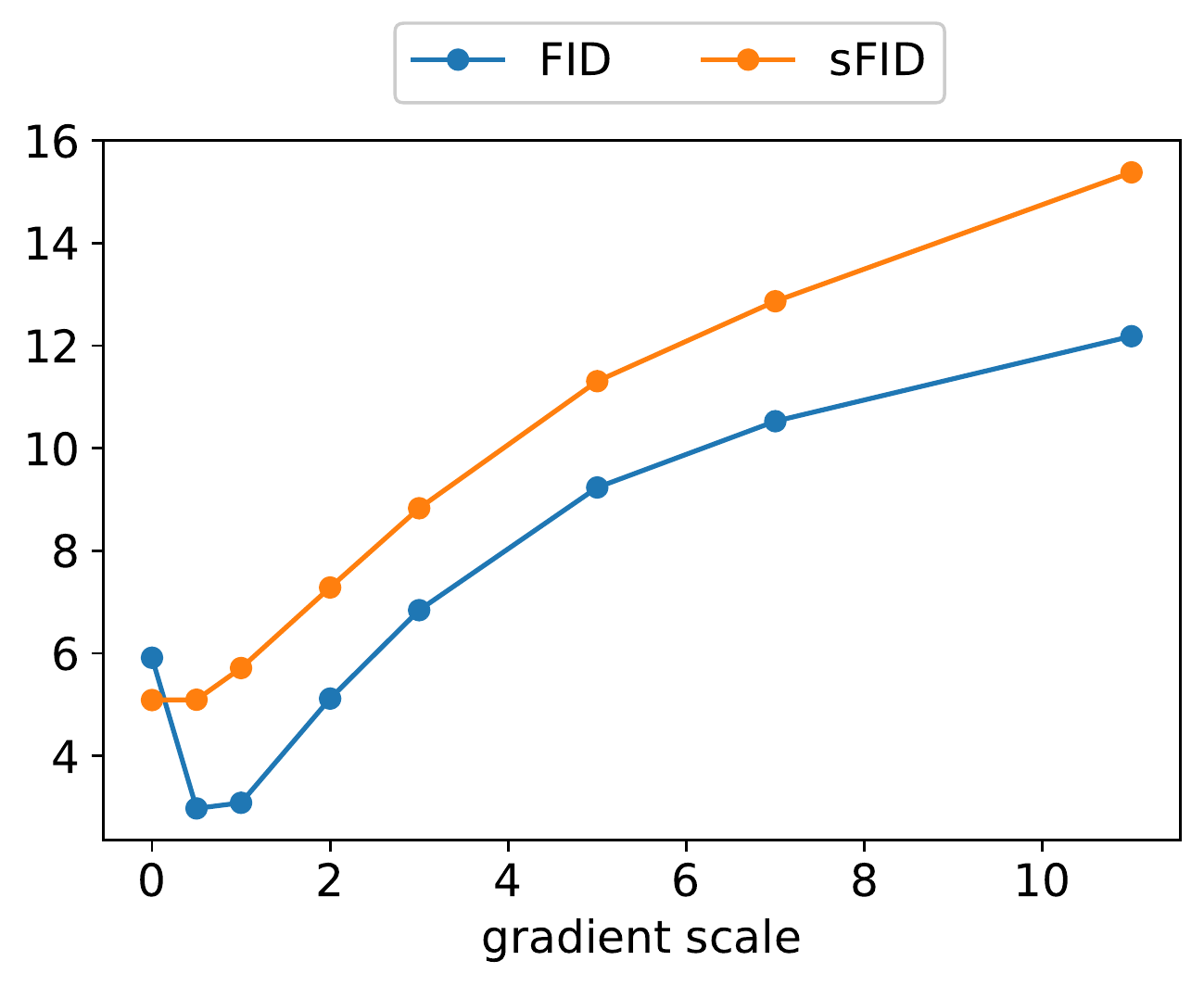}}
    \end{subfigure}
    \begin{subfigure}{0.32\textwidth}
        \centerline{\includegraphics[width=\textwidth]{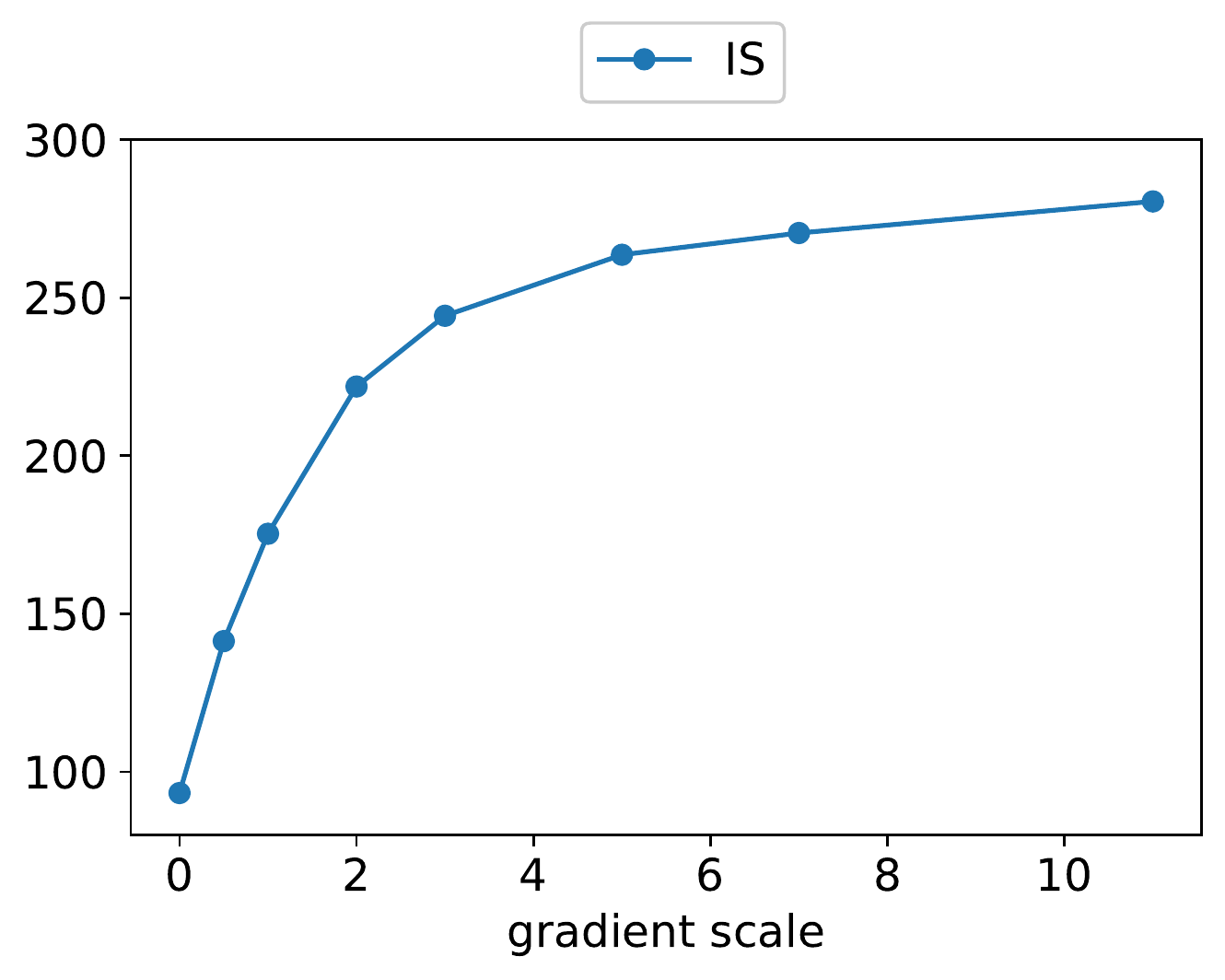}}
    \end{subfigure}
    \begin{subfigure}{0.32\textwidth}
        \centerline{\includegraphics[width=\textwidth]{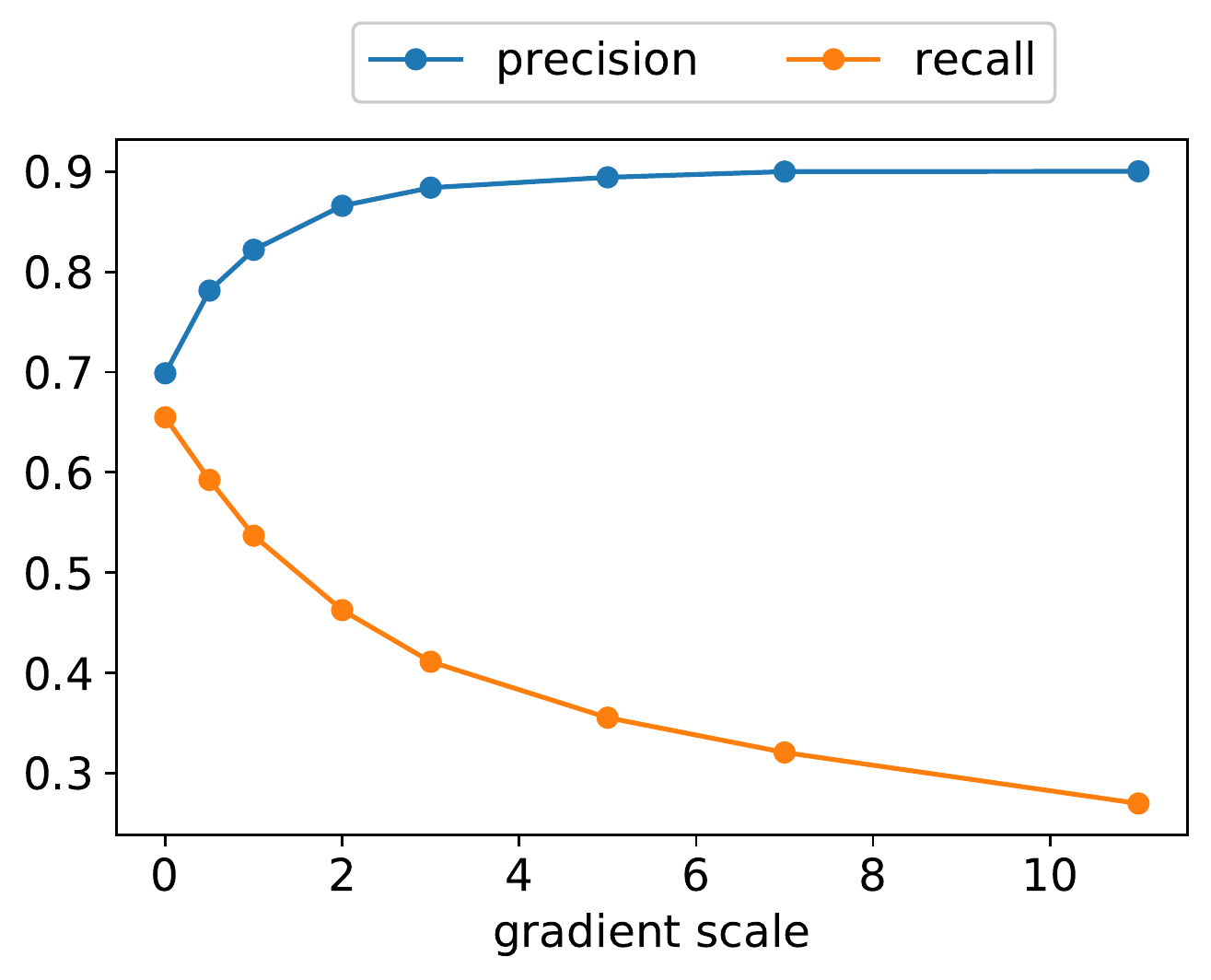}}
    \end{subfigure}
    \end{center}
    \caption{Change in sample quality as we vary scale of the classifier gradients for a class-conditional ImageNet 128$\times$128 model.} 
    \label{fig:guidescale}
\end{figure}

\begin{figure}[t!]
    \begin{center}
    \begin{subfigure}[]{0.48\textwidth}
    \centerline{\includegraphics[width=\textwidth]{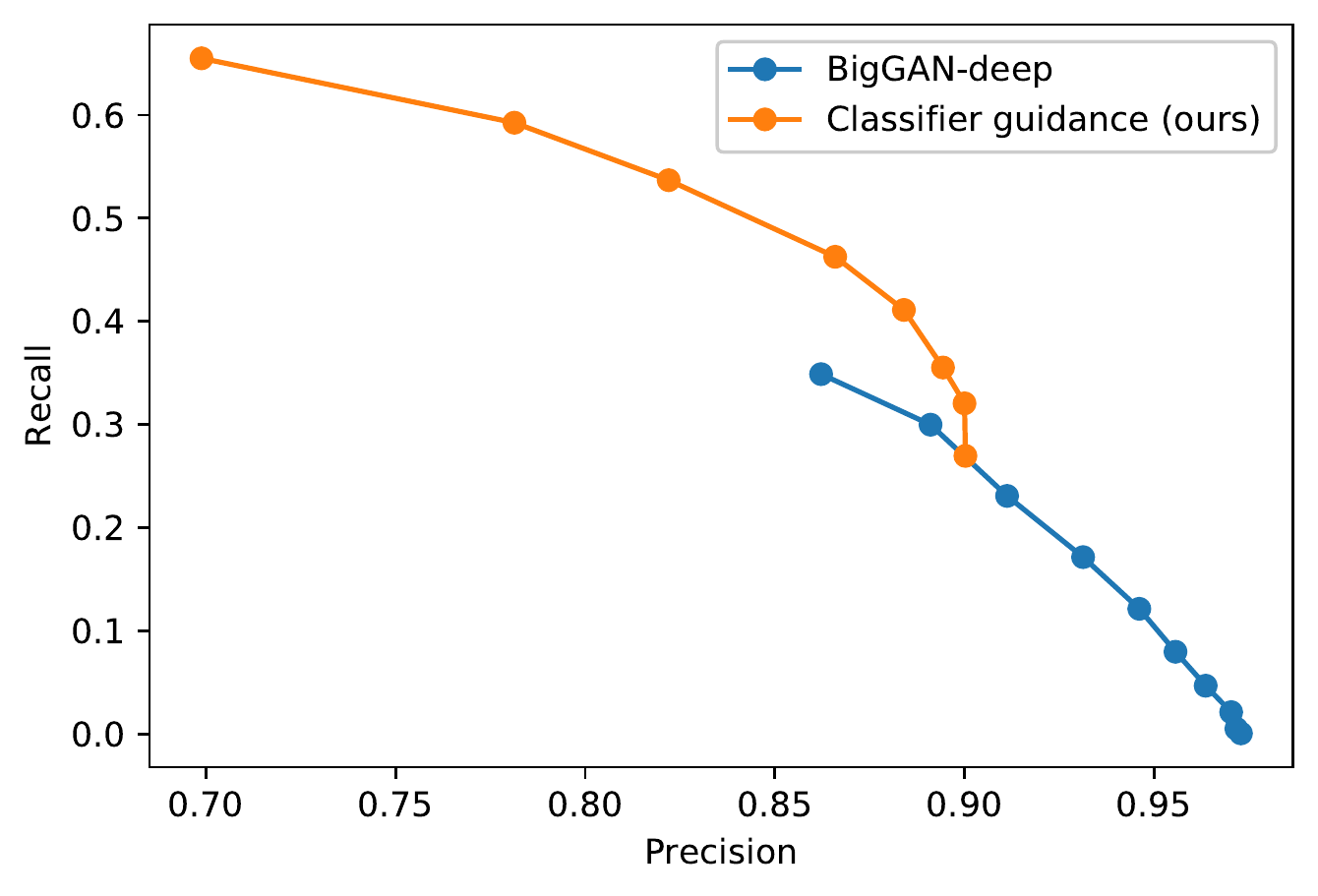}}
    \end{subfigure}\quad
    \begin{subfigure}[]{0.48\textwidth}
    \centerline{\includegraphics[width=\textwidth]{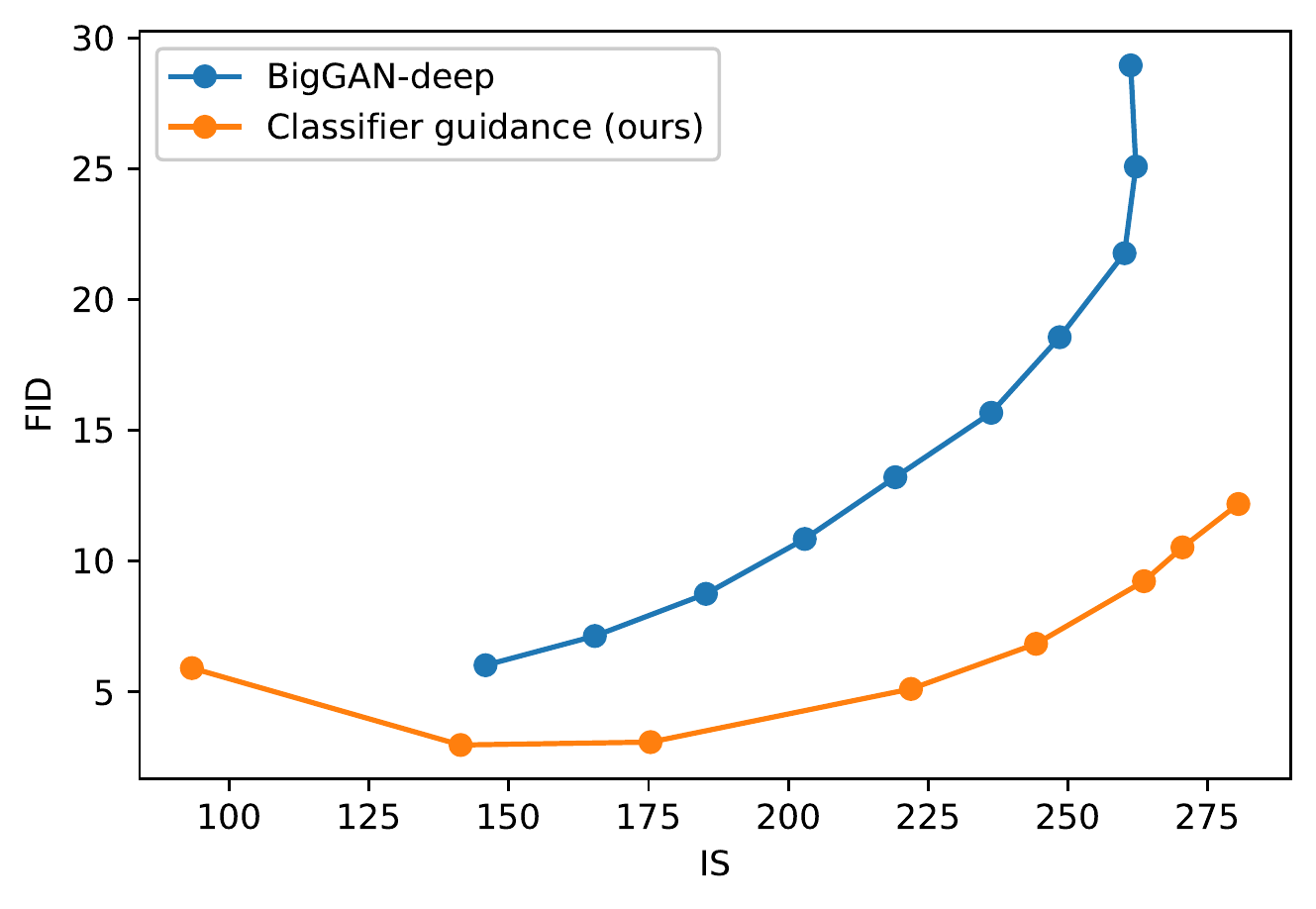}}
    \end{subfigure}
    \caption{\label{fig:gan_comparison_pareto} Trade-offs when varying truncation for BigGAN-deep and gradient scale for classifier guidance. Models are evaluated on ImageNet 128$\times$128. The BigGAN-deep results were produced using the TFHub model \shortcite{tfhubbiggan} at truncation levels $[0.1, 0.2, 0.3, ..., 1.0]$.}
    \end{center}
    \vskip -0.2in
\end{figure}

In the above derivations, we assumed that the underlying diffusion model was unconditional, modeling $p(x)$. It is also possible to train conditional diffusion models, $p(x|y)$, and use classifier guidance in the exact same way. 
Table \ref{tab:guide} shows that the sample quality of both unconditional and conditional models can be greatly improved by classifier guidance. We see that, with a high enough scale, the guided unconditional model can get quite close to the FID of an unguided conditional model, although training directly with the class labels still helps. Guiding a conditional model further improves FID. 

Table \ref{tab:guide} also shows that classifier guidance improves precision at the cost of recall, thus introducing a trade-off in sample fidelity versus diversity. We explicitly evaluate how this trade-off varies with the gradient scale in Figure \ref{fig:guidescale}. We see that scaling the gradients beyond 1.0 smoothly trades off recall (a measure of diversity) for higher precision and IS (measures of fidelity). Since FID and sFID depend on both diversity and fidelity, their best values are obtained at an intermediate point. We also compare our guidance with the truncation trick from BigGAN in Figure \ref{fig:gan_comparison_pareto}. We find that classifier guidance is strictly better than BigGAN-deep when trading off FID for Inception Score. Less clear cut is the precision/recall trade-off, which shows that classifier guidance is only a better choice up until a certain precision threshold, after which point it cannot achieve better precision.

\section{Results}
\label{sec:results}
To evaluate our improved model architecture on unconditional image generation, we train separate diffusion models on three LSUN \shortcite{lsun} classes: bedroom, horse, and cat. To evaluate classifier guidance, we train conditional diffusion models on the ImageNet \shortcite{imagenet} dataset at 128$\times$128, 256$\times$256, and 512$\times$512 resolution.

\subsection{State-of-the-art Image Synthesis}

\begin{table}[t]
    \setlength\tabcolsep{4pt}
    \begin{center}
    \begin{small}
    \begin{subtable}[t]{.5\linewidth}
    \begin{tabular}[t]{lcccc}
    \toprule
    Model            & FID          & sFID       & Prec      & Rec \\
    \\
    \multicolumn{5}{l}{\bf{LSUN Bedrooms} 256$\times$256} \\
    \toprule
    
    DCTransformer${}^\dagger$ \shortcite{dctransformer}    & 6.40          & 6.66      & 0.44      & \bf{0.56} \\
    DDPM \shortcite{ddpm} & 4.89 & 9.07 & 0.60 & 0.45 \\
    IDDPM \shortcite{improved}            & 4.24             & 8.21      & 0.62    & 0.46 \\
    StyleGAN \shortcite{stylegan}         & 2.35	         & 6.62      & 0.59      & 0.48 \\
    \bf{\ablaname{} (dropout)}   & \bf{1.90}          & \bf{5.59}      & \bf{0.66}      & 0.51 \\
    
    \\
    \multicolumn{5}{l}{\bf{LSUN Horses} 256$\times$256} \\
    \toprule
    
    StyleGAN2 \shortcite{stylegan2}        & 3.84          & 6.46      &  0.63      & 0.48 \\
    \bf{\ablaname{}}             & 2.95          & \bf{5.94} &  0.69      & \bf{0.55} \\
    \bf{\ablaname{} (dropout)}   & \bf{2.57}     & 6.81 &  \bf{0.71}  & \bf{0.55} \\
    \\
    \multicolumn{5}{l}{\bf{LSUN Cats} 256$\times$256} \\
    \toprule
    DDPM \shortcite{ddpm} & 17.1 & 12.4 & 0.53 & 0.48 \\
    StyleGAN2 \shortcite{stylegan2}        & 7.25          & \bf{6.33} &  0.58      & 0.43  \\
    \bf{\ablaname{} (dropout)}   & \bf{5.57}     & 6.69      &  \bf{0.63}  & \bf{0.52}  \\
    \\
    \multicolumn{5}{l}{\bf{ImageNet} 64$\times$64} \\
    \toprule
    BigGAN-deep* \shortcite{biggan}     & 4.06         & 3.96      & \bf{0.79} & 0.48  \\
    IDDPM \shortcite{improved}   & 2.92         & \bf{3.79} & 0.74      & 0.62 \\
    \bf{\ablaname{}}             & 2.61         & \bf{3.77} & 0.73      & 0.63  \\
    \bf{\ablaname{} (dropout)}   & \bf{2.07}    & 4.29      & 0.74      & \bf{0.63}  \\     
    \end{tabular}
    \end{subtable}%
    \begin{subtable}[t]{.5\linewidth}
    \begin{tabular}[t]{lcccc}
    \toprule
    Model            & FID          & sFID       & Prec & Rec \\
    \\
    \multicolumn{5}{l}{\bf{ImageNet} 128$\times$128} \\
    \toprule
    BigGAN-deep \shortcite{biggan}              & 6.02          & 7.18          & \bf{0.86}     & 0.35          \\
    LOGAN${}^\dagger$ \shortcite{logan}  & 3.36          &               &               &               \\
    \bf{\ablaname{}}                     & 5.91          & \bf{5.09}     & 0.70          & \bf{0.65}     \\
    \bf{\guidedname{} (25 steps)}      & 5.98          & 7.04          & 0.78          & 0.51          \\
    \bf{\guidedname{}}                & \bf{2.97}	 & \bf{5.09}     & 0.78          & 0.59          \\
    \\
    \multicolumn{5}{l}{\bf{ImageNet} 256$\times$256} \\
    \toprule
    DCTransformer${}^\dagger$ \shortcite{dctransformer}            & 36.51         & 8.24          & 0.36          & \bf{0.67}  \\
    VQ-VAE-2${}^\dagger{}^\ddag$ \shortcite{vqvae2} & 31.11 & 17.38 & 0.36 & 0.57 \\
    IDDPM${}^\ddag$ \shortcite{improved} & 12.26 & 5.42 & 0.70 & 0.62 \\
    SR3${}^\dagger{}^\ddag$ \shortcite{sr3} & 11.30 & & & \\
    BigGAN-deep \shortcite{biggan}              & 6.95          & 7.36          & \bf{0.87}     & 0.28       \\
    \bf{\ablaname{}}                     & 10.94         & 6.02          & 0.69          & 0.63       \\
    \bf{\guidedname{} (25 steps)}      & 5.44	         & 5.32          & 0.81	         & 0.49       \\
    \bf{\guidedname{}}                & \bf{4.59}     & \bf{5.25}     & 0.82          & 0.52       \\
    \\
    \multicolumn{5}{l}{\bf{ImageNet} 512$\times$512} \\
    \toprule
    BigGAN-deep \shortcite{biggan}        & 8.43          & 8.13          & \bf{0.88} & 0.29          \\
    \bf{\ablaname{}}                & 23.24         & 10.19         & 0.73      & \bf{0.60}                \\
    \bf{\guidedname{} (25 steps)} & 8.41 &	9.67    & 0.83      & 0.47               \\
    \bf{\guidedname{}}           & \bf{7.72}     & \bf{6.57}     & 0.87      & 0.42               \\
    \end{tabular}
    \end{subtable}
    \end{small}
    \end{center}
    \caption{Sample quality comparison with state-of-the-art generative models for each task. \ablaname{} refers to our \textbf{a}blated \textbf{d}iffusion \textbf{m}odel, and \guidedname{} additionally uses classifier \textbf{g}uidance. LSUN diffusion models are sampled using 1000 steps (see Appendix \ref{app:lsunsweep}). ImageNet diffusion models are sampled using 250 steps, except when we use the DDIM sampler with 25 steps. *No BigGAN-deep model was available at this resolution, so we trained our own. ${}^\dagger$Values are taken from a previous paper, due to lack of public models or samples. ${}^\ddag$Results use two-resolution stacks.}
    \label{tab:sota}
\end{table}

Table \ref{tab:sota} summarizes our results. Our diffusion models can obtain the best FID on each task, and the best sFID on all but one task. With the improved architecture, we already obtain state-of-the-art image generation on LSUN and ImageNet 64$\times$64. For higher resolution ImageNet, we observe that classifier guidance allows our models to substantially outperform the best GANs. These models obtain perceptual quality similar to GANs, while maintaining a higher coverage of the distribution as measured by recall, and can even do so using only 25 diffusion steps.

\begin{figure}[ht]
    \begin{center}
    \begin{subfigure}[]{0.31\textwidth}
    \centerline{\includegraphics[width=\textwidth]{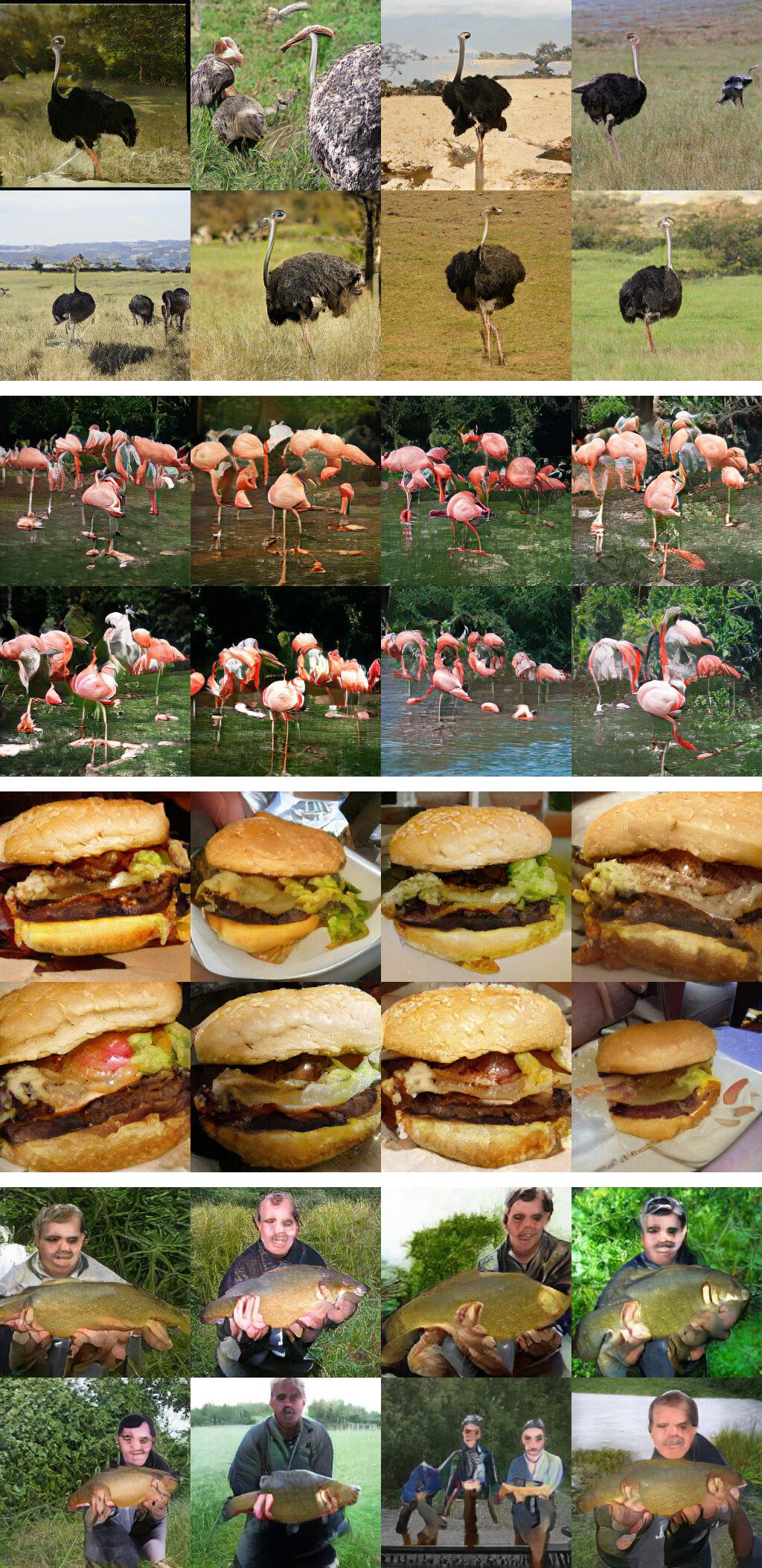}}
    \end{subfigure}\quad
    \begin{subfigure}[]{0.31\textwidth}
    \centerline{\includegraphics[width=\textwidth]{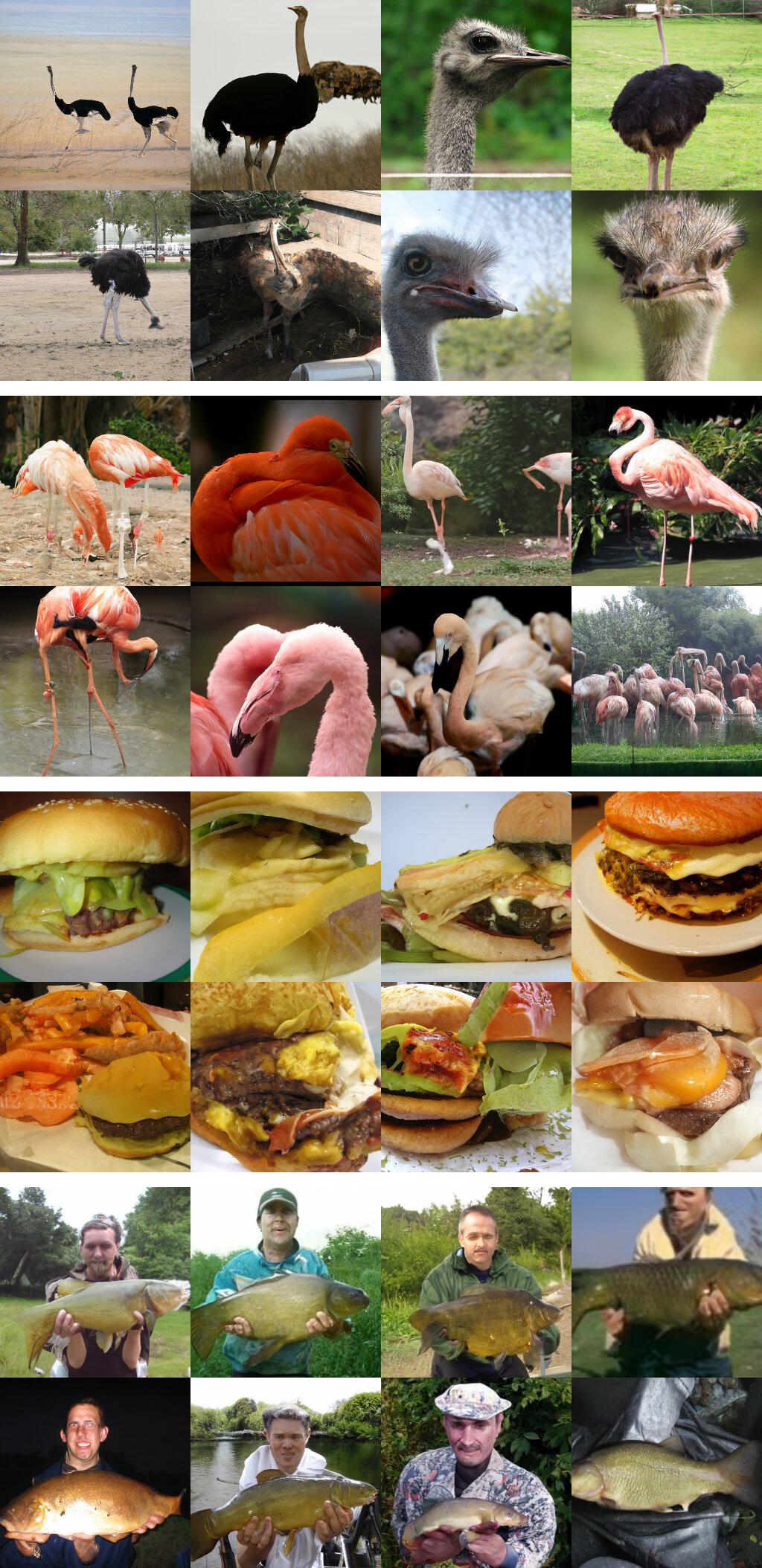}}
    \end{subfigure}\quad
    \begin{subfigure}[]{0.31\textwidth}
    \centerline{\includegraphics[width=\textwidth]{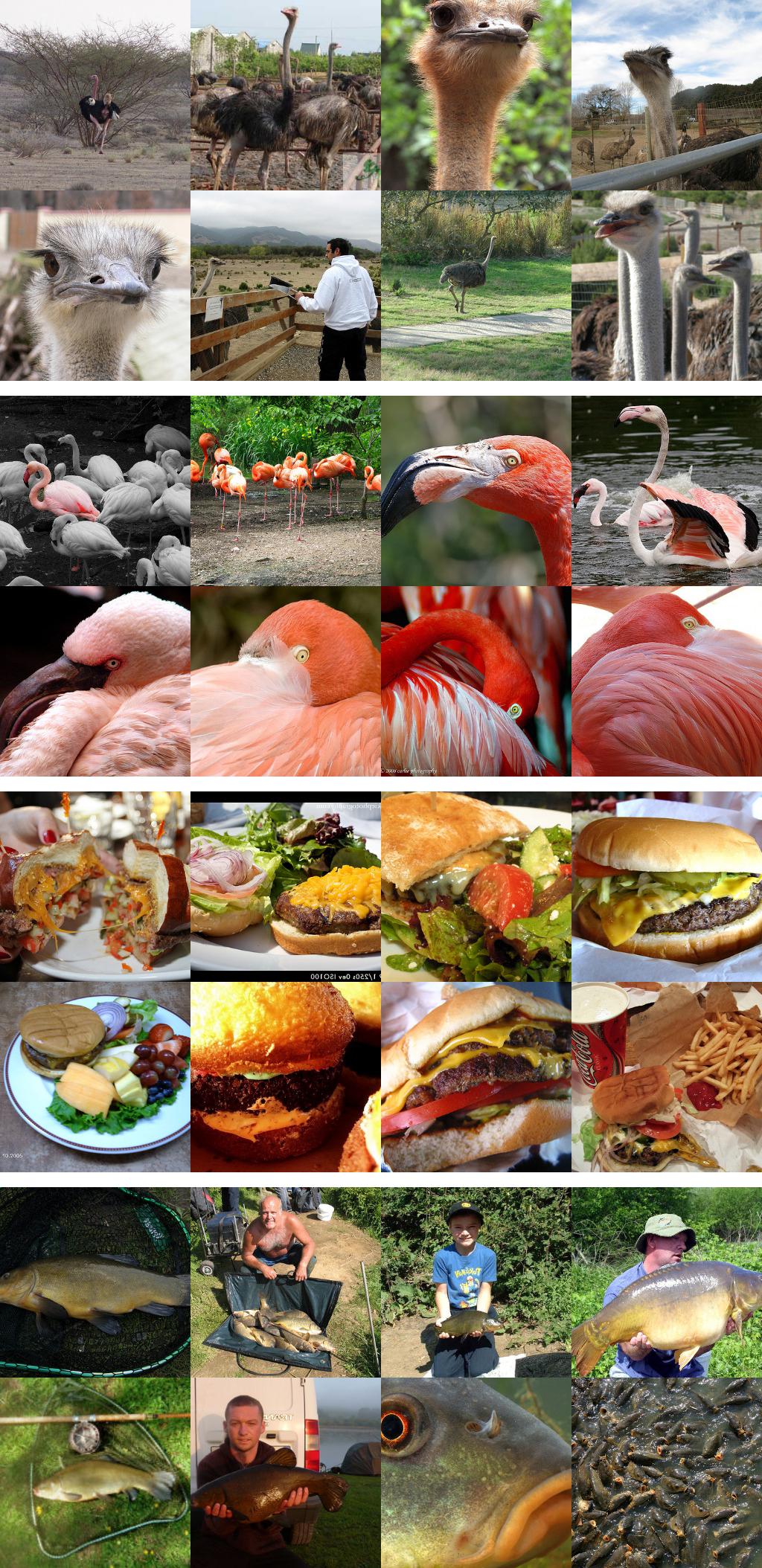}}
    \end{subfigure}
    \caption{\label{fig:diversity} Samples from BigGAN-deep with truncation 1.0 (FID 6.95, left) vs samples from our diffusion model with guidance (FID 4.59, middle) and samples from the training set (right).}
    \end{center}
    \vskip -0.2in
\end{figure}

Figure \ref{fig:diversity} compares random samples from the best BigGAN-deep model to our best diffusion model. While the samples are of similar perceptual quality, the diffusion model contains more modes than the GAN, such as zoomed ostrich heads, single flamingos, different orientations of cheeseburgers, and a tinca fish with no human holding it. We also check our generated samples for nearest neighbors in the Inception-V3 feature space in Appendix \ref{app:neighbors}, and we show additional samples in Appendices \ref{app:fullsamples}-\ref{app:endfullsamples}.

\subsection{Comparison to Upsampling}

We also compare guidance to using a two-stage upsampling stack. \namecite{improved} and \namecite{sr3} train two-stage diffusion models by combining a low-resolution diffusion model with a corresponding upsampling diffusion model. In this approach, the upsampling model is trained to upsample images from the training set, and conditions on low-resolution images that are concatenated channel-wise to the model input using a simple interpolation (e.g. bilinear). During sampling, the low-resolution model produces a sample, and then the upsampling model is conditioned on this sample. This greatly improves FID on ImageNet 256$\times$256, but does not reach the same performance as state-of-the-art models like BigGAN-deep \shortcite{improved,sr3}, as seen in Table \ref{tab:sota}. 

In Table \ref{tab:upsample}, we show that guidance and upsampling improve sample quality along different axes. While upsampling improves precision while keeping a high recall, guidance provides a knob to trade off diversity for much higher precision. We achieve the best FIDs by using guidance at a lower resolution before upsampling to a higher resolution, indicating that these approaches complement one another.

\begin{table}[t]
    \begin{center}
    \begin{small}
    \begin{tabular}{lccccccc}
    \toprule
    Model & $S_{\textit{base}}$ & $S_{\textit{upsample}}$ & FID   & sFID  & IS     & Precision & Recall \\
    \\
    \multicolumn{8}{l}{\bf{ImageNet} 256$\times$256} \\
    \toprule
    ADM & 250 & & 10.94 & 6.02 & 100.98 & 0.69 & \bf{0.63} \\
    ADM-U & 250 & 250 & 7.49 & \bf{5.13} & 127.49 & 0.72 & \bf{0.63} \\
    ADM-G & 250 & & 4.59 & 5.25 & 186.70 & 0.82 & 0.52 \\
    ADM-G, ADM-U & 250 & 250 & \bf{3.94} & 6.14      & \bf{215.84} & \bf{0.83} & 0.53 \\
    \\
    \multicolumn{8}{l}{\bf{ImageNet} 512$\times$512} \\
    \toprule
    ADM & 250   &    & 23.24 &	10.19 & 58.06 & 0.73  & 0.60 \\
    ADM-U & 250 & 250  & 9.96      & \bf{5.62} & 121.78      & 0.75      & \bf{0.64} \\
    ADM-G & 250    &   & 7.72      & 6.57      & 172.71      & \bf{0.87} & 0.42 \\
    ADM-G, ADM-U & 25 & 25    & 5.96      & 12.10     & 187.87      & 0.81      & 0.54 \\
    ADM-G, ADM-U & 250 & 25   & 4.11      & 9.57      & 219.29      & 0.83      & 0.55 \\
    ADM-G, ADM-U & 250 & 250  & \bf{3.85} & 5.86      & \bf{221.72} & 0.84      & 0.53 \\
    \end{tabular}
    \end{small}
    \end{center}
    \caption{Comparing our single, upsampling and classifier guided models. For upsampling, we use the \textbf{u}psampling stack from \namecite{improved} combined with our architecture improvements, which we refer to as ADM-U. The base resolution for the two-stage upsampling models is $64$ and $128$ for the $256$ and $512$ models, respectively. When combining classifier guidance with upsampling, we only guide the lower resolution model.}
    \label{tab:upsample}
    \vskip -0.2in
\end{table}

\section{Related Work}
\label{sec:relwork}

Score based generative models were introduced by \namecite{scorematching} as a way of modeling a data distribution using its gradients, and then sampling using Langevin dynamics \shortcite{langevin}. \namecite{ddpm} found a connection between this method and diffusion models \shortcite{dickstein}, and achieved excellent sample quality by leveraging this connection. After this breakthrough work, many works followed up with more promising results: \namecite{diffwave} and \namecite{wavegrad} demonstrated that diffusion models work well for audio; \namecite{adversarial} found that a GAN-like setup could improve samples from these models; \namecite{sde} explored ways to leverage techniques from stochastic differential equations to improve the sample quality obtained by score-based models; \namecite{ddim} and \namecite{improved} proposed methods to improve sampling speed; \namecite{improved} and \namecite{sr3} demonstrated promising results on the difficult ImageNet generation task using upsampling diffusion models. Also related to diffusion models, and following the work of \namecite{dickstein}, \namecite{variationalwalkback} described a technique for learning a model with learned iterative generation steps, and found that it could achieve good image samples when trained with a likelihood objective.

One missing element from previous work on diffusion models is a way to trade off diversity for fidelity. Other generative techniques provide natural levers for this trade-off. \namecite{biggan} introduced the truncation trick for GANs, wherein the latent vector is sampled from a truncated normal distribution. They found that increasing truncation naturally led to a decrease in diversity but an increase in fidelity. More recently, \namecite{vqvae2} proposed to use classifier rejection sampling to filter out bad samples from an autoregressive likelihood-based model, and found that this technique improved FID. Most likelihood-based models also allow for low-temperature sampling \shortcite{temperature}, which provides a natural way to emphasize modes of the data distribution (see Appendix \ref{app:temperature}).

Other likelihood-based models have been shown to produce high-fidelity image samples. VQ-VAE \shortcite{vqvae} and VQ-VAE-2 \shortcite{vqvae2} are autoregressive models trained on top of quantized latent codes, greatly reducing the computational resources required to train these models on large images. These models produce diverse and high quality images, but still fall short of GANs without expensive rejection sampling and special metrics to compensate for blurriness. DCTransformer \shortcite{dctransformer} is a related method which relies on a more intelligent compression scheme. VAEs are another promising class of likelihood-based models, and recent methods such as NVAE \shortcite{nvae} and VDVAE \shortcite{vdvae} have successfully been applied to difficult image generation domains. Energy-based models are another class of likelihood-based models with a rich history \shortcite{temperature,helmholtz,contrastive}. Sampling from the EBM distribution is challenging, and \namecite{genconv} demonstrate that Langevin dynamics can be used to sample coherent images from these models. \namecite{yilunenergy} further improve upon this approach, obtaining high quality images. More recently, \namecite{diffusionebm} incorporate diffusion steps into an energy-based model, and find that doing so improves image samples from these models.

Other works have controlled generative models with a pre-trained classifier. For example, an emerging body of work \shortcite{clipglass,styleclip,bigsleep} aims to optimize GAN latent spaces for text prompts using pre-trained CLIP \shortcite{clip} models. More similar to our work, \namecite{sde} uses a classifier to generate class-conditional CIFAR-10 images with a diffusion model. In some cases, classifiers can act as stand-alone generative models. For example, \namecite{robustgeneration} demonstrate that a robust image classifier can be used as a stand-alone generative model, and \namecite{jem} train a model which is jointly a classifier and an energy-based model.

\section{Limitations and Future Work}

While we believe diffusion models are an extremely promising direction for generative modeling, they are still slower than GANs at sampling time due to the use of multiple denoising steps (and therefore forward passes). One promising work in this direction is from \namecite{ddimdistill}, who explore a way to distill the DDIM sampling process into a single step model. The samples from the single step model are not yet competitive with GANs, but are much better than previous single-step likelihood-based models. Future work in this direction might be able to completely close the sampling speed gap between diffusion models and GANs without sacrificing image quality.

Our proposed classifier guidance technique is currently limited to labeled datasets, and we have provided no effective strategy for trading off diversity for fidelity on unlabeled datasets. In the future, our method could be extended to unlabeled data by clustering samples to produce synthetic labels \shortcite{unlabeledgan} or by training discriminative models to predict when samples are in the true data distribution or from the sampling distribution.

The effectiveness of classifier guidance demonstrates that we can obtain powerful generative models from the gradients of a classification function. This could be used to condition pre-trained models in a plethora of ways, for example by conditioning an image generator with a text caption using a noisy version of CLIP \shortcite{clip}, similar to recent methods that guide GANs using text prompts \shortcite{clipglass,styleclip,bigsleep}. It also suggests that large unlabeled datasets could be leveraged in the future to pre-train powerful diffusion models that can later be improved by using a classifier with desirable properties.

\section{Conclusion}
\label{sec:conclusion}
We have shown that diffusion models, a class of likelihood-based models with a stationary training objective, can obtain better sample quality than state-of-the-art GANs. Our improved architecture is sufficient to achieve this on unconditional image generation tasks, and our classifier guidance technique allows us to do so on class-conditional tasks. In the latter case, we find that the scale of the classifier gradients can be adjusted to trade off diversity for fidelity. These guided diffusion models can reduce the sampling time gap between GANs and diffusion models, although diffusion models still require multiple forward passes during sampling. Finally, by combining guidance with upsampling, we can further improve sample quality on high-resolution conditional image synthesis. 

\section{Acknowledgements}
We thank Alec Radford, Mark Chen, Pranav Shyam and Raul Puri for providing feedback on this work. 

\setcitestyle{numbers}
\bibliographystyle{plainnat}
\bibliography{main.bib}

\begin{thebibliography}{73}
\providecommand{\natexlab}[1]{#1}
\providecommand{\url}[1]{\texttt{#1}}
\expandafter\ifx\csname urlstyle\endcsname\relax
  \providecommand{\doi}[1]{doi: #1}\else
  \providecommand{\doi}{doi: \begingroup \urlstyle{rm}\Url}\fi

\bibitem[Ackley et~al.(1985)Ackley, Hinton, and Sejnowski]{temperature}
David Ackley, Geoffrey Hinton, and Terrence Sejnowski.
\newblock A learning algorithm for boltzmann machines.
\newblock \emph{Cognitive science, 9(1):147-169}, 1985.

\bibitem[Adverb(2021)]{bigsleep}
Adverb.
\newblock The big sleep.
\newblock \url{https://twitter.com/advadnoun/status/1351038053033406468}, 2021.

\bibitem[Barratt and Sharma(2018)]{noteoninceptionscore}
Shane Barratt and Rishi Sharma.
\newblock A note on the inception score.
\newblock \emph{\href{https://arxiv.org/abs/1801.01973}{arXiv:1801.01973}},
  2018.

\bibitem[Brock et~al.(2016)Brock, Lim, Ritchie, and Weston]{orthoreg}
Andrew Brock, Theodore Lim, J.~M. Ritchie, and Nick Weston.
\newblock Neural photo editing with introspective adversarial networks.
\newblock \emph{\href{https://arxiv.org/abs/1609.07093}{arXiv:1609.07093}},
  2016.

\bibitem[Brock et~al.(2018)Brock, Donahue, and Simonyan]{biggan}
Andrew Brock, Jeff Donahue, and Karen Simonyan.
\newblock Large scale gan training for high fidelity natural image synthesis.
\newblock \emph{\href{https://arxiv.org/abs/1809.11096}{arXiv:1809.11096}},
  2018.

\bibitem[Brown et~al.(2020)Brown, Mann, Ryder, Subbiah, Kaplan, Dhariwal,
  Neelakantan, Shyam, Sastry, Askell, Agarwal, Herbert-Voss, Krueger, Henighan,
  Child, Ramesh, Ziegler, Wu, Winter, Hesse, Chen, Sigler, Litwin, Gray, Chess,
  Clark, Berner, McCandlish, Radford, Sutskever, and Amodei]{gpt3}
Tom~B. Brown, Benjamin Mann, Nick Ryder, Melanie Subbiah, Jared Kaplan,
  Prafulla Dhariwal, Arvind Neelakantan, Pranav Shyam, Girish Sastry, Amanda
  Askell, Sandhini Agarwal, Ariel Herbert-Voss, Gretchen Krueger, Tom Henighan,
  Rewon Child, Aditya Ramesh, Daniel~M. Ziegler, Jeffrey Wu, Clemens Winter,
  Christopher Hesse, Mark Chen, Eric Sigler, Mateusz Litwin, Scott Gray,
  Benjamin Chess, Jack Clark, Christopher Berner, Sam McCandlish, Alec Radford,
  Ilya Sutskever, and Dario Amodei.
\newblock Language models are few-shot learners.
\newblock \emph{\href{https://arxiv.org/abs/2005.14165}{arXiv:2005.14165}},
  2020.

\bibitem[Chen et~al.(2020{\natexlab{a}})Chen, Radford, Child, Wu, Jun, Luan,
  and Sutskever]{igpt}
Mark Chen, Alec Radford, Rewon Child, Jeffrey Wu, Heewoo Jun, David Luan, and
  Ilya Sutskever.
\newblock Generative pretraining from pixels.
\newblock In \emph{International Conference on Machine Learning}, pages
  1691--1703. PMLR, 2020{\natexlab{a}}.

\bibitem[Chen et~al.(2020{\natexlab{b}})Chen, Zhang, Zen, Weiss, Norouzi, and
  Chan]{wavegrad}
Nanxin Chen, Yu~Zhang, Heiga Zen, Ron~J. Weiss, Mohammad Norouzi, and William
  Chan.
\newblock Wavegrad: Estimating gradients for waveform generation.
\newblock \emph{\href{https://arxiv.org/abs/2009.00713}{arXiv:2009.00713}},
  2020{\natexlab{b}}.

\bibitem[Child(2021)]{vdvae}
Rewon Child.
\newblock Very deep vaes generalize autoregressive models and can outperform
  them on images.
\newblock \emph{\href{https://arxiv.org/abs/2011.10650}{arXiv:2011.10650}},
  2021.

\bibitem[Dayan et~al.(1995)Dayan, Hinton, Neal, and Zemel]{helmholtz}
Peter Dayan, Geoffrey~E Hinton, Radford~M Neal, and Richard~S Zemel.
\newblock The helmholtz machine.
\newblock \emph{Neural computation}, 7\penalty0 (5):\penalty0 889--904, 1995.

\bibitem[de~Vries et~al.(2017)de~Vries, Strub, Mary, Larochelle, Pietquin, and
  Courville]{cganmodulat}
Harm de~Vries, Florian Strub, Jérémie Mary, Hugo Larochelle, Olivier
  Pietquin, and Aaron Courville.
\newblock Modulating early visual processing by language.
\newblock \emph{\href{https://arxiv.org/abs/1707.00683}{arXiv:1707.00683}},
  2017.

\bibitem[DeepMind(2018)]{tfhubbiggan}
DeepMind.
\newblock Biggan-deep 128x128 on tensorflow hub.
\newblock \url{https://tfhub.dev/deepmind/biggan-deep-128/1}, 2018.

\bibitem[Dhariwal et~al.(2020)Dhariwal, Jun, Payne, Kim, Radford, and
  Sutskever]{jukebox}
Prafulla Dhariwal, Heewoo Jun, Christine Payne, Jong~Wook Kim, Alec Radford,
  and Ilya Sutskever.
\newblock Jukebox: A generative model for music.
\newblock \emph{\href{https://arxiv.org/abs/2005.00341}{arXiv:2005.00341}},
  2020.

\bibitem[Donahue and Simonyan(2019)]{bigbigan}
Jeff Donahue and Karen Simonyan.
\newblock Large scale adversarial representation learning.
\newblock \emph{\href{https://arxiv.org/abs/1907.02544}{arXiv:1907.02544}},
  2019.

\bibitem[Du and Mordatch(2019)]{yilunenergy}
Yilun Du and Igor Mordatch.
\newblock Implicit generation and generalization in energy-based models.
\newblock \emph{\href{https://arxiv.org/abs/1903.08689}{arXiv:1903.08689}},
  2019.

\bibitem[Dumoulin et~al.(2017)Dumoulin, Shlens, and Kudlur]{cganbn}
Vincent Dumoulin, Jonathon Shlens, and Manjunath Kudlur.
\newblock A learned representation for artistic style.
\newblock \emph{\href{https://arxiv.org/abs/1610.07629}{arXiv:1610.07629}},
  2017.

\bibitem[Galatolo et~al.(2021)Galatolo, Cimino, and Vaglini]{clipglass}
Federico~A. Galatolo, Mario G. C.~A. Cimino, and Gigliola Vaglini.
\newblock Generating images from caption and vice versa via clip-guided
  generative latent space search.
\newblock \emph{\href{https://arxiv.org/abs/2102.01645}{arXiv:2102.01645}},
  2021.

\bibitem[Gao et~al.(2020)Gao, Song, Poole, Wu, and Kingma]{diffusionebm}
Ruiqi Gao, Yang Song, Ben Poole, Ying~Nian Wu, and Diederik~P. Kingma.
\newblock Learning energy-based models by diffusion recovery likelihood.
\newblock \emph{\href{https://arxiv.org/abs/2012.08125}{arXiv:2012.08125}},
  2020.

\bibitem[Goodfellow et~al.(2014)Goodfellow, Pouget-Abadie, Mirza, Xu,
  Warde-Farley, Ozair, Courville, and Bengio]{gan}
Ian~J. Goodfellow, Jean Pouget-Abadie, Mehdi Mirza, Bing Xu, David
  Warde-Farley, Sherjil Ozair, Aaron Courville, and Yoshua Bengio.
\newblock Generative adversarial networks.
\newblock \emph{\href{https://arxiv.org/abs/1406.2661}{arXiv:1406.2661}}, 2014.

\bibitem[Google(2018)]{tpu}
Google.
\newblock Cloud tpus.
\newblock \url{https://cloud.google.com/tpu/}, 2018.

\bibitem[Goyal et~al.(2017)Goyal, Ke, Ganguli, and Bengio]{variationalwalkback}
Anirudh Goyal, Nan~Rosemary Ke, Surya Ganguli, and Yoshua Bengio.
\newblock Variational walkback: Learning a transition operator as a stochastic
  recurrent net.
\newblock \emph{\href{https://arxiv.org/abs/1711.02282}{arXiv:1711.02282}},
  2017.

\bibitem[Grathwohl et~al.(2019)Grathwohl, Wang, Jacobsen, Duvenaud, Norouzi,
  and Swersky]{jem}
Will Grathwohl, Kuan-Chieh Wang, Jörn-Henrik Jacobsen, David Duvenaud,
  Mohammad Norouzi, and Kevin Swersky.
\newblock Your classifier is secretly an energy based model and you should
  treat it like one.
\newblock \emph{\href{https://arxiv.org/abs/1912.03263}{arXiv:1912.03263}},
  2019.

\bibitem[Heusel et~al.(2017)Heusel, Ramsauer, Unterthiner, Nessler, and
  Hochreiter]{fid}
Martin Heusel, Hubert Ramsauer, Thomas Unterthiner, Bernhard Nessler, and Sepp
  Hochreiter.
\newblock Gans trained by a two time-scale update rule converge to a local nash
  equilibrium.
\newblock \emph{Advances in Neural Information Processing Systems 30 (NIPS
  2017)}, 2017.

\bibitem[Hinton(2002)]{contrastive}
Geoffrey~E Hinton.
\newblock Training products of experts by minimizing contrastive divergence.
\newblock \emph{Neural computation}, 14\penalty0 (8):\penalty0 1771--1800,
  2002.

\bibitem[Ho et~al.(2020)Ho, Jain, and Abbeel]{ddpm}
Jonathan Ho, Ajay Jain, and Pieter Abbeel.
\newblock Denoising diffusion probabilistic models.
\newblock \emph{\href{https://arxiv.org/abs/2006.11239}{arXiv:2006.11239}},
  2020.

\bibitem[Jolicoeur-Martineau et~al.(2020)Jolicoeur-Martineau, Piché-Taillefer,
  des Combes, and Mitliagkas]{adversarial}
Alexia Jolicoeur-Martineau, Rémi Piché-Taillefer, Rémi~Tachet des Combes,
  and Ioannis Mitliagkas.
\newblock Adversarial score matching and improved sampling for image
  generation.
\newblock \emph{\href{https://arxiv.org/abs/2009.05475}{arXiv:2009.05475}},
  2020.

\bibitem[Karras et~al.(2019{\natexlab{a}})Karras, Laine, and Aila]{stylegan}
Tero Karras, Samuli Laine, and Timo Aila.
\newblock A style-based generator architecture for generative adversarial
  networks.
\newblock
  \emph{\href{https://arxiv.org/abs/arXiv:1812.04948}{arXiv:arXiv:1812.04948}},
  2019{\natexlab{a}}.

\bibitem[Karras et~al.(2019{\natexlab{b}})Karras, Laine, Aittala, Hellsten,
  Lehtinen, and Aila]{stylegan2}
Tero Karras, Samuli Laine, Miika Aittala, Janne Hellsten, Jaakko Lehtinen, and
  Timo Aila.
\newblock Analyzing and improving the image quality of stylegan.
\newblock \emph{\href{https://arxiv.org/abs/1912.04958}{arXiv:1912.04958}},
  2019{\natexlab{b}}.

\bibitem[Kingma and Ba(2014)]{adam}
Diederik~P. Kingma and Jimmy Ba.
\newblock Adam: A method for stochastic optimization.
\newblock \emph{\href{https://arxiv.org/abs/1412.6980}{arXiv:1412.6980}}, 2014.

\bibitem[Kong et~al.(2020)Kong, Ping, Huang, Zhao, and Catanzaro]{diffwave}
Zhifeng Kong, Wei Ping, Jiaji Huang, Kexin Zhao, and Bryan Catanzaro.
\newblock Diffwave: A versatile diffusion model for audio synthesis.
\newblock \emph{\href{https://arxiv.org/abs/2009.09761}{arXiv:2009.09761}},
  2020.

\bibitem[Krizhevsky et~al.(2009)Krizhevsky, Nair, and Hinton]{cifar10}
Alex Krizhevsky, Vinod Nair, and Geoffrey Hinton.
\newblock {CIFAR-10 (Canadian Institute for Advanced Research)}, 2009.
\newblock URL \url{http://www.cs.toronto.edu/~kriz/cifar.html}.

\bibitem[Kynkäänniemi et~al.(2019)Kynkäänniemi, Karras, Laine, Lehtinen,
  and Aila]{precrecall}
Tuomas Kynkäänniemi, Tero Karras, Samuli Laine, Jaakko Lehtinen, and Timo
  Aila.
\newblock Improved precision and recall metric for assessing generative models.
\newblock \emph{\href{https://arxiv.org/abs/1904.06991}{arXiv:1904.06991}},
  2019.

\bibitem[Lin et~al.(2016)Lin, Milan, Shen, and Reid]{refinenet}
Guosheng Lin, Anton Milan, Chunhua Shen, and Ian Reid.
\newblock Refinenet: Multi-path refinement networks for high-resolution
  semantic segmentation.
\newblock \emph{\href{https://arxiv.org/abs/1611.06612}{arXiv:1611.06612}},
  2016.

\bibitem[Liu et~al.(2015)Liu, Luo, Wang, and Tang]{celeba}
Ziwei Liu, Ping Luo, Xiaogang Wang, and Xiaoou Tang.
\newblock Deep learning face attributes in the wild.
\newblock In \emph{Proceedings of International Conference on Computer Vision
  (ICCV)}, December 2015.

\bibitem[Loshchilov and Hutter(2017)]{adamw}
Ilya Loshchilov and Frank Hutter.
\newblock Decoupled weight decay regularization.
\newblock \emph{\href{https://arxiv.org/abs/1711.05101}{arXiv:1711.05101}},
  2017.

\bibitem[Lucic et~al.(2019)Lucic, Tschannen, Ritter, Zhai, Bachem, and
  Gelly]{unlabeledgan}
Mario Lucic, Michael Tschannen, Marvin Ritter, Xiaohua Zhai, Olivier Bachem,
  and Sylvain Gelly.
\newblock High-fidelity image generation with fewer labels.
\newblock \emph{\href{https://arxiv.org/abs/1903.02271}{arXiv:1903.02271}},
  2019.

\bibitem[Luhman and Luhman(2021)]{ddimdistill}
Eric Luhman and Troy Luhman.
\newblock Knowledge distillation in iterative generative models for improved
  sampling speed.
\newblock \emph{\href{https://arxiv.org/abs/2101.02388}{arXiv:2101.02388}},
  2021.

\bibitem[Micikevicius et~al.(2017)Micikevicius, Narang, Alben, Diamos, Elsen,
  Garcia, Ginsburg, Houston, Kuchaiev, Venkatesh, and Wu]{lossscaling}
Paulius Micikevicius, Sharan Narang, Jonah Alben, Gregory Diamos, Erich Elsen,
  David Garcia, Boris Ginsburg, Michael Houston, Oleksii Kuchaiev, Ganesh
  Venkatesh, and Hao Wu.
\newblock Mixed precision training.
\newblock \emph{\href{https://arxiv.org/abs/1710.03740}{arXiv:1710.03740}},
  2017.

\bibitem[Mirza and Osindero(2014)]{cgan}
Mehdi Mirza and Simon Osindero.
\newblock Conditional generative adversarial nets.
\newblock \emph{\href{https://arxiv.org/abs/1411.1784}{arXiv:1411.1784}}, 2014.

\bibitem[Miyato and Koyama(2018)]{cganproj}
Takeru Miyato and Masanori Koyama.
\newblock cgans with projection discriminator.
\newblock \emph{\href{https://arxiv.org/abs/1802.05637}{arXiv:1802.05637}},
  2018.

\bibitem[Miyato et~al.(2018)Miyato, Kataoka, Koyama, and Yoshida]{sngan}
Takeru Miyato, Toshiki Kataoka, Masanori Koyama, and Yuichi Yoshida.
\newblock Spectral normalization for generative adversarial networks.
\newblock \emph{\href{https://arxiv.org/abs/1802.05957}{arXiv:1802.05957}},
  2018.

\bibitem[Nash et~al.(2021)Nash, Menick, Dieleman, and Battaglia]{dctransformer}
Charlie Nash, Jacob Menick, Sander Dieleman, and Peter~W. Battaglia.
\newblock Generating images with sparse representations.
\newblock \emph{\href{https://arxiv.org/abs/2103.03841}{arXiv:2103.03841}},
  2021.

\bibitem[Nichol and Dhariwal(2021)]{improved}
Alex Nichol and Prafulla Dhariwal.
\newblock Improved denoising diffusion probabilistic models.
\newblock \emph{\href{https://arxiv.org/abs/2102.09672}{arXiv:2102.09672}},
  2021.

\bibitem[NVIDIA(2019)]{stylegan2repo}
NVIDIA.
\newblock Stylegan2.
\newblock \url{https://github.com/NVlabs/stylegan2}, 2019.

\bibitem[Parmar et~al.(2021)Parmar, Zhang, and Zhu]{cleanfid}
Gaurav Parmar, Richard Zhang, and Jun-Yan Zhu.
\newblock On buggy resizing libraries and surprising subtleties in fid
  calculation.
\newblock \emph{\href{https://arxiv.org/abs/2104.11222}{arXiv:2104.11222}},
  2021.

\bibitem[Paszke et~al.(2019)Paszke, Gross, Massa, Lerer, Bradbury, Chanan,
  Killeen, Lin, Gimelshein, Antiga, et~al.]{pytorch}
Adam Paszke, Sam Gross, Francisco Massa, Adam Lerer, James Bradbury, Gregory
  Chanan, Trevor Killeen, Zeming Lin, Natalia Gimelshein, Luca Antiga, et~al.
\newblock Pytorch: An imperative style, high-performance deep learning library.
\newblock \emph{\href{https://arxiv.org/abs/1912.01703}{arXiv:1912.01703}},
  2019.

\bibitem[Patashnik et~al.(2021)Patashnik, Wu, Shechtman, Cohen-Or, and
  Lischinski]{styleclip}
Or~Patashnik, Zongze Wu, Eli Shechtman, Daniel Cohen-Or, and Dani Lischinski.
\newblock Styleclip: Text-driven manipulation of stylegan imagery.
\newblock \emph{\href{https://arxiv.org/abs/2103.17249}{arXiv:2103.17249}},
  2021.

\bibitem[Perez et~al.(2017)Perez, Strub, de~Vries, Dumoulin, and
  Courville]{film}
Ethan Perez, Florian Strub, Harm de~Vries, Vincent Dumoulin, and Aaron
  Courville.
\newblock Film: Visual reasoning with a general conditioning layer.
\newblock \emph{\href{https://arxiv.org/abs/1709.07871}{arXiv:1709.07871}},
  2017.

\bibitem[Radford et~al.(2021)Radford, Kim, Hallacy, Ramesh, Goh, Agarwal,
  Sastry, Askell, Mishkin, Clark, Krueger, and Sutskever]{clip}
Alec Radford, Jong~Wook Kim, Chris Hallacy, Aditya Ramesh, Gabriel Goh,
  Sandhini Agarwal, Girish Sastry, Amanda Askell, Pamela Mishkin, Jack Clark,
  Gretchen Krueger, and Ilya Sutskever.
\newblock Learning transferable visual models from natural language
  supervision.
\newblock \emph{\href{https://arxiv.org/abs/2103.00020}{arXiv:2103.00020}},
  2021.

\bibitem[Ramesh et~al.(2021)Ramesh, Pavlov, Goh, Gray, Voss, Radford, Chen, and
  Sutskever]{dalle}
Aditya Ramesh, Mikhail Pavlov, Gabriel Goh, Scott Gray, Chelsea Voss, Alec
  Radford, Mark Chen, and Ilya Sutskever.
\newblock Zero-shot text-to-image generation.
\newblock \emph{\href{https://arxiv.org/abs/2102.12092}{arXiv:2102.12092}},
  2021.

\bibitem[Razavi et~al.(2019)Razavi, van~den Oord, and Vinyals]{vqvae2}
Ali Razavi, Aaron van~den Oord, and Oriol Vinyals.
\newblock Generating diverse high-fidelity images with {VQ-VAE-2}.
\newblock \emph{\href{https://arxiv.org/abs/1906.00446}{arXiv:1906.00446}},
  2019.

\bibitem[Russakovsky et~al.(2014)Russakovsky, Deng, Su, Krause, Satheesh, Ma,
  Huang, Karpathy, Khosla, Bernstein, Berg, and Fei-Fei]{imagenet}
Olga Russakovsky, Jia Deng, Hao Su, Jonathan Krause, Sanjeev Satheesh, Sean Ma,
  Zhiheng Huang, Andrej Karpathy, Aditya Khosla, Michael Bernstein,
  Alexander~C. Berg, and Li~Fei-Fei.
\newblock Imagenet large scale visual recognition challenge.
\newblock \emph{\href{https://arxiv.org/abs/1409.0575}{arXiv:1409.0575}}, 2014.

\bibitem[Saharia et~al.(2021)Saharia, Ho, Chan, Salimans, Fleet, and
  Norouzi]{sr3}
Chitwan Saharia, Jonathan Ho, William Chan, Tim Salimans, David~J. Fleet, and
  Mohammad Norouzi.
\newblock Image super-resolution via iterative refinement.
\newblock
  \emph{\href{https://arxiv.org/abs/arXiv:2104.07636}{arXiv:arXiv:2104.07636}},
  2021.

\bibitem[Salimans et~al.(2016)Salimans, Goodfellow, Zaremba, Cheung, Radford,
  and Chen]{inceptionscore}
Tim Salimans, Ian Goodfellow, Wojciech Zaremba, Vicki Cheung, Alec Radford, and
  Xi~Chen.
\newblock Improved techniques for training gans.
\newblock \emph{\href{https://arxiv.org/abs/1606.03498}{arXiv:1606.03498}},
  2016.

\bibitem[Santurkar et~al.(2019)Santurkar, Tsipras, Tran, Ilyas, Engstrom, and
  Madry]{robustgeneration}
Shibani Santurkar, Dimitris Tsipras, Brandon Tran, Andrew Ilyas, Logan
  Engstrom, and Aleksander Madry.
\newblock Image synthesis with a single (robust) classifier.
\newblock \emph{\href{https://arxiv.org/abs/1906.09453}{arXiv:1906.09453}},
  2019.

\bibitem[Sohl-Dickstein et~al.(2015)Sohl-Dickstein, Weiss, Maheswaranathan, and
  Ganguli]{dickstein}
Jascha Sohl-Dickstein, Eric~A. Weiss, Niru Maheswaranathan, and Surya Ganguli.
\newblock Deep unsupervised learning using nonequilibrium thermodynamics.
\newblock \emph{\href{https://arxiv.org/abs/1503.03585}{arXiv:1503.03585}},
  2015.

\bibitem[Song et~al.(2020{\natexlab{a}})Song, Meng, and Ermon]{ddim}
Jiaming Song, Chenlin Meng, and Stefano Ermon.
\newblock Denoising diffusion implicit models.
\newblock \emph{\href{https://arxiv.org/abs/2010.02502}{arXiv:2010.02502}},
  2020{\natexlab{a}}.

\bibitem[Song and Ermon(2020{\natexlab{a}})]{improvedscore}
Yang Song and Stefano Ermon.
\newblock Improved techniques for training score-based generative models.
\newblock \emph{\href{https://arxiv.org/abs/2006.09011}{arXiv:2006.09011}},
  2020{\natexlab{a}}.

\bibitem[Song and Ermon(2020{\natexlab{b}})]{scorematching}
Yang Song and Stefano Ermon.
\newblock Generative modeling by estimating gradients of the data distribution.
\newblock
  \emph{\href{https://arxiv.org/abs/arXiv:1907.05600}{arXiv:arXiv:1907.05600}},
  2020{\natexlab{b}}.

\bibitem[Song et~al.(2020{\natexlab{b}})Song, Sohl-Dickstein, Kingma, Kumar,
  Ermon, and Poole]{sde}
Yang Song, Jascha Sohl-Dickstein, Diederik~P. Kingma, Abhishek Kumar, Stefano
  Ermon, and Ben Poole.
\newblock Score-based generative modeling through stochastic differential
  equations.
\newblock \emph{\href{https://arxiv.org/abs/2011.13456}{arXiv:2011.13456}},
  2020{\natexlab{b}}.

\bibitem[Szegedy et~al.(2013)Szegedy, Zaremba, Sutskever, Bruna, Erhan,
  Goodfellow, and Fergus]{adversarialexamples}
Christian Szegedy, Wojciech Zaremba, Ilya Sutskever, Joan Bruna, Dumitru Erhan,
  Ian Goodfellow, and Rob Fergus.
\newblock Intriguing properties of neural networks.
\newblock \emph{\href{https://arxiv.org/abs/1312.6199}{arXiv:1312.6199}}, 2013.

\bibitem[Szegedy et~al.(2015)Szegedy, Vanhoucke, Ioffe, Shlens, and
  Wojna]{inceptionv3}
Christian Szegedy, Vincent Vanhoucke, Sergey Ioffe, Jonathon Shlens, and
  Zbigniew Wojna.
\newblock Rethinking the inception architecture for computer vision.
\newblock \emph{\href{https://arxiv.org/abs/1512.00567}{arXiv:1512.00567}},
  2015.

\bibitem[Vahdat and Kautz(2020)]{nvae}
Arash Vahdat and Jan Kautz.
\newblock Nvae: A deep hierarchical variational autoencoder.
\newblock \emph{\href{https://arxiv.org/abs/2007.03898}{arXiv:2007.03898}},
  2020.

\bibitem[van~den Oord et~al.(2016)van~den Oord, Dieleman, Zen, Simonyan,
  Vinyals, Graves, Kalchbrenner, Senior, and Kavukcuoglu]{wavenet}
Aaron van~den Oord, Sander Dieleman, Heiga Zen, Karen Simonyan, Oriol Vinyals,
  Alex Graves, Nal Kalchbrenner, Andrew Senior, and Koray Kavukcuoglu.
\newblock Wavenet: A generative model for raw audio.
\newblock \emph{\href{https://arxiv.org/abs/1609.03499}{arXiv:1609.03499}},
  2016.

\bibitem[van~den Oord et~al.(2017)van~den Oord, Vinyals, and
  Kavukcuoglu]{vqvae}
Aaron van~den Oord, Oriol Vinyals, and Koray Kavukcuoglu.
\newblock Neural discrete representation learning.
\newblock \emph{\href{https://arxiv.org/abs/1711.00937}{arXiv:1711.00937}},
  2017.

\bibitem[Vaswani et~al.(2017)Vaswani, Shazeer, Parmar, Uszkoreit, Jones, Gomez,
  Kaiser, and Polosukhin]{transformer}
Ashish Vaswani, Noam Shazeer, Niki Parmar, Jakob Uszkoreit, Llion Jones,
  Aidan~N. Gomez, Lukasz Kaiser, and Illia Polosukhin.
\newblock Attention is all you need.
\newblock \emph{\href{https://arxiv.org/abs/1706.03762}{arXiv:1706.03762}},
  2017.

\bibitem[Welling and Teh(2011)]{langevin}
Max Welling and Yee~W Teh.
\newblock Bayesian learning via stochastic gradient langevin dynamics.
\newblock In \emph{Proceedings of the 28th international conference on machine
  learning (ICML-11)}, pages 681--688. Citeseer, 2011.

\bibitem[Wu et~al.(2019)Wu, Donahue, Balduzzi, Simonyan, and Lillicrap]{logan}
Yan Wu, Jeff Donahue, David Balduzzi, Karen Simonyan, and Timothy Lillicrap.
\newblock Logan: Latent optimisation for generative adversarial networks.
\newblock \emph{\href{https://arxiv.org/abs/1912.00953}{arXiv:1912.00953}},
  2019.

\bibitem[Wu and He(2018)]{groupnorm}
Yuxin Wu and Kaiming He.
\newblock Group normalization.
\newblock \emph{\href{https://arxiv.org/abs/1803.08494}{arXiv:1803.08494}},
  2018.

\bibitem[Xie et~al.(2016)Xie, Lu, Zhu, and Wu]{genconv}
Jianwen Xie, Yang Lu, Song-Chun Zhu, and Ying~Nian Wu.
\newblock A theory of generative convnet.
\newblock \emph{\href{https://arxiv.org/abs/1602.03264}{arXiv:1602.03264}},
  2016.

\bibitem[Yu et~al.(2015)Yu, Seff, Zhang, Song, Funkhouser, and Xiao]{lsun}
Fisher Yu, Ari Seff, Yinda Zhang, Shuran Song, Thomas Funkhouser, and Jianxiong
  Xiao.
\newblock Lsun: Construction of a large-scale image dataset using deep learning
  with humans in the loop.
\newblock \emph{\href{https://arxiv.org/abs/1506.03365}{arXiv:1506.03365}},
  2015.

\bibitem[Zhang et~al.(2016)Zhang, Xu, Li, Zhang, Wang, Huang, and
  Metaxas]{stackgan}
Han Zhang, Tao Xu, Hongsheng Li, Shaoting Zhang, Xiaogang Wang, Xiaolei Huang,
  and Dimitris Metaxas.
\newblock Stackgan: Text to photo-realistic image synthesis with stacked
  generative adversarial networks.
\newblock \emph{\href{https://arxiv.org/abs/1612.03242}{arXiv:1612.03242}},
  2016.

\bibitem[Zhu(2018)]{thoprepo}
Ligeng Zhu.
\newblock Thop.
\newblock \url{https://github.com/Lyken17/pytorch-OpCounter}, 2018.

\end{thebibliography}

\newpage
\appendix
\section{Computational Requirements}
\label{app:compute}

Compute is essential to modern machine learning applications, and more compute typically yields better results. It is thus important to compare our method's compute requirements to competing methods. In this section, we demonstrate that we can achieve results better than StyleGAN2 and BigGAN-deep with the same or lower compute budget.

\subsection{Throughput}
We first benchmark the throughput of our models in Table \ref{tab:throughput}. For the theoretical throughput, we measure the theoretical FLOPs for our model using THOP \shortcite{thoprepo}, and assume 100\% utilization of an NVIDIA Tesla V100 (120 TFLOPs), while for the actual throughput we use measured wall-clock time. We include communication time across two machines whenever our training batch size doesn't fit on a single machine, where each of our machines has 8 V100s.

We find that a naive implementation of our models in PyTorch 1.7 is very inefficient, utilizing only 20-30\% of the hardware. We also benchmark our optimized version, which use larger per-GPU batch sizes, fused GroupNorm-Swish and fused Adam CUDA ops. For our ImageNet 128$\times$128 model in particular, we find that we can increase the per-GPU batch size from 4 to 32 while still fitting in GPU memory, and this makes a large utilization difference. Our implementation is still far from optimal, and further optimizations should allow us to reach higher levels of utilization.

\begin{table}[h]
    \begin{center}
    \begin{small}
    \begin{tabular}{llccc}
    \toprule
    \multirow{2}*{Model} & \multirow{2}*{Implementation} & Batch Size & Throughput & \multirow{2}*{Utilization} \\
    & & per GPU & Imgs per V100-sec & \\
    \midrule
    \multirow{3}*{64$\times$64} & Theoretical & - & 182.3 & 100\% \\
    & Naive & 32 & 37.0 & 20\% \\
    & Optimized & 96 & 74.1 & 41\% \\
    \midrule
    \multirow{3}*{128$\times$128} & Theoretical & - & 65.2 & 100\% \\
    & Naive & 4 & 11.5 & 18\% \\
    & Optimized & 32 & 24.8 & 38\% \\
    \midrule
    \multirow{3}*{256$\times$256} & Theoretical & - & 17.9 & 100\% \\
    & Naive & 4 & 4.4 & 25\% \\
    & Optimized & 8 & 6.4 & 36\%\\
    \midrule
    \multirow{3}*{64 $\to$ 256} & Theoretical & - & 31.7 & 100\% \\
    & Naive & 4 & 6.3 & 20\% \\
    & Optimized & 12 & 9.5 & 30\% \\
    \midrule
    \multirow{3}*{128 $\to$ 512} & Theoretical & - & 8.0 & 100\% \\
    & Naive & 2 & 1.9 & 24\% \\
    & Optimized & 2 & 2.3 & 29\% \\
    \bottomrule
    \end{tabular}
    \end{small}
    \end{center}
    \caption{Throughput of our ImageNet models, measured in Images per V100-sec.}
    \label{tab:throughput}
    \vskip -0.2in
\end{table}

\subsection{Early stopping}
In addition, we can train for many fewer iterations while maintaining sample quality superior to BigGAN-deep. Table \ref{tab:imagenet128early} and \ref{tab:imagenet256early} evaluate our ImageNet 128$\times$128 and 256$\times$256 models throughout training. We can see that the ImageNet 128$\times$128 model beats BigGAN-deep's FID (6.02) after 500K training iterations, only one eighth of the way through training. Similarly, the ImageNet 256$\times$256 model beats BigGAN-deep after 750K iterations, roughly a third of the way through training.

\begin{table}[h]
    \begin{center}
    \begin{small}
    \begin{tabular}{ccccc}
    \toprule
    Iterations & FID & sFID & Precision & Recall \\
    \midrule
    250K & 7.97 & 6.48 & 0.80 & 0.50 \\
    500K & 5.31 & 5.97 & 0.83 & 0.49 \\
    1000K & 4.10 & 5.80 & 0.81 & 0.51 \\
    2000K & 3.42 & 5.69 & 0.83 & 0.53 \\
    4360K & 3.09 & 5.59 & 0.82 & 0.54 \\
    \bottomrule
    \end{tabular}
    \end{small}
    \end{center}
    \caption{Evaluating an ImageNet 128$\times$128 model throughout training (classifier scale 1.0).}
    \label{tab:imagenet128early}
    \vskip -0.2in
\end{table}

\begin{table}[h]
    \begin{center}
    \begin{small}
    \begin{tabular}{ccccc}
    \toprule
    Iterations & FID & sFID & Precision & Recall \\
    \midrule
    250K & 12.21 & 6.15 & 0.78 & 0.50 \\
    500K & 7.95 & 5.51 & 0.81 & 0.50 \\
    750K & 6.49 & 5.39 & 0.81 & 0.50 \\
    1000K & 5.74 & 5.29 & 0.81 & 0.52 \\
    1500K & 5.01 & 5.20 & 0.82 & 0.52 \\
    1980K & 4.59 & 5.25 & 0.82  & 0.52 \\
    \bottomrule
    \end{tabular}
    \end{small}
    \end{center}
    \caption{Evaluating an ImageNet 256$\times$256 model throughout training (classifier scale 1.0).}
    \label{tab:imagenet256early}
    \vskip -0.2in
\end{table}

\subsection{Compute comparison}
Finally, in Table \ref{tab:computecomparison} we compare the compute of our models with StyleGAN2 and BigGAN-deep, and show we can obtain better FIDs with a similar compute budget. For BigGAN-deep, \namecite{biggan} do not explicitly describe the compute requirements for training their models, but rather provide rough estimates in terms of days on a Google TPUv3 pod \shortcite{tpu}. We convert their TPU-v3 estimates to V100 days according to 2 TPU-v3 day = 1 V100 day. For StyleGAN2, we use the reported throughput of 25M images over 32 days 13 hour on one V100 for config-f \shortcite{stylegan2repo}. We note that our classifier training is relatively lightweight compared to training the generative model. 

\begin{table}[h]
    \setlength\tabcolsep{3.5pt}
    \begin{center}
    \begin{small}
    \begin{tabular}{lccccccc}
    \toprule
    Model & Generator & Classifier & Total   & FID & sFID & Precision & Recall \\
          & Compute    & Compute  & Compute &     &      &           &        \\
    \toprule
    \\
    \multicolumn{6}{l}{\bf{LSUN Horse} 256$\times$256} \\
    \toprule
    StyleGAN2 \shortcite{stylegan2}  & & & 130 & 3.84          & 6.46      &  0.63      & 0.48 \\
    ADM (250K)       & 116 & - & \bf{116} & 2.95          & \bf{5.94} &  0.69      & \bf{0.55} \\
    ADM (dropout, 250K)                     & 116 & - & \bf{116} & \bf{2.57}    & 6.81 &  \bf{0.71}  & \bf{0.55} \\
    \\
    \multicolumn{6}{l}{\bf{LSUN Cat} 256$\times$256} \\
    \toprule
    StyleGAN2 \shortcite{stylegan2}  & & & 115 & 7.25          & \bf{6.33} &  0.58      & 0.43 \\
    ADM (dropout, 200K)                     & 92  & - & \bf{92} & \bf{5.57}    & 6.69      &  \bf{0.63}  & \bf{0.52} \\
    \\
    \multicolumn{6}{l}{\bf{ImageNet} 128$\times$128} \\
    \toprule
    BigGAN-deep \shortcite{biggan}  & & & 64-128 & 6.02      & 7.18      & \bf{0.86} & 0.35 \\
    ADM-G (4360K)                   & 521 & 9 & 530       & \bf 3.09 & \bf 5.59 & 0.82 & \bf 0.54 \\
    ADM-G (450K)                    & 54 & 9 & \bf{63} & 5.67 & 6.19 & 0.82 & 0.49 \\    
    \\
    \multicolumn{6}{l}{\bf{ImageNet} 256$\times$256} \\
    \toprule
    BigGAN-deep \shortcite{biggan}  &     &    & 128-256  & 6.95      & 7.36      & \bf{0.87}& 0.28 \\
    ADM-G (1980K)                   & 916 & 46 & 962      & 4.59      & \bf{5.25} & 0.82     & 0.52 \\
    ADM-G (750K)                    & 347 & 46 & 393      & 6.49      & 5.39      & 0.81     & 0.50 \\
    ADM-G (750K)                    & 347 & 14${}^\dagger$ & 361      & 6.68      & 5.34      & 0.81     & 0.51 \\
    ADM-G (540K), ADM-U (500K)      & 329 & 30 & 359      & \bf{3.85} & 5.86      & 0.84     & 0.53 \\
    ADM-G (540K), ADM-U (150K)      & 219 & 30 & 249      & 4.15      & 6.14      & 0.82	 & \bf{0.54} \\
    ADM-G (200K), ADM-U (150K)      & 110 & 10${}^\ddagger$ & \bf{126} & 4.93      & 5.82      & 0.82     & 0.52  \\
    \\
    \multicolumn{6}{l}{\bf{ImageNet} 512$\times$512} \\
    \toprule
    BigGAN-deep \shortcite{biggan}  & &    & 256-512 & 8.43      & 8.13      & \bf{0.88} & 0.29 \\
    ADM-G (4360K), ADM-U (1050K)   & 1878 & 36 & 1914 & \bf{3.85} & \bf{5.86} & 0.84      & \bf{0.53} \\
    ADM-G (500K), ADM-U (100K) & 189 & 9* & \bf{198} & 7.59 & 6.84 & 0.84 & \bf{0.53}  \\
    \bottomrule
    \end{tabular}
    \end{small}
    \end{center}
    \caption{Training compute requirements for our diffusion models compared to StyleGAN2 and BigGAN-deep. Training iterations for each diffusion model are mentioned in parenthesis. Compute is measured in V100-days. ${}^\dagger$ImageNet 256$\times$256 classifier with 150K iterations (instead of 500K). ${}^\ddagger$ImageNet 64$\times$64 classifier with batch size 256 (instead of 1024). *ImageNet 128$\times$128 classifier with batch size 256 (instead of 1024).}
    \label{tab:computecomparison}
    \vskip -0.2in
\end{table}
\clearpage
\section{Detailed Formulation of DDPM}
\label{app:detailedbackground}
Here, we provide a detailed review of the formulation of Gaussian diffusion models from \namecite{ddpm}. We start by defining our data distribution $x_0 \sim q(x_0)$ and a Markovian noising process $q$ which gradually adds noise to the data to produce noised samples $x_1$ through $x_T$. In particular, each step of the noising process adds Gaussian noise according to some variance schedule given by $\beta_t$:
\begin{alignat}{2}
q(x_t|x_{t-1}) &\coloneqq \mathcal{N}(x_t; \sqrt{1-\beta_t}x_{t-1}, \beta_t \mathbf{I})
\end{alignat}

\namecite{ddpm} note that we need not apply $q$ repeatedly to sample from $x_t \sim q(x_t|x_0)$. Instead, $q(x_t|x_0)$ can be expressed as a Gaussian distribution. With $\alpha_t \coloneqq 1 - \beta_t$ and $\bar{\alpha}_t \coloneqq \prod_{s=0}^{t} \alpha_s$
\begin{alignat}{2}
    q(x_t|x_0) &= \mathcal{N}(x_t; \sqrt{\bar{\alpha}_t} x_0, (1-\bar{\alpha}_t) \mathbf{I}) \\
    &= \sqrt{\bar{\alpha}_t} x_0 + \epsilon \sqrt{1-\bar{\alpha}_t},\text{  } \epsilon \sim \mathcal{N}(0, \mathbf{I}) \label{eq:jumpnoise}
\end{alignat}

Here, $1 - \bar{\alpha}_t$ tells us the variance of the noise for an arbitrary timestep, and we could equivalently use this to define the noise schedule instead of $\beta_t$.

Using Bayes theorem, one finds that the posterior $q(x_{t-1}|x_t,x_0)$ is also a Gaussian with mean $\tilde{\mu}_t(x_t,x_0)$ and variance $\tilde{\beta}_t$ defined as follows:

\begin{alignat}{2}
    \tilde{\mu}_t(x_t,x_0) &\coloneqq \frac{\sqrt{\bar{\alpha}_{t-1}}\beta_t}{1-\bar{\alpha}_t}x_0 + \frac{\sqrt{\alpha_t}(1-\bar{\alpha}_{t-1})}{1-\bar{\alpha}_t} x_t \label{eq:mutilde} \\
    \tilde{\beta}_t &\coloneqq \frac{1-\bar{\alpha}_{t-1}}{1-\bar{\alpha}_t} \beta_t \label{eq:betatilde} \\
    q(x_{t-1}|x_t,x_0) &= \mathcal{N}(x_{t-1}; \tilde{\mu}(x_t, x_0), \tilde{\beta}_t \mathbf{I}) \label{eq:posterior}
\end{alignat}

If we wish to sample from the data distribution $q(x_0)$, we can first sample from $q(x_T)$ and then sample reverse steps $q(x_{t-1}|x_t)$ until we reach $x_0$. Under reasonable settings for $\beta_t$ and $T$, the distribution $q(x_T)$ is nearly an isotropic Gaussian distribution, so sampling $x_T$ is trivial. All that is left is to approximate $q(x_{t-1}|x_t)$ using a neural network, since it cannot be computed exactly when the data distribution is unknown. To this end, \namecite{dickstein} note that $q(x_{t-1}|x_t)$ approaches a diagonal Gaussian distribution as $T \to \infty$ and correspondingly $\beta_t \to 0$, so it is sufficient to train a neural network to predict a mean $\mu_{\theta}$ and a diagonal covariance matrix $\Sigma_{\theta}$:
\begin{alignat}{2}
p_{\theta}(x_{t-1}|x_t) &\coloneqq \mathcal{N}(x_{t-1};\mu_{\theta}(x_t, t), \Sigma_{\theta}(x_t, t)) \label{eq:ptheta}
\end{alignat}

To train this model such that $p(x_0)$ learns the true data distribution $q(x_0)$, we can optimize the following variational lower-bound $L_{\text{vlb}}$ for $p_{\theta}(x_0)$:
\begin{alignat}{2}
    L_{\text{vlb}} &\coloneqq L_0 + L_1 + ... + L_{T-1} + L_T \label{eq:loss} \\
    L_{0} &\coloneqq -\log p_{\theta}(x_0 | x_1) \label{eq:loss0} \\
    L_{t-1} &\coloneqq \kld{q(x_{t-1}|x_t,x_0)}{p_{\theta}(x_{t-1}|x_t)} \label{eq:losst} \\
    L_{T} &\coloneqq \kld{q(x_T | x_0)}{p(x_T)} \label{eq:lossT}
\end{alignat}

While the above objective is well-justified, \namecite{ddpm} found that a different objective produces better samples in practice. In particular, they do not directly parameterize $\mu_{\theta}(x_t,t)$ as a neural network, but instead train a model $\epsilon_{\theta}(x_t,t)$ to predict $\epsilon$ from Equation \ref{eq:jumpnoise}. This simplified objective is defined as follows:
\begin{alignat}{2}
    L_{\text{simple}} &\coloneqq E_{t \sim [1,T],x_0 \sim q(x_0), \epsilon \sim \mathcal{N}(0, \mathbf{I})}[||\epsilon - \epsilon_{\theta}(x_t, t)||^2] \label{eq:lsimple}
\end{alignat}

During sampling, we can use substitution to derive $\mu_{\theta}(x_t,t)$ from $\epsilon_{\theta}(x_t,t)$:
\begin{alignat}{2}
    \mu_{\theta}(x_t,t) &= \frac{1}{\sqrt{\alpha_t}}\left(x_t-\frac{1-\alpha_t}{\sqrt{1-\bar{\alpha}_t}}\epsilon_{\theta}(x_t, t)\right) \label{eq:mufromeps}
\end{alignat}

Note that $L_{\text{simple}}$ does not provide any learning signal for $\Sigma_{\theta}(x_t,t)$. \namecite{ddpm} find that instead of learning $\Sigma_{\theta}(x_t,t)$, they can fix it to a constant, choosing either $\beta_t \mathbf{I}$ or $\tilde{\beta}_t \mathbf{I}$. These values correspond to upper and lower bounds for the true reverse step variance.
\section{Nearest Neighbors for Samples}
\label{app:neighbors}
\begin{figure}[h]
    \begin{center}
    \includegraphics[width=0.9\textwidth]{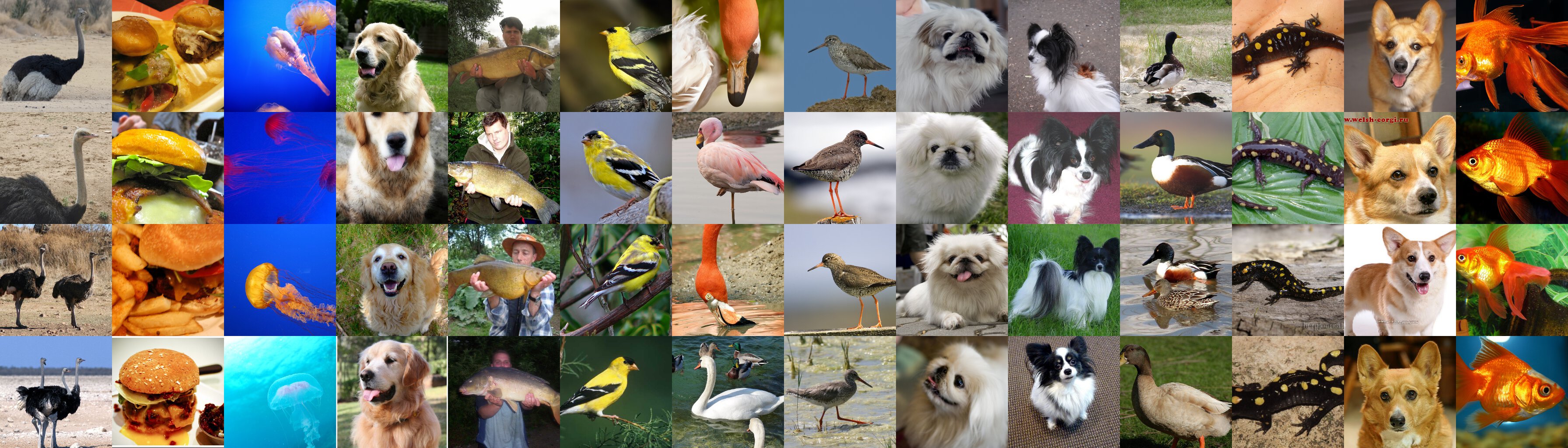} \\
    \vspace{0.3cm}
    \includegraphics[width=0.9\textwidth]{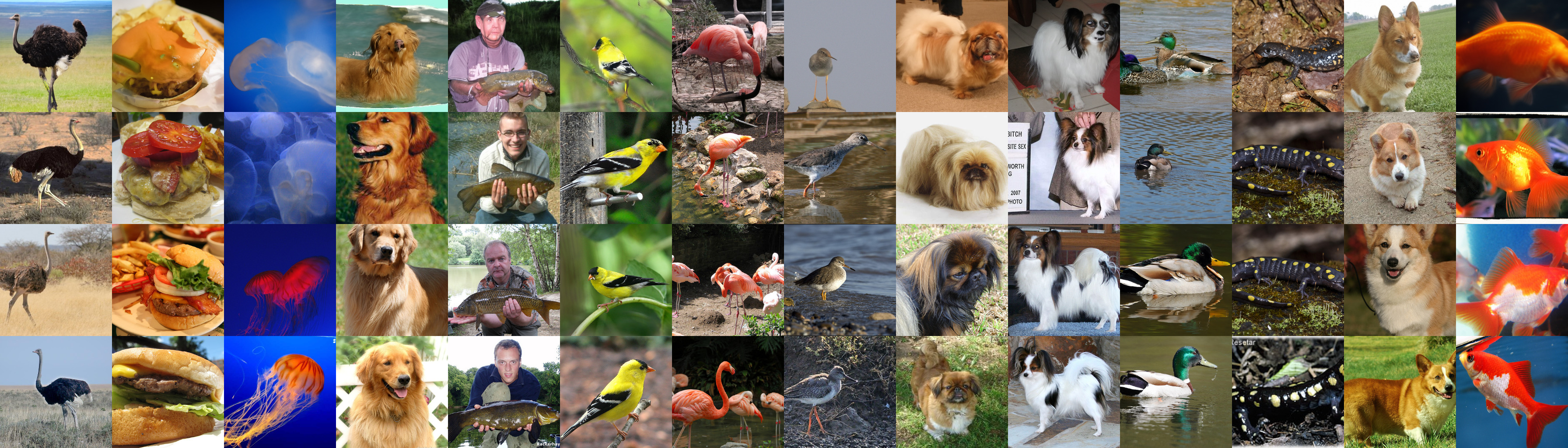}
    
    \caption{\label{fig:neighbors} Nearest neighbors for samples from a classifier guided model on ImageNet 256$\times$256. For each image, the top row is a sample, and the remaining rows are the top 3 nearest neighbors from the dataset. The top samples were generated with classifier scale 1 and 250 diffusion sampling steps (FID 4.59). The bottom samples were generated with classifier scale 2.5 and 25 DDIM steps (FID 5.44).}
    \end{center}
\end{figure}

Our models achieve their best FID when using a classifier to reduce the diversity of the generations. One might fear that such a process could cause the model to recall existing images from the training dataset, especially as the classifier scale is increased. To test this, we looked at the nearest neighbors (in InceptionV3 \shortcite{inceptionv3} feature space) for a handful of samples. Figure \ref{fig:neighbors} shows our results, revealing that the samples are indeed unique and not stored in the training set.

\section{Effect of Varying the Classifier Scale}
\begin{figure}[h]
    \centering
    \includegraphics[width=0.9\textwidth]{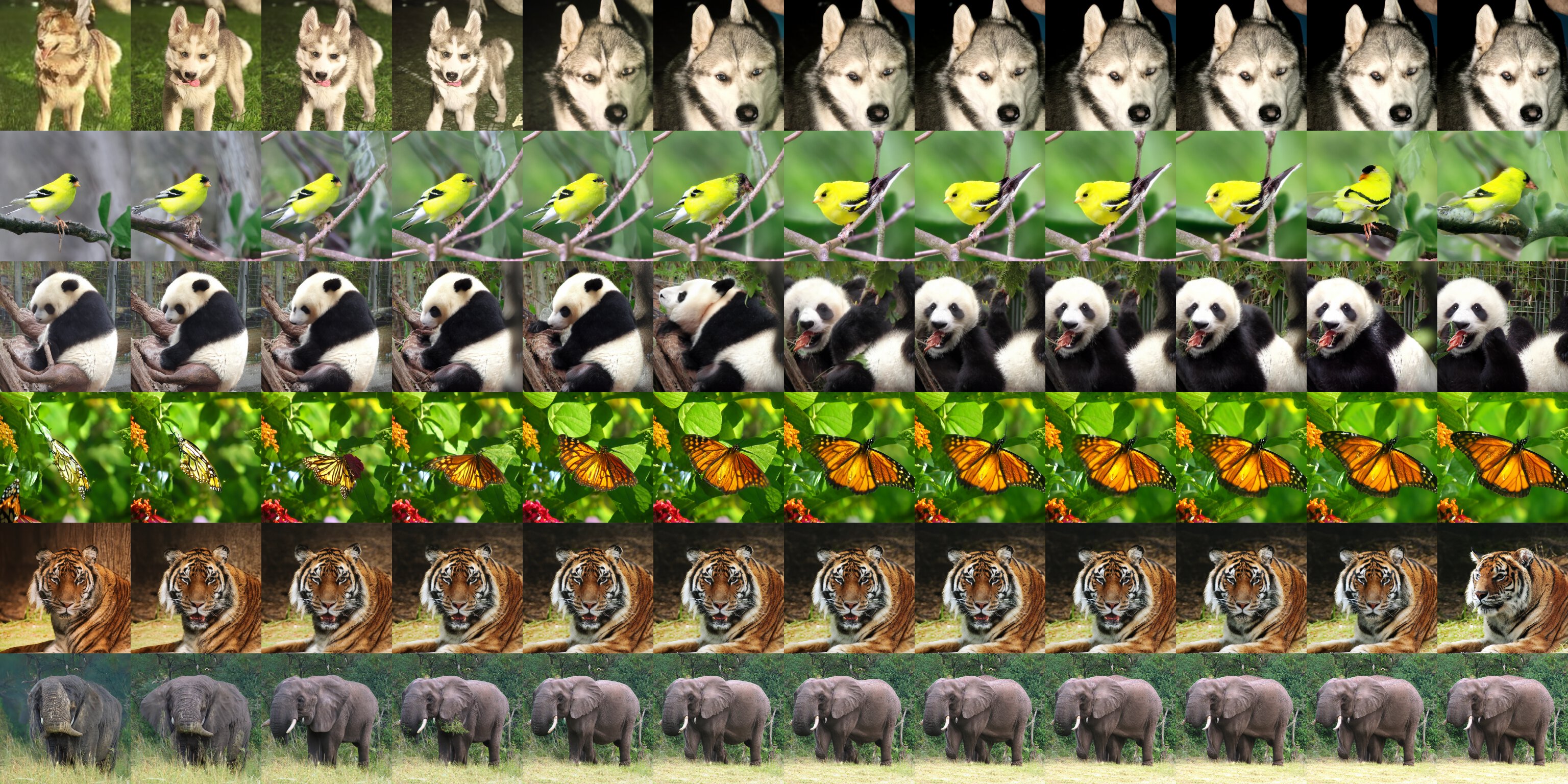}
    \caption{Samples when increasing the classifier scale from 0.0 (left) to 5.5 (right). Each row corresponds to a fixed noise seed. We observe that the classifier drastically changes some images, while leaving others relatively unaffected.}
    \label{fig:varyingscale}
\end{figure}

\clearpage
\section{LSUN Diversity Comparison}

\begin{figure}[h]
    \begin{center}
    \begin{subfigure}[]{0.31\textwidth}
    \centerline{\includegraphics[width=\textwidth]{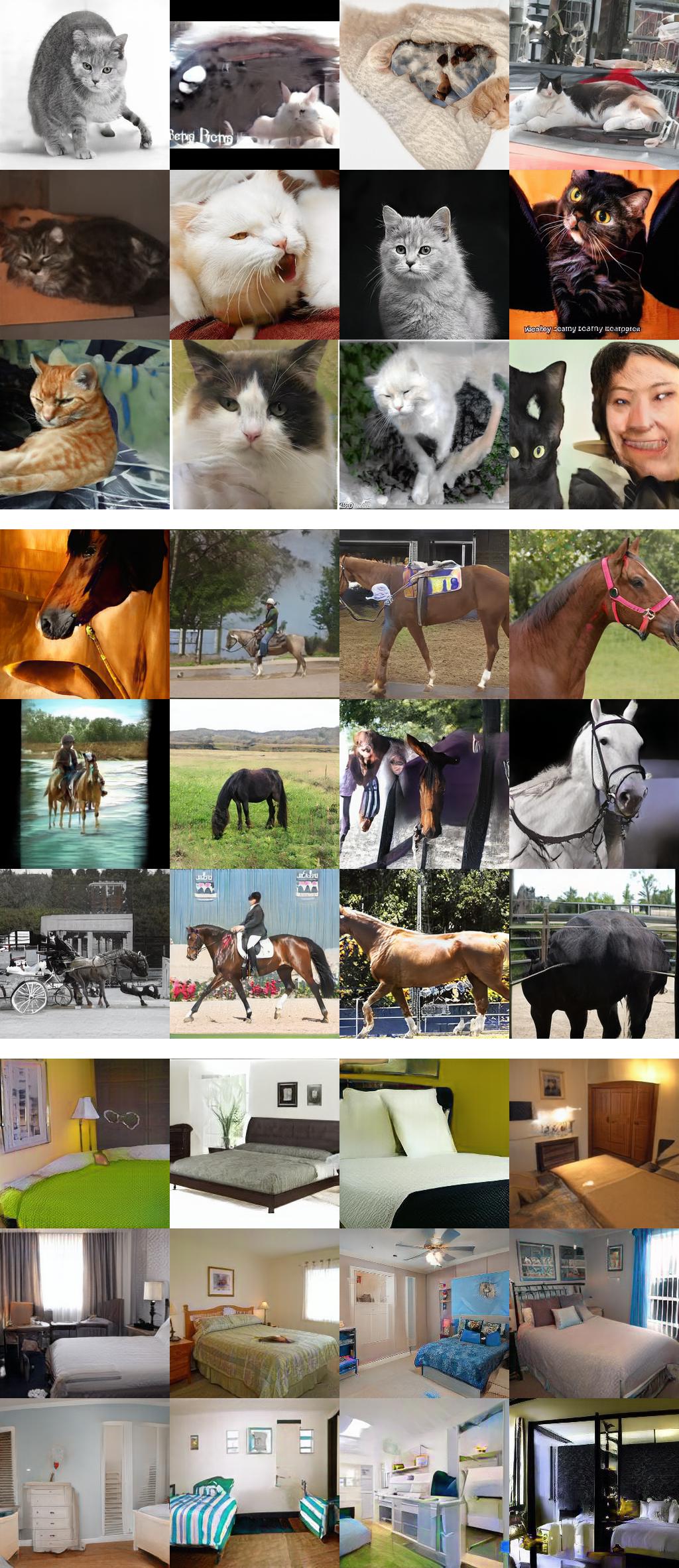}}
    \end{subfigure}\quad
    \begin{subfigure}[]{0.31\textwidth}
    \centerline{\includegraphics[width=\textwidth]{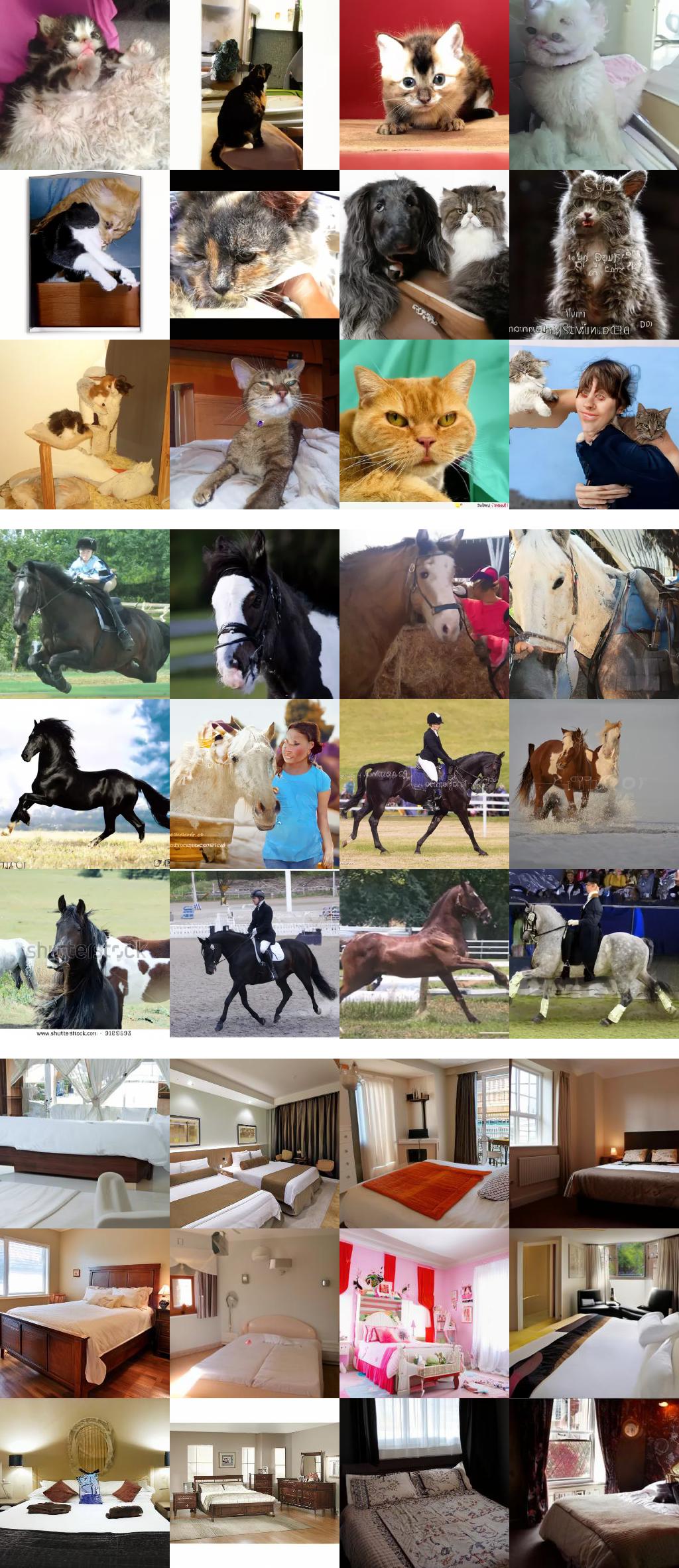}}
    \end{subfigure}\quad
    \begin{subfigure}[]{0.31\textwidth}
    \centerline{\includegraphics[width=\textwidth]{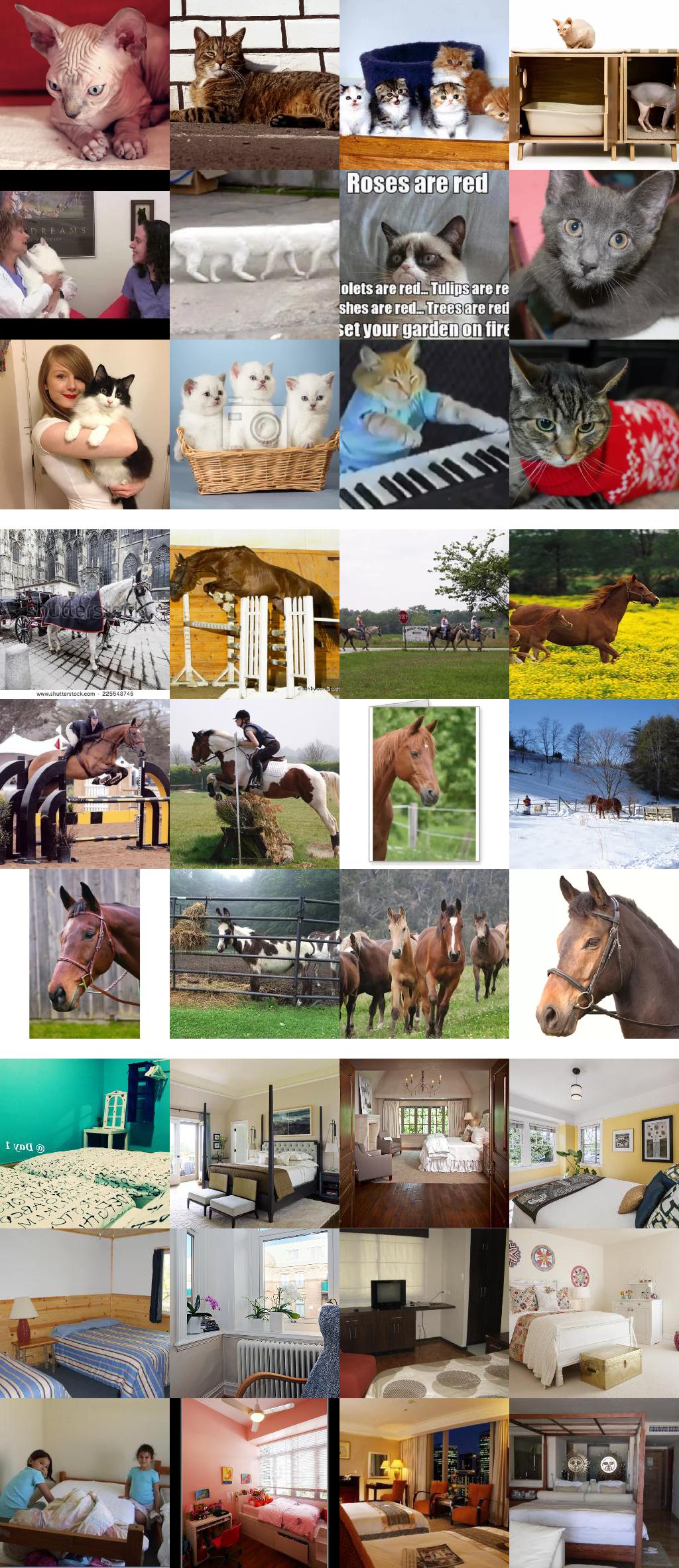}}
    \end{subfigure}
    \caption{\label{fig:diversity_lsun} Samples from StyleGAN2 (or StyleGAN for bedrooms) with truncation 1.0 (left) vs samples from our diffusion models (middle) and samples from the training set (right).}
    \end{center}
\end{figure}

\clearpage
\section{Interpolating Between Dataset Images Using DDIM}

The DDIM \shortcite{ddim} sampling process is deterministic given the initial noise $x_T$, thus giving rise to an implicit latent space. It corresponds to integrating an ODE in the forward direction, and we can run the process in reverse to get the latents that produce a given real image. Here, we experiment with encoding real images into this latent space and then interpolating between them.

Equation 13 for the generative pass in DDIM looks like
\begin{equation*}
x_{t-1} - x_{t} = \sqrt{\bar{\alpha}_{t-1}} \left[ \left(  \sqrt{1/\bar{\alpha}_{t}}  - \sqrt{1/\bar{\alpha}_{t-1}}\right) x_t + \left(\sqrt{1/\bar{\alpha}_{t-1} - 1} - \sqrt{1/\bar{\alpha}_{t} - 1} \right) \epsilon_\theta(x_t) \right]
\end{equation*}

Thus, in the limit of small steps, we can expect the reversal of this ODE in the forward direction looks like
\begin{equation*}
x_{t+1} - x_{t} = \sqrt{\bar{\alpha}_{t+1}} \left[ \left( \sqrt{1/\bar{\alpha}_{t}} - \sqrt{1/\bar{\alpha}_{t+1}}\right) x_t + \left(\sqrt{1/\bar{\alpha}_{t+1} - 1} - \sqrt{1/\bar{\alpha}_{t} - 1} \right) \epsilon_\theta(x_t) \right]
\end{equation*}

We found that this reverse ODE approximation gives latents with reasonable reconstructions, even with as few as 250 reverse steps. However, we noticed some noise artifacts when reversing all 250 steps, and find that reversing the first 249 steps gives much better reconstructions. To interpolate the latents, class embeddings, and classifier log probabilities, we use $cos(\theta) x_0 + sin(\theta) x_1$ where $\theta$ sweeps linearly from 0 to $\frac{\pi}{2}$.

Figures $\ref{fig:interpfirst}$ through $\ref{fig:interplast}$ show DDIM latent space interpolations on a class-conditional 256$\times$256 model, while varying the classifier scale. The left and rightmost images are ground truth dataset examples, and between them are reconstructed interpolations in DDIM latent space (including both endpoints). We see that the model with no guidance has almost perfect reconstructions due to its high recall, whereas raising the guidance scale to 2.5 only finds approximately similar reconstructions.

\begin{figure}[h]
    \begin{center}
    \begin{subfigure}{\textwidth}
        \centering
        \includegraphics[width=0.85\textwidth]{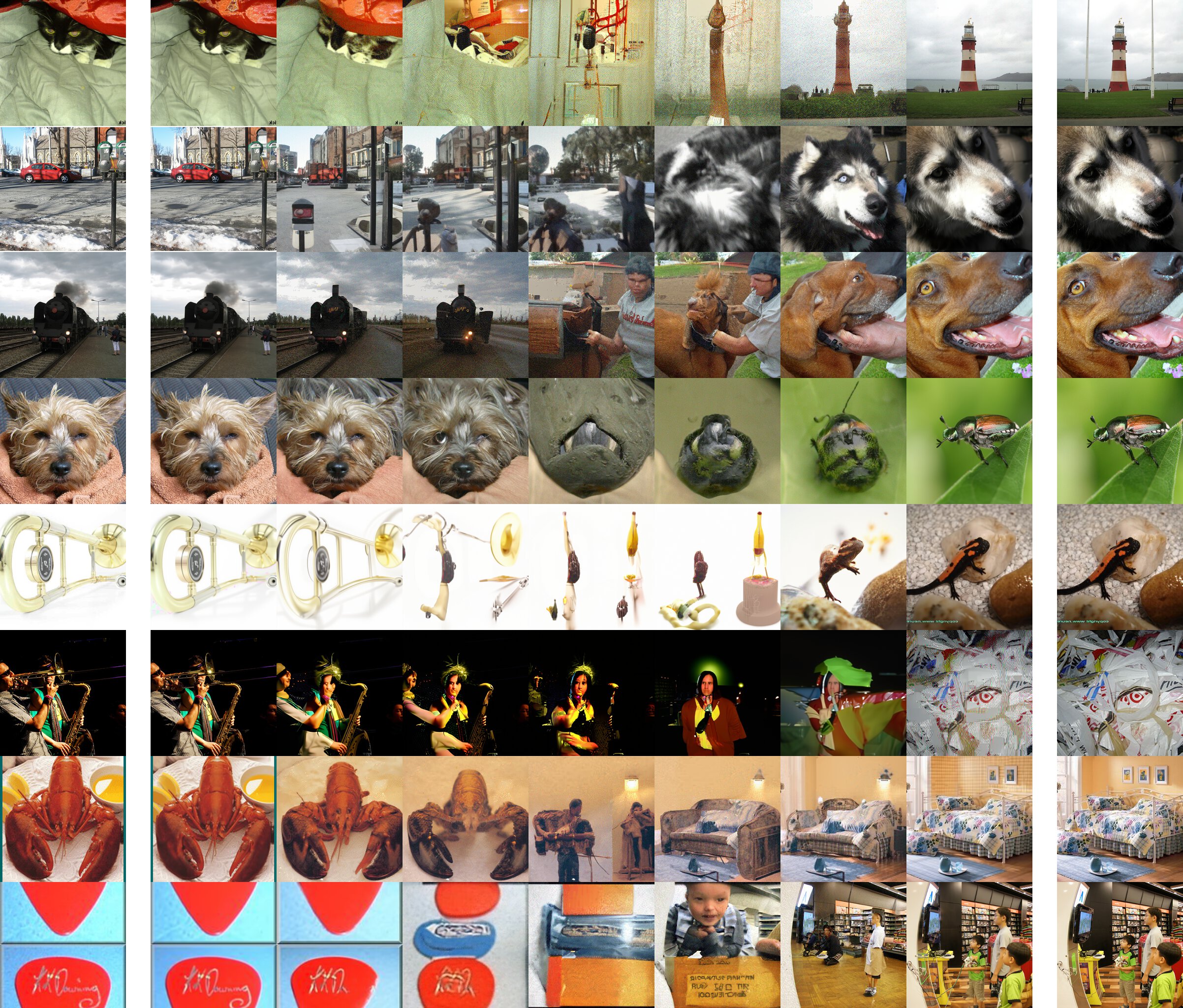}
        \caption{\label{fig:interpfirst} DDIM latent reconstructions and interpolations on real images with no classifier guidance.}
    \end{subfigure}
        \end{center}
\end{figure}
\clearpage   
\begin{figure}[tb]\ContinuedFloat
    \begin{center}
    \begin{subfigure}{\textwidth}
        \centering
        \includegraphics[width=0.85\textwidth]{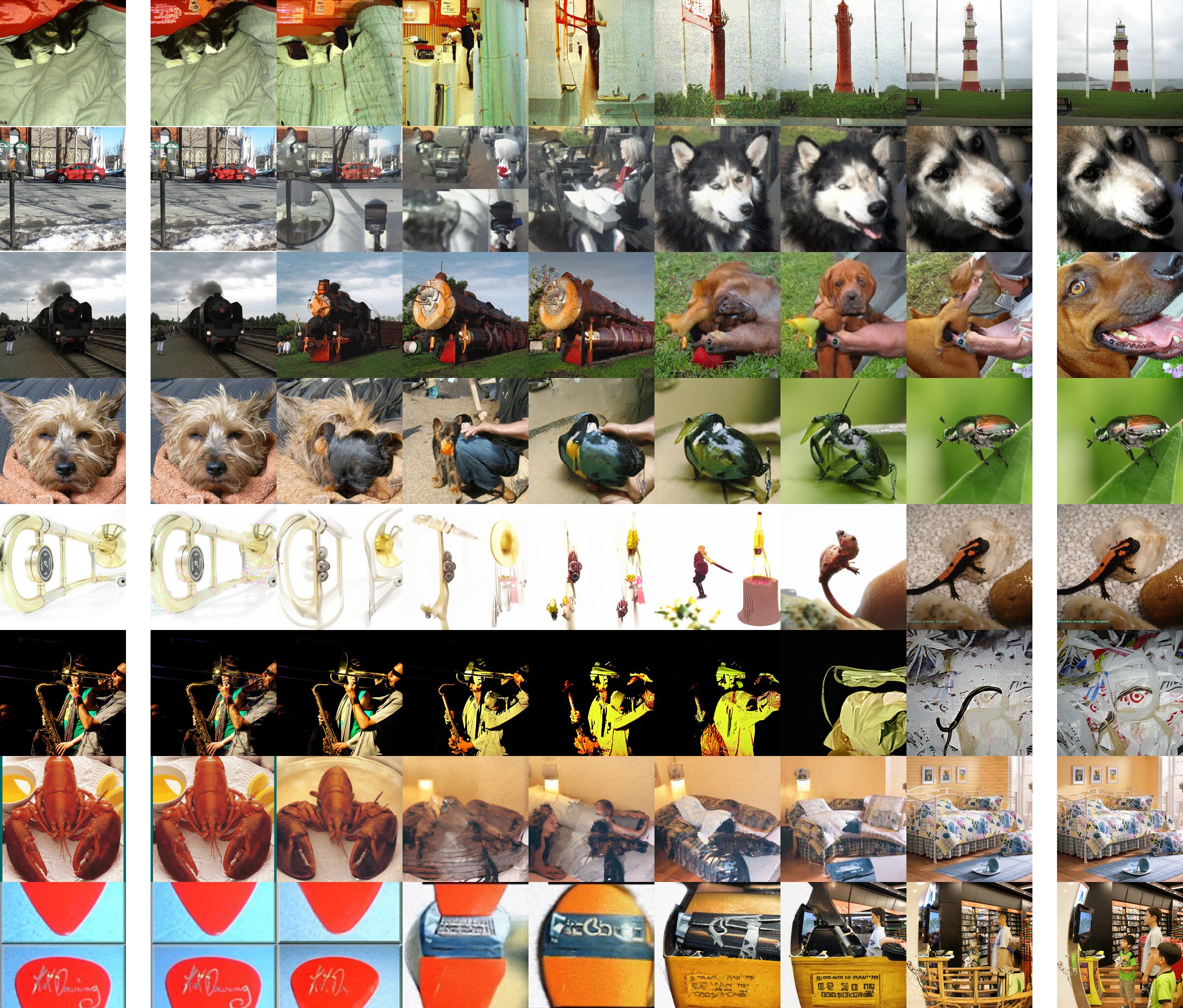}
        \caption{DDIM latent reconstructions and interpolations on real images with classifier scale 1.0.}
    \end{subfigure}
    \par\medskip
    \begin{subfigure}{\textwidth}
        \centering
        \includegraphics[width=0.85\textwidth]{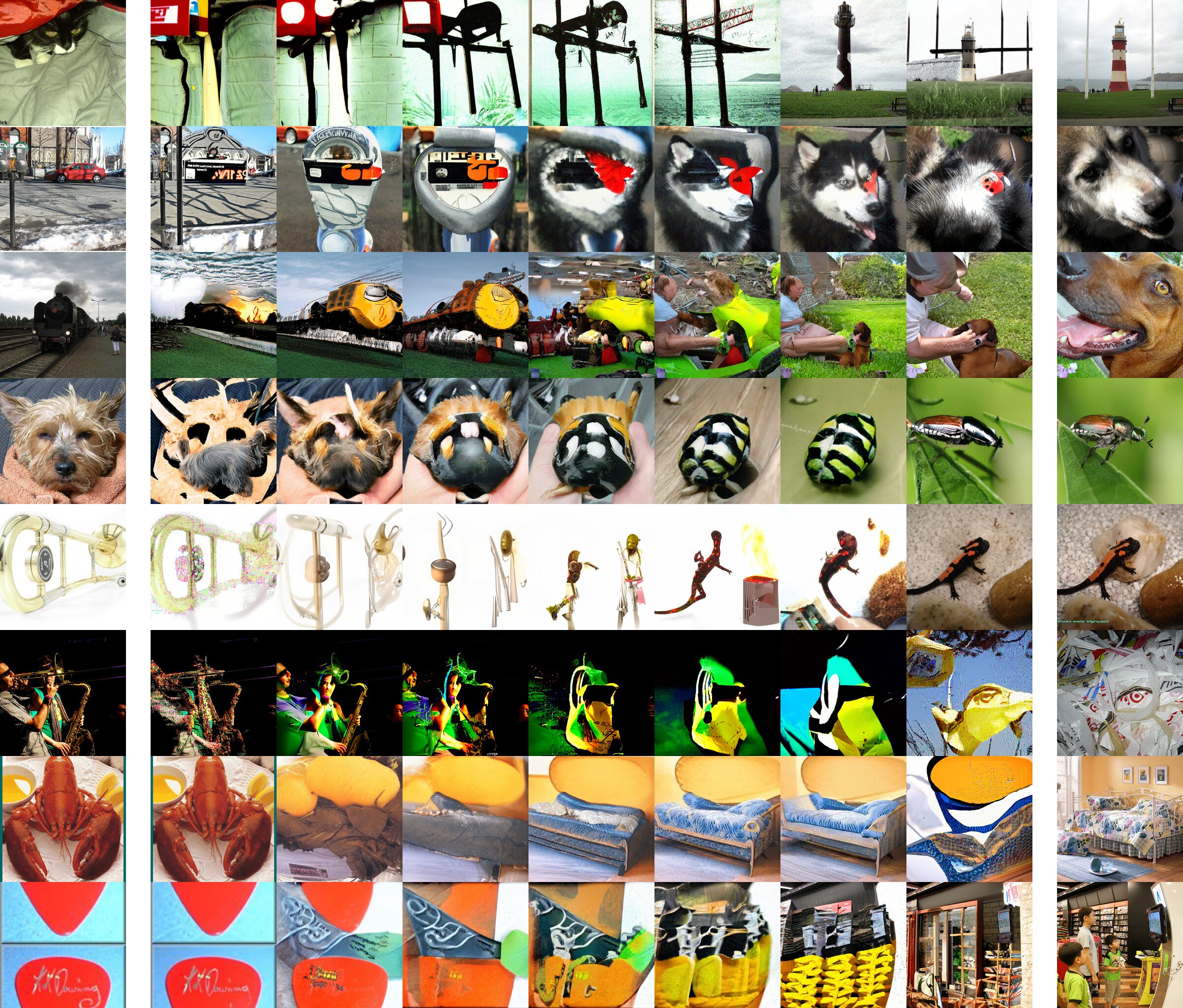}
        \caption{\label{fig:interplast} DDIM latent reconstructions and interpolations on real images with classifier scale 2.5.}
    \end{subfigure}
    \end{center}
\end{figure}

\clearpage
\section{Reduced Temperature Sampling}
\label{app:temperature}

We achieved our best ImageNet samples by reducing the diversity of our models using classifier guidance. For many classes of generative models, there is a much simpler way to reduce diversity: reducing the temperature \shortcite{temperature}. The temperature parameter $\tau$ is typically setup so that $\tau=1.0$ corresponds to standard sampling, and $\tau < 1.0$ focuses more on high-density samples. We experimented with two ways of implementing this for diffusion models: first, by scaling the Gaussian noise used for each transition by $\tau$, and second by dividing $\epsilon_{\theta}(x_t)$ by $\tau$. The latter implementation makes sense when thinking about $\epsilon$ as a re-scaled score function (see Section \ref{sec:ddimguide}), and scaling up the score function is similar to scaling up classifier gradients.

\begin{figure}[h]
    \begin{center}
    \begin{subfigure}{0.32\textwidth}
        \centerline{\includegraphics[width=\textwidth]{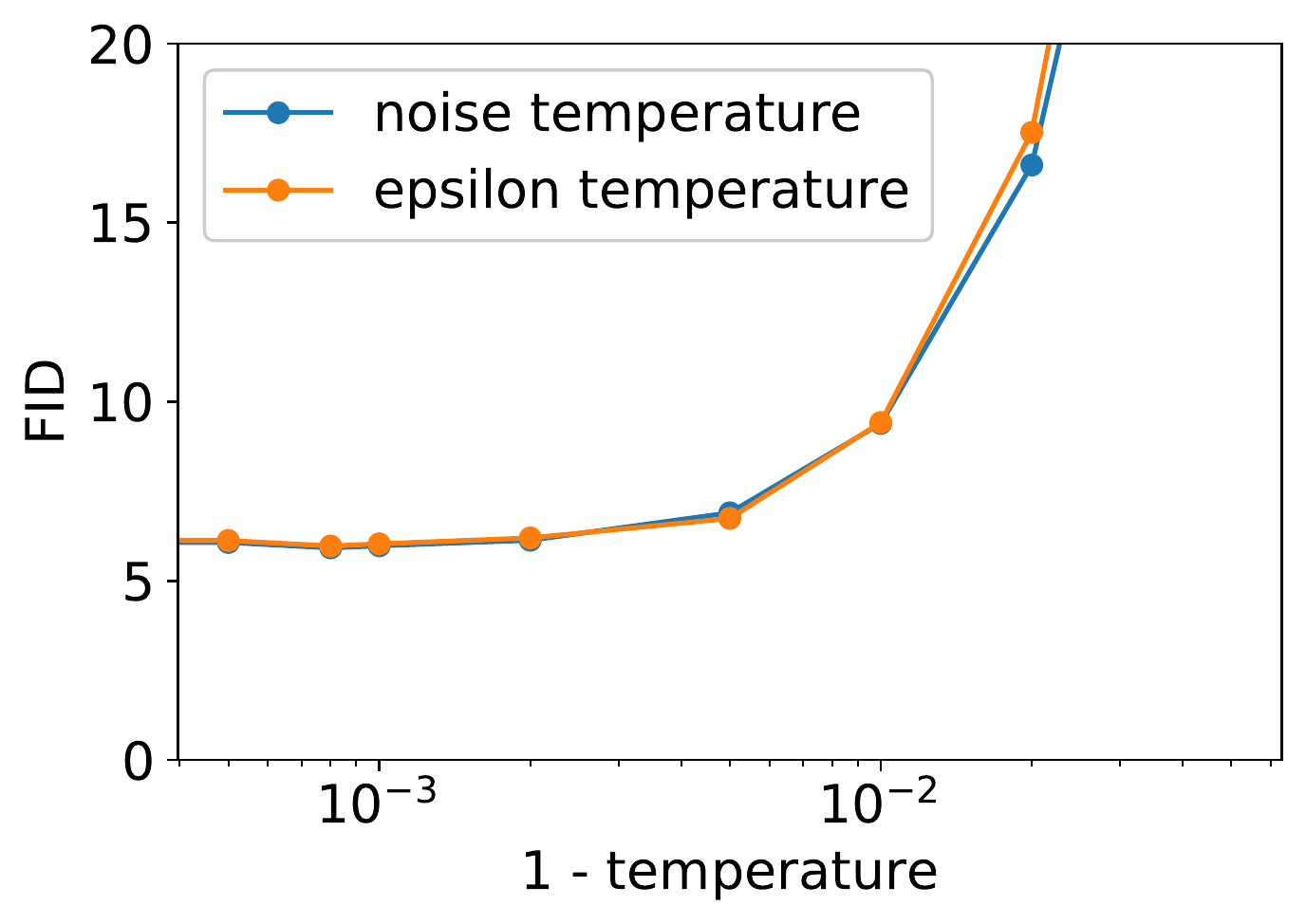}}
    \end{subfigure}
    \begin{subfigure}{0.32\textwidth}
        \centerline{\includegraphics[width=\textwidth]{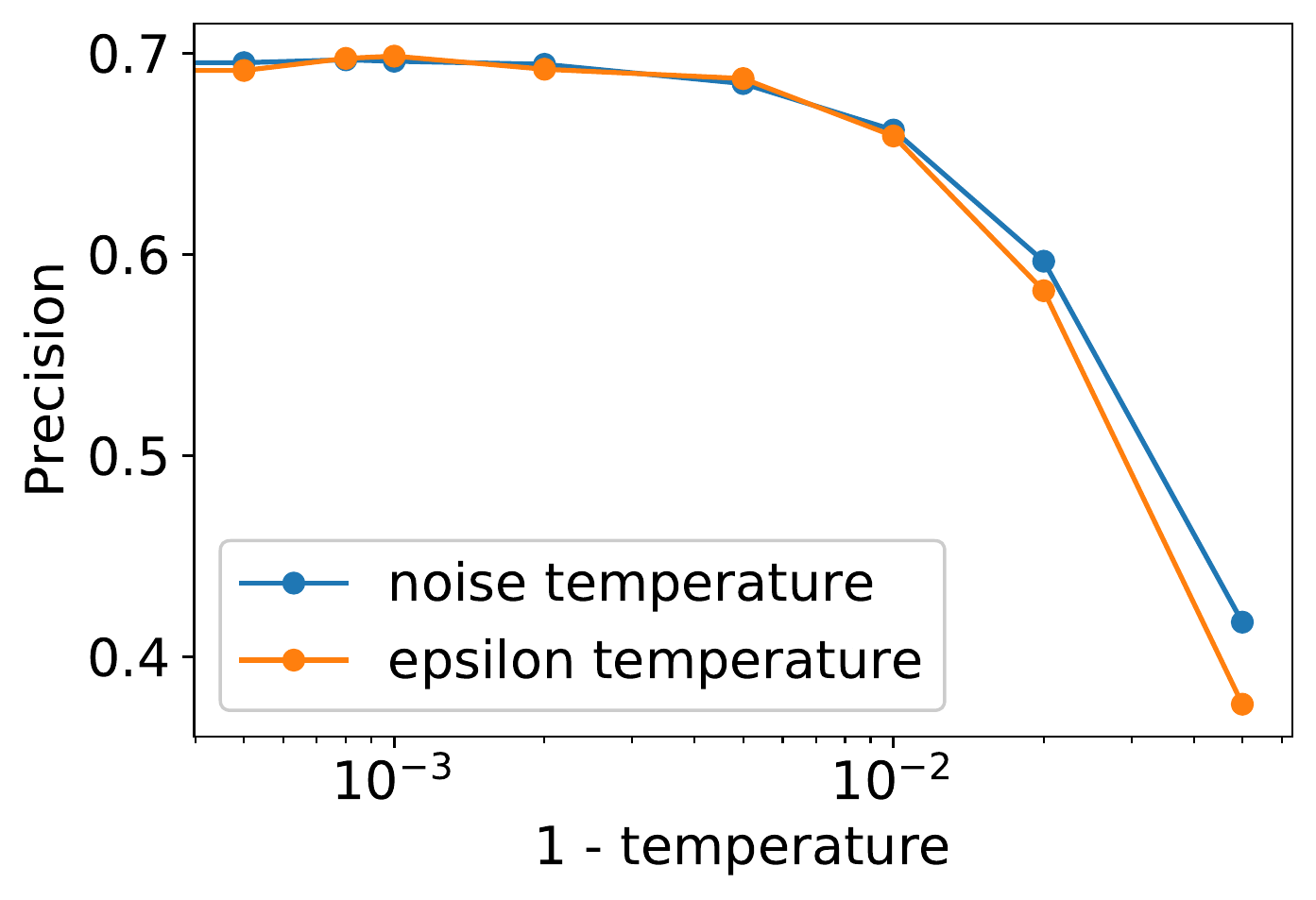}}
    \end{subfigure}
    \begin{subfigure}{0.32\textwidth}
        \centerline{\includegraphics[width=\textwidth]{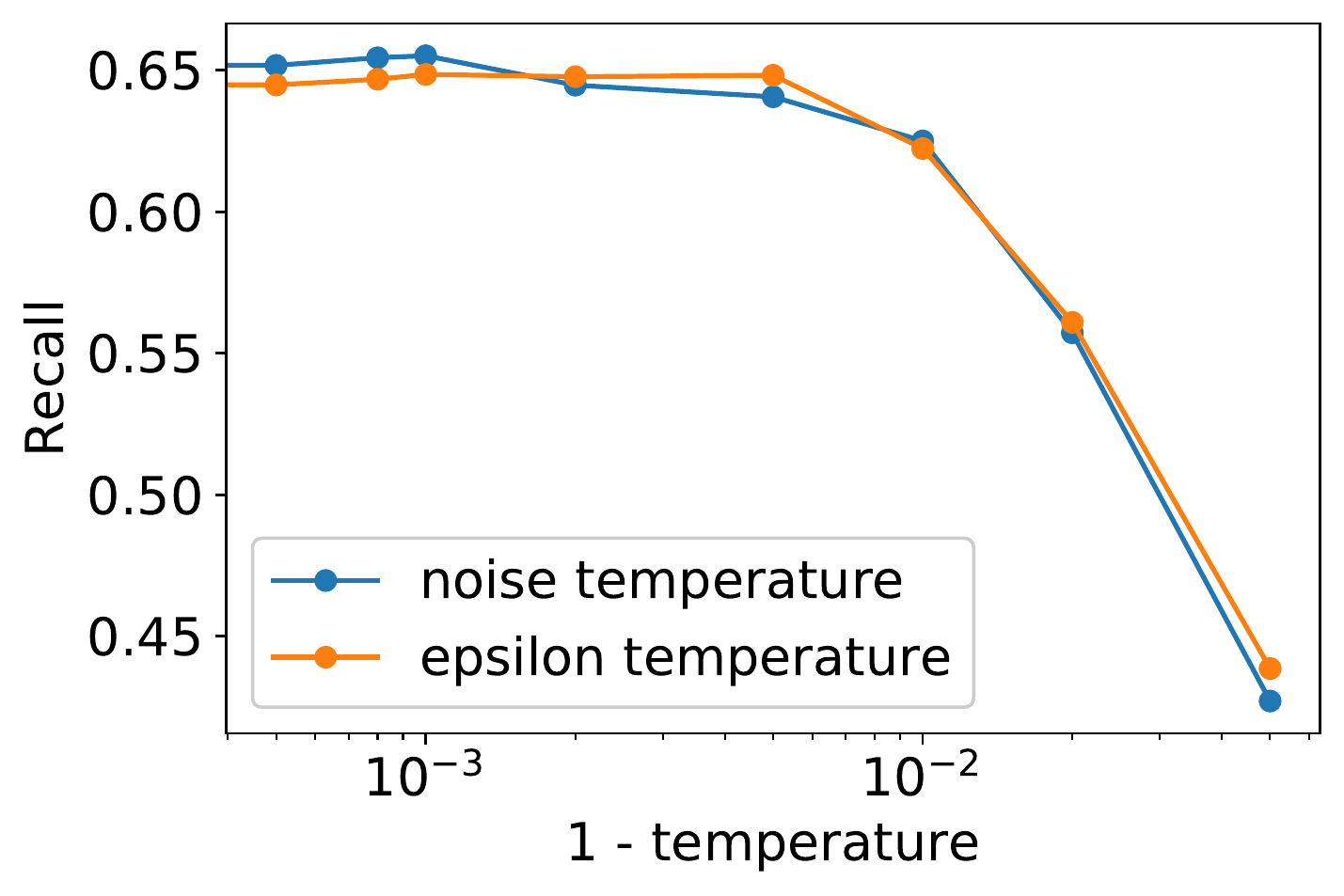}}
    \end{subfigure}
    \end{center}
    \caption{The effect of changing temperature for an ImageNet 128$\times$128 model.} 
    \label{fig:tempchanging}
\end{figure}

\begin{figure}[h]
    \begin{center}
    \begin{subfigure}{0.45\textwidth}
        \centerline{\includegraphics[width=\textwidth]{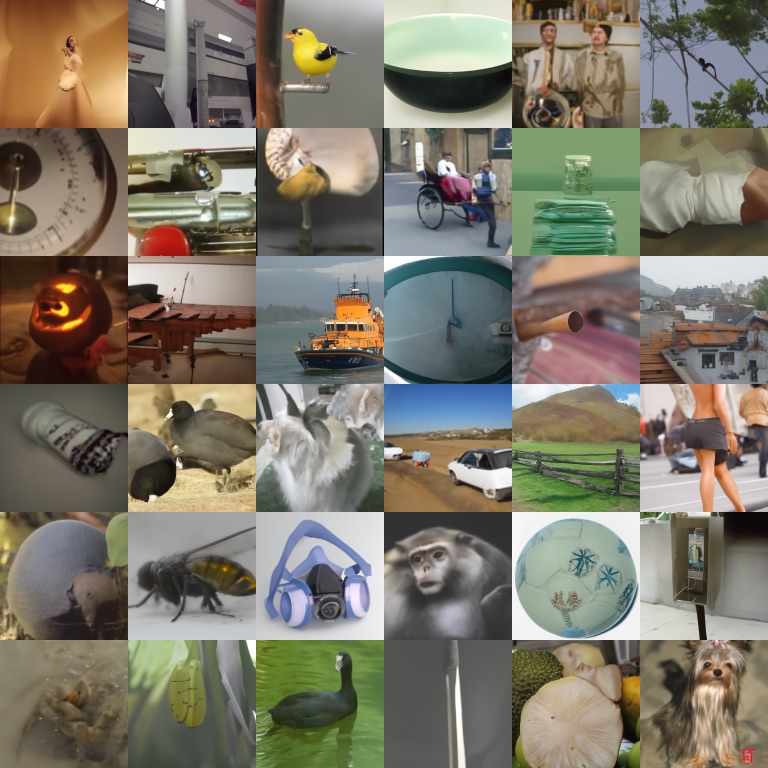}}
    \end{subfigure}
    \hspace{0.05\textwidth}
    \begin{subfigure}{0.45\textwidth}
        \centerline{\includegraphics[width=\textwidth]{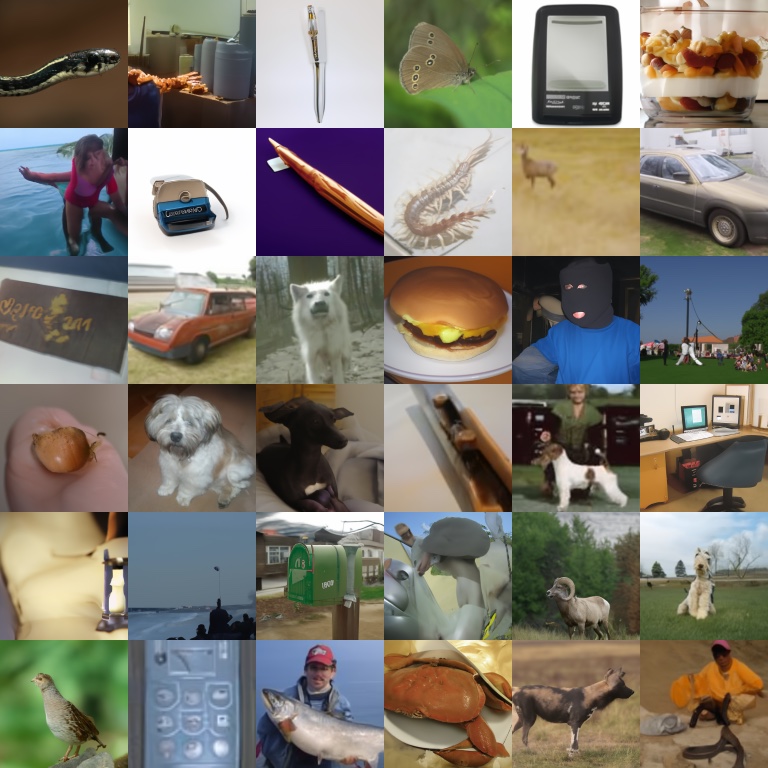}}
    \end{subfigure}
    \end{center}
    \caption{Samples at temperature 0.98 with epsilon scaling (left) and noise scaling (right).} 
    \label{fig:lowtempsamples}
\end{figure}

To measure how temperature scaling affects samples, we experimented with our ImageNet 128$\times$128 model, evaluating FID, Precision, and Recall across different temperatures (Figure \ref{fig:tempchanging}). We find that two techniques behave similarly, and neither technique provides any substantial improvement in our evaluation metrics. We also find that low temperatures have both low precision and low recall, indicating that the model is not focusing on modes of the real data distribution. Figure \ref{fig:lowtempsamples} highlights this effect, indicating that reducing temperature produces blurry, smooth images.

\clearpage
\section{Conditional Diffusion Process}
\label{app:conditional}

In this section, we show that conditional sampling can be achieved with a transition operator proportional to $p_{\theta}(x_t|x_{t+1}) p_{\phi}(y|x_t)$, where $p_{\theta}(x_t|x_{t+1})$ approximates $q(x_t|x_{t+1})$ and $p_{\phi}(y|x_t)$ approximates the label distribution for a noised sample $x_t$.

We start by defining a conditional Markovian noising process $\hat q$ similar to $q$, and assume that $\hat{q}(y|x_0)$ is a known and readily available label distribution for each sample.
\begin{alignat}{2}
\hat{q}(x_0) &\coloneqq q(x_0) \\
\hat{q}(y|x_0) &\coloneqq \text{Known labels per sample} \\
\hat{q}(x_{t+1}|x_t,y) &\coloneqq q(x_{t+1} | x_t) \label{eq:condnoise} \\
\hat{q}(x_{1:T}|x_0,y) &\coloneqq \prod_{t=1}^T \hat{q}(x_t|x_{t-1},y)
\end{alignat}

While we defined the noising process $\hat{q}$ conditioned on $y$, we can prove that $\hat{q}$ behaves exactly like $q$ when not conditioned on $y$. Along these lines, we first derive the unconditional noising operator $\hat{q}(x_{t+1}|x_t)$:
\begin{alignat}{2}
\hat{q}(x_{t+1}|x_t) &= \int_{y} \hat{q}(x_{t+1},y|x_t) \,dy \\
               &= \int_{y} \hat{q}(x_{t+1}|x_t,y) \hat{q}(y|x_t) \,dy \\
               &= \int_{y} q(x_{t+1}|x_t) \hat{q}(y|x_t) \,dy \\
               &= q(x_{t+1}|x_t) \int_{y} \hat{q}(y|x_t) \,dy \\
               &= q(x_{t+1}|x_t) \\
               &= \hat{q}(x_{t+1}|x_t,y)
\end{alignat}

Following similar logic, we find the joint distribution $\hat{q}(x_{1:T}|x_0)$:
\begin{alignat}{2}
\hat{q}(x_{1:T}|x_0) &= \int_{y} \hat{q}(x_{1:T},y|x_0) \,dy \\
&= \int_{y} \hat{q}(y|x_0) \hat{q}(x_{1:T}|x_0,y) \,dy \\
&= \int_{y} \hat{q}(y|x_0) \prod_{t=1}^T \hat{q}(x_t|x_{t-1},y) \,dy \\
&= \int_{y} \hat{q}(y|x_0) \prod_{t=1}^T q(x_t|x_{t-1}) \,dy \\
&= \prod_{t=1}^T q(x_t|x_{t-1}) \int_{y} \hat{q}(y|x_0) \,dy \\
&= \prod_{t=1}^T q(x_t|x_{t-1}) \\
&= q(x_{1:T}|x_0) \label{eq:condjointdist}
\end{alignat}

Using Equation \ref{eq:condjointdist}, we can now derive $\hat{q}(x_t)$:
\begin{alignat}{2}
\hat{q}(x_t) &= \int_{x_{0:t-1}} \hat{q}(x_0, ..., x_t) \,dx_{0:t-1} \\
&= \int_{x_{0:t-1}} \hat{q}(x_0) \hat{q}(x_1,...,x_t|x_0) \,dx_{0:t-1} \\
&= \int_{x_{0:t-1}} q(x_0) q(x_1,...,x_t|x_0) \,dx_{0:t-1} \\
&= \int_{x_{0:t-1}} q(x_0,...,x_t) \,dx_{0:t-1} \\
&= q(x_t) \\
\end{alignat}

Using the identities $\hat q(x_t) = q(x_t)$ and $\hat q(x_{t+1}|x_t) = q(x_{t+1}|x_t)$, it is trivial to show via Bayes rule that the unconditional reverse process $\hat{q}(x_t|x_{t+1}) = q(x_t|x_{t+1})$.

One observation about $\hat{q}$ is that it gives rise to a noisy classification function, $\hat{q}(y|x_t)$. We can show that this classification distribution does not depend on $x_{t+1}$ (a noisier version of $x_t$), a fact which we will later use:
\begin{alignat}{2}
\hat{q}(y|x_t,x_{t+1}) &= \hat{q}(x_{t+1}|x_t,y) \frac{\hat{q}(y|x_t)}{\hat{q}(x_{t+1}|x_t)} \\
                 &= \hat{q}(x_{t+1}|x_t) \frac{\hat{q}(y|x_t)}{\hat{q}(x_{t+1}|x_t)} \\
                 &= \hat{q}(y|x_t) \\
\end{alignat}

We can now derive the conditional reverse process:
\begin{alignat}{2}
\hat{q}(x_t|x_{t+1},y) &= \frac{\hat{q}(x_t,x_{t+1},y)}{\hat{q}(x_{t+1},y)} \\
                 &= \frac{\hat{q}(x_t,x_{t+1},y)}{\hat{q}(y|x_{t+1})\hat{q}(x_{t+1})} \\
                 &= \frac{\hat{q}(x_t|x_{t+1})\hat{q}(y|x_t,x_{t+1})\hat{q}(x_{t+1})}{\hat{q}(y|x_{t+1})\hat{q}(x_{t+1})} \\
                 &= \frac{\hat{q}(x_t|x_{t+1})\hat{q}(y|x_t,x_{t+1})}{\hat{q}(y|x_{t+1})} \\
                 &= \frac{\hat{q}(x_t|x_{t+1})\hat{q}(y|x_t)}{\hat{q}(y|x_{t+1})} \\
                 &= \frac{q(x_t|x_{t+1})\hat{q}(y|x_t)}{\hat{q}(y|x_{t+1})} \\
\end{alignat}

The $\hat{q}(y|x_{t+1})$ term can be treated as a constant since it does not depend on $x_t$. We thus want to sample from the distribution $Z q(x_t|x_{t+1})\hat{q}(y|x_t)$ where $Z$ is a normalizing constant. We already have a neural network approximation of $q(x_t|x_{t+1})$, called $p_{\theta}(x_t|x_{t+1})$, so all that is left is an approximation of $\hat{q}(y|x_t)$. This can be obtained by training a classifier $p_{\phi}(y|x_t)$ on noised images $x_t$ derived by sampling from $q(x_t)$.

\clearpage
\section{Hyperparameters}
\label{app:hyperparameters}

When choosing optimal classifier scales for our sampler, we swept over $[0.5, 1, 2]$ for ImageNet 128$\times$128 and ImageNet 256$\times$256, and $[1, 2, 3, 3.5, 4, 4.5, 5]$ for ImageNet 512$\times$512. For DDIM, we swept over values $[0.5, 0.75, 1.0, 1.25, 2]$ for ImageNet 128$\times$128, $[0.5, 1, 1.5, 2, 2.5, 3, 3.5]$ for ImageNet 256$\times$256, and $[3,4,5,6,7,9,11]$ for ImageNet 512$\times$512.

Hyperparameters for training the diffusion and classification models are in Table \ref{tab:hpsdiff} and Table \ref{tab:hpscf} respectively. Hyperparameters for guided sampling are in Table \ref{tab:hpguided}. Hyperparameters used to train upsampling models are in Table \ref{tab:hpsupsample}. We train all of our models using Adam \shortcite{adam} or AdamW \shortcite{adamw} with $\beta_1=0.9$ and $\beta_2=0.999$. We train in 16-bit precision using loss-scaling \shortcite{lossscaling}, but maintain 32-bit weights, EMA, and optimizer state. We use an EMA rate of 0.9999 for all experiments. We use PyTorch \shortcite{pytorch}, and train on NVIDIA Tesla V100s.

For all architecture ablations, we train with batch size 256, and sample using 250 sampling steps. For our attention heads ablations, we use 128 base channels, 2 residual blocks per resolution, multi-resolution attention, and BigGAN up/downsampling, and we train the models for 700K iterations. By default, all of our experiments use adaptive group normalization, except when explicitly ablating for it.

When sampling with 1000 timesteps, we use the same noise schedule as for training. On ImageNet, we use the uniform stride from \namecite{improved} for 250 step samples and the slightly different uniform stride from \namecite{ddim} for 25 step DDIM.

\begin{table}[h]
    \setlength\tabcolsep{4pt}
    \begin{center}
    \begin{small}
    \begin{tabular}{lccccc}
    \toprule
     & LSUN & ImageNet 64 & ImageNet 128 & ImageNet 256 & ImageNet 512\\
    \midrule
    Diffusion steps & 1000 & 1000 & 1000 & 1000 & 1000 \\
    Noise Schedule & linear & cosine & linear & linear & linear \\
    Model size & 552M & 296M & 422M & 554M & 559M \\
    Channels & 256 & 192 & 256 & 256 & 256  \\
    Depth & 2 & 3 & 2 & 2 & 2 \\
    Channels multiple & 1,1,2,2,4,4 & 1,2,3,4 & 1,1,2,3,4 & 1,1,2,2,4,4 & 0.5,1,1,2,2,4,4 \\
    Heads & & & 4 & & \\
    Heads Channels & 64 & 64 & & 64 & 64 \\
    Attention resolution & 32,16,8 & 32,16,8 & 32,16,8 & 32,16,8 & 32,16,8 \\
    BigGAN up/downsample & \cmark & \cmark & \cmark & \cmark & \cmark \\
    Dropout & 0.1 & 0.1 & 0.0 & 0.0 & 0.0 \\
    Batch size & 256 & 2048 & 256 & 256 & 256 \\
    Iterations & varies* & 540K & 4360K & 1980K & 1940K \\
    Learning Rate & 1e-4 & 3e-4 & 1e-4 & 1e-4 & 1e-4 \\
    \bottomrule
    \end{tabular}
    \end{small}
    \end{center}
    \caption{Hyperparameters for diffusion models. *We used 200K iterations for LSUN cat, 250K for LSUN horse, and 500K for LSUN bedroom.}
    \label{tab:hpsdiff}
     \vskip -0.2in
\end{table}

\begin{table}[h]
    \begin{center}
    \begin{small}
    \begin{tabular}{lccccc}
    \toprule
     & ImageNet 64 & ImageNet 128 & ImageNet 256 & ImageNet 512 \\
    \midrule
    Diffusion steps & 1000 & 1000 & 1000 & 1000 \\
    Noise Schedule & cosine & linear & linear & linear \\
    Model size & 65M & 43M & 54M & 54M \\
    Channels & 128 & 128 & 128 & 128 \\
    Depth & 4 & 2 & 2 & 2 \\
    Channels multiple & 1,2,3,4 & 1,1,2,3,4 & 1,1,2,2,4,4 & 0.5,1,1,2,2,4,4 \\
    Heads Channels & 64 & 64 & 64 & 64 \\
    Attention resolution & 32,16,8 & 32,16,8 & 32,16,8 & 32,16,8 \\
    BigGAN up/downsample & \cmark & \cmark & \cmark & \cmark \\
    Attention pooling & \cmark & \cmark & \cmark & \cmark \\
    Weight decay & 0.2 & 0.05 & 0.05 & 0.05 \\
    Batch size & 1024 & 256* & 256 & 256 \\
    Iterations & 300K & 300K & 500K & 500K \\
    Learning rate & 6e-4 & 3e-4* & 3e-4 & 3e-4 \\
    \bottomrule
    \end{tabular}
    \end{small}
    \end{center}
    \caption{Hyperparameters for classification models. *For our ImageNet 128$\times$128 $\to$ 512$\times$512 upsamples, we used a different classifier for the base model, with batch size 1024 and learning rate 6e-5.}
    \label{tab:hpscf}
    \vskip -0.2in
\end{table}

\begin{table}[h]
    \setlength\tabcolsep{4pt}
    \begin{center}
    \begin{small}
    \begin{tabular}{lccccc}
    \toprule
     & ImageNet $64 \rightarrow 256$ & ImageNet $128 \rightarrow 512$\\
    \midrule
    Diffusion steps & 1000 & 1000 \\
    Noise Schedule & linear & linear \\
    Model size & 312M & 309M & \\
    Channels & 192 & 192 \\
    Depth & 2 & 2 \\
    Channels multiple & 1,1,2,2,4,4 & 1,1,2,2,4,4* \\
    Heads & 4 & \\
    Heads Channels & & 64 \\
    Attention resolution & 32,16,8 & 32,16,8 \\
    BigGAN up/downsample & \cmark & \cmark \\
    Dropout & 0.0 & 0.0 \\
    Batch size & 256 & 256 \\
    Iterations & 500K & 1050K \\
    Learning Rate & 1e-4 & 1e-4 \\
    \bottomrule
    \end{tabular}
    \end{small}
    \end{center}
    \caption{Hyperparameters for upsampling diffusion models. *We chose this as an optimization, with the intuition that a lower-resolution path should be unnecessary for upsampling 128x128 images.}
    \label{tab:hpsupsample}
    \vskip -0.2in
\end{table}

\begin{table}[h]
    \begin{center}
    \begin{small}
    \begin{tabular}{lccccc}
    \toprule
    & ImageNet 64 & ImageNet 128 & ImageNet 256 & ImageNet 512 \\
    \midrule
    Gradient Scale (250 steps) & 1.0 & 0.5 & 1.0 & 4.0 \\
    Gradient Scale (DDIM, 25 steps) & - & 1.25 & 2.5 & 9.0 \\
    \bottomrule
    \end{tabular}
    \end{small}
    \end{center}
    \caption{Hyperparameters for classifier-guided sampling.}
    \label{tab:hpguided}
    \vskip -0.2in
\end{table}

\clearpage
\section{Using Fewer Sampling Steps on LSUN}
\label{app:lsunsweep}

We initially found that our LSUN models achieved much better results when sampling with 1000 steps rather than 250 steps, contrary to previous results from \namecite{improved}. To address this, we conducted a sweep over sampling-time noise schedules, finding that an improved schedule can largely close the gap. We swept over schedules on LSUN bedrooms, and selected the schedule with the best FID for use on the other two datasets. Table \ref{tab:schedulesweep} details the findings of this sweep, and Table \ref{tab:schedulesweep_all} applies this schedule to three LSUN datasets.

While sweeping over sampling schedules is not as expensive as re-training models from scratch, it does require a significant amount of sampling compute. As a result, we did not conduct an exhaustive sweep, and superior schedules are likely to exist.

\begin{table}[h]
    \begin{center}
    \begin{small}
    \begin{tabular}{lc}
    \toprule
    Schedule & FID \\
    \midrule
    $50,50,50,50,50$ & 2.31 \\
    $70,60,50,40,30$ & 2.17 \\
    $90,50,40,40,30$ & 2.10 \\
    $90,60,50,30,20$ & 2.09 \\
    $80,60,50,30,30$ & 2.09 \\
    $90,50,50,30,30$ & 2.07 \\
    $100,50,40,30,30$ & 2.03 \\
    $90,60,60,20,20$ & \bf 2.02 \\
    \bottomrule
    \end{tabular}
    \end{small}
    \end{center}
    \caption{Results of sweeping over 250 step sampling schedules on LSUN bedrooms. The schedule is expressed as a sequence of five integers, where each integer is the number of steps allocated to one fifth of the diffusion process. The first integer corresponding to $t \in [0,199]$ and the last to $t \in [T-200,T-1]$. Thus, $50,50,50,50,50$ is a uniform schedule, and $250,0,0,0,0$ is a schedule where all timesteps are spent near $t=0$.}
    \label{tab:schedulesweep}
\end{table}

\begin{table}[h]
    \begin{center}
    \begin{small}
    \begin{tabular}[t]{lcccc}
    \toprule
    Schedule & FID & sFID & Prec & Rec \\
    \\
    \multicolumn{5}{l}{\bf{LSUN Bedrooms} 256$\times$256} \\
    \toprule
    1000 steps & 1.90 & 5.59 & 0.66 & 0.51  \\
    250 steps (uniform) & 2.31 & 6.12 & 0.65 & 0.50 \\
    250 steps (sweep) & 2.02 & 6.12 & 0.67 & 0.50 \\
    \\
    \multicolumn{5}{l}{\bf{LSUN Horses} 256$\times$256} \\
    1000 steps & 2.57 & 6.81 & 0.71 & 0.55 \\
    250 steps (uniform) & 3.45 & 7.55 & 0.68 & 0.56 \\
    250 steps (sweep) & 2.83 & 7.08 & 0.69 & 0.56 \\
    \\
    \multicolumn{5}{l}{\bf{LSUN Cat} 256$\times$256} \\
    1000 steps & 5.57 & 6.69 & 0.63 & 0.52 \\
    250 steps (uniform) & 7.03 & 8.24 & 0.60 & 0.53 \\
    250 steps (sweep) & 5.94 & 7.43 & 0.62 & 0.52 \\
    \bottomrule
    \end{tabular}
    \end{small}
    \end{center}
    \caption{Evaluations on LSUN bedrooms, horses, and cats using different sampling schedules. We find that the sweep schedule produces better results than the uniform 250 step schedule on all three datasets, and mostly bridges the gap to the 1000 step schedule.}
    \label{tab:schedulesweep_all}
\end{table}

\clearpage
\section{Samples from ImageNet \texorpdfstring{512$\times$512}{512x512}}
\label{app:fullsamples}

\begin{figure}[h]
    \centerline{\includegraphics[width=\textwidth]{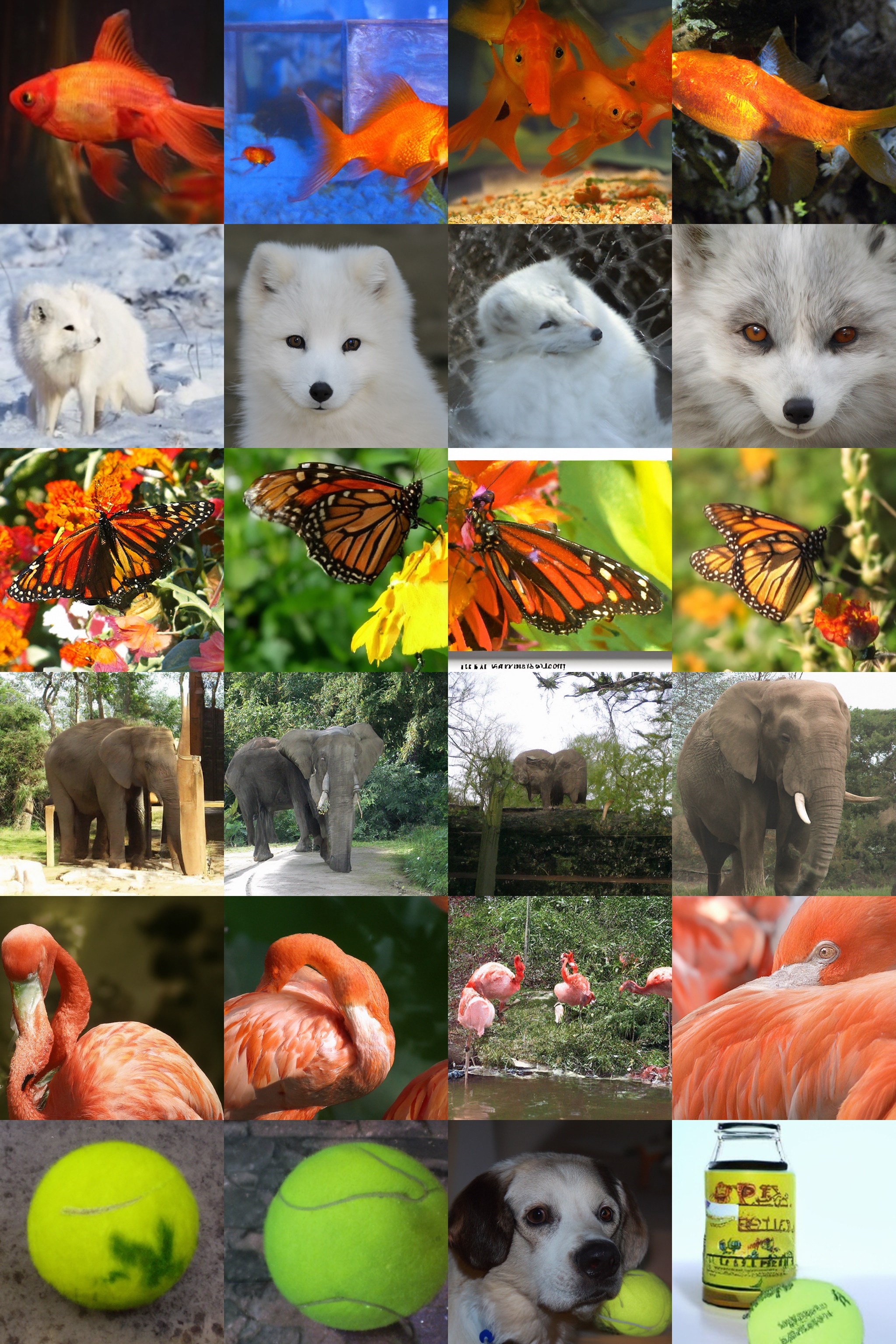}}
    \caption{Samples from our best 512$\times$512 model (FID: 3.85). Classes are 1: goldfish, 279: arctic fox, 323: monarch butterfly, 386: african elephant, 130: flamingo, 852: tennis ball.}
    \vskip -1in
\end{figure}
\clearpage
\begin{figure}[h]
    \centerline{\includegraphics[width=\textwidth]{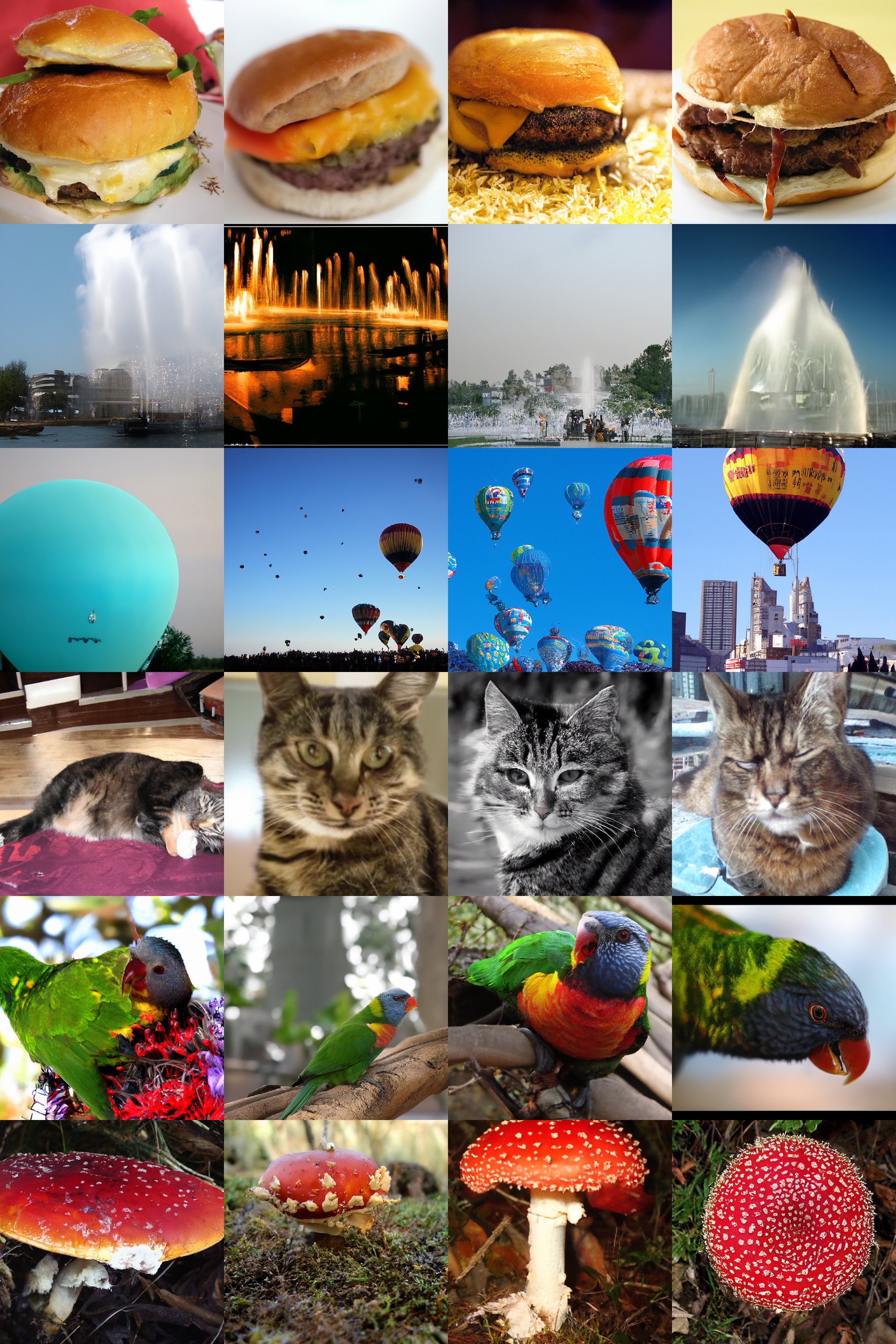}}
    \caption{Samples from our best 512$\times$512 model (FID: 3.85). Classes are 933: cheeseburger, 562: fountain, 417: balloon, 281: tabby cat, 90: lorikeet, 992: agaric.}
    \vskip -1in
\end{figure}

\clearpage
\begin{figure}[h]
    \centerline{\includegraphics[width=\textwidth]{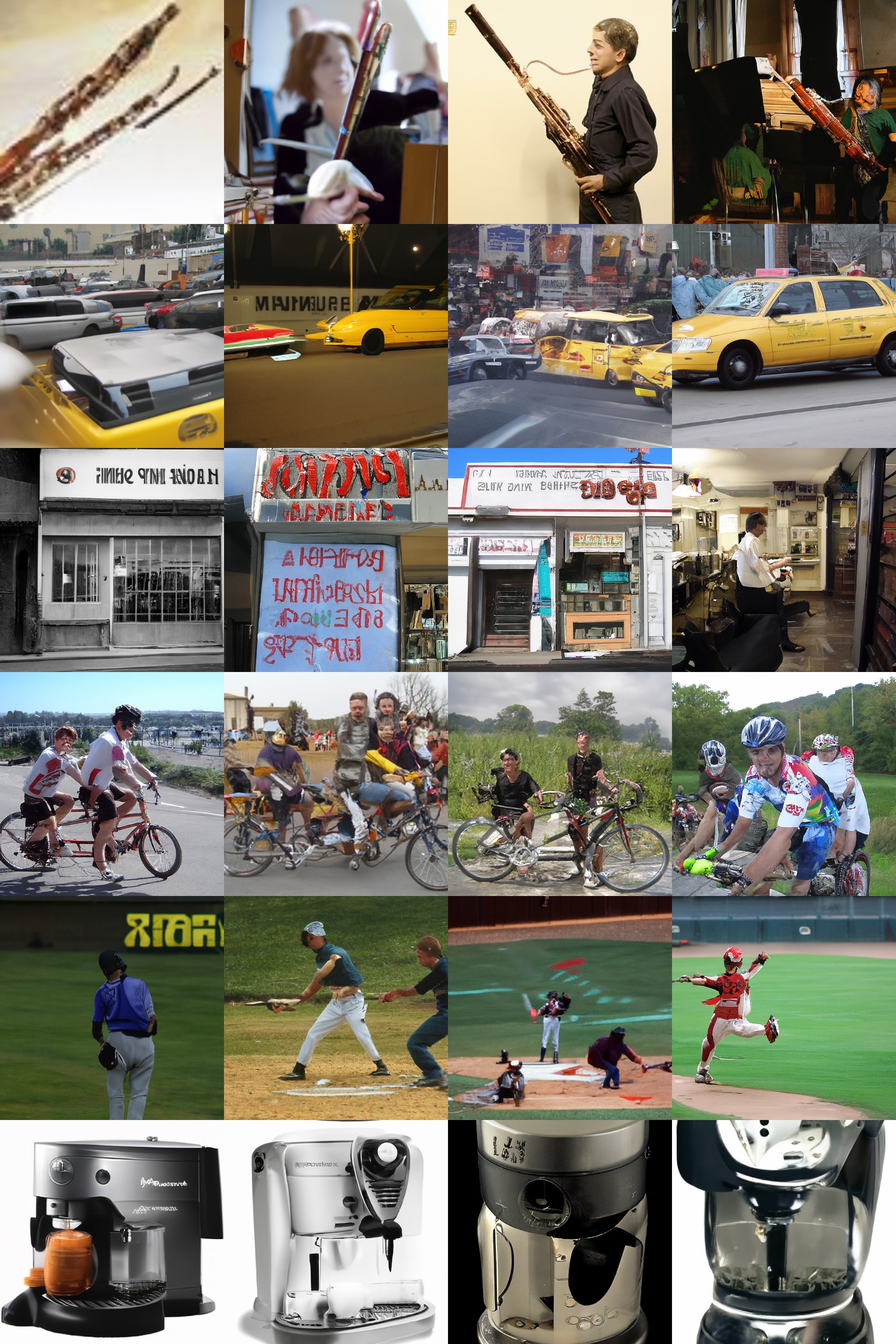}}
    \caption{Difficult class samples from our best 512$\times$512 model (FID: 3.85). Classes are 432: bassoon, 468: cab, 424: barbershop, 444: bicycle-built-for-two, 981: ballplayer, 550: espresso maker.}
    \vskip -1in
\end{figure}

\clearpage

\begin{figure}[h]
    \centerline{\includegraphics[width=\textwidth]{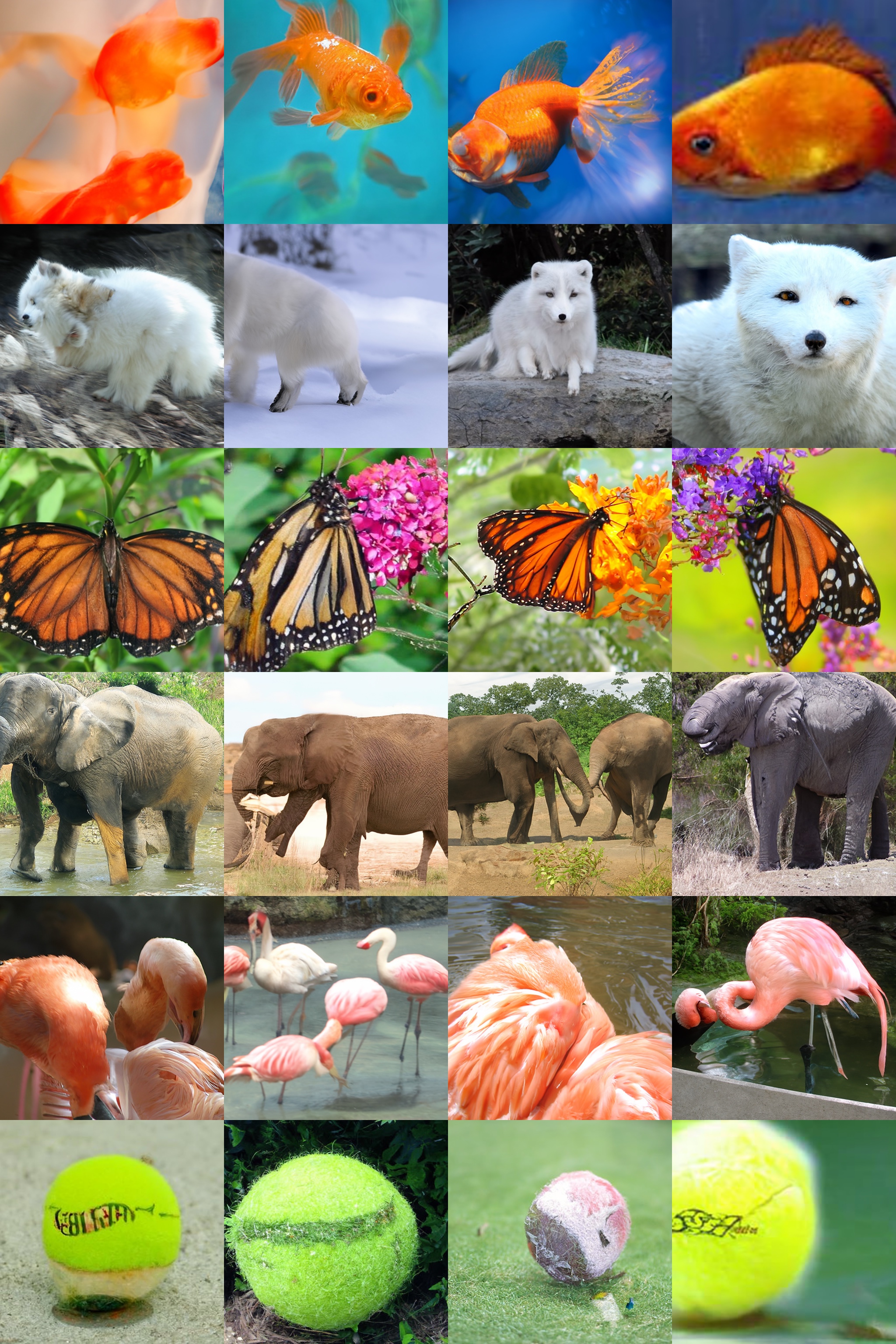}}
    \caption{Samples from our guided 512$\times$512 model using 250 steps with classifier scale 4.0 (FID 7.72). Classes are 1: goldfish, 279: arctic fox, 323: monarch butterfly, 386: african elephant, 130: flamingo, 852: tennis ball.}
\end{figure}
\clearpage
\begin{figure}[h]
    \centerline{\includegraphics[width=\textwidth]{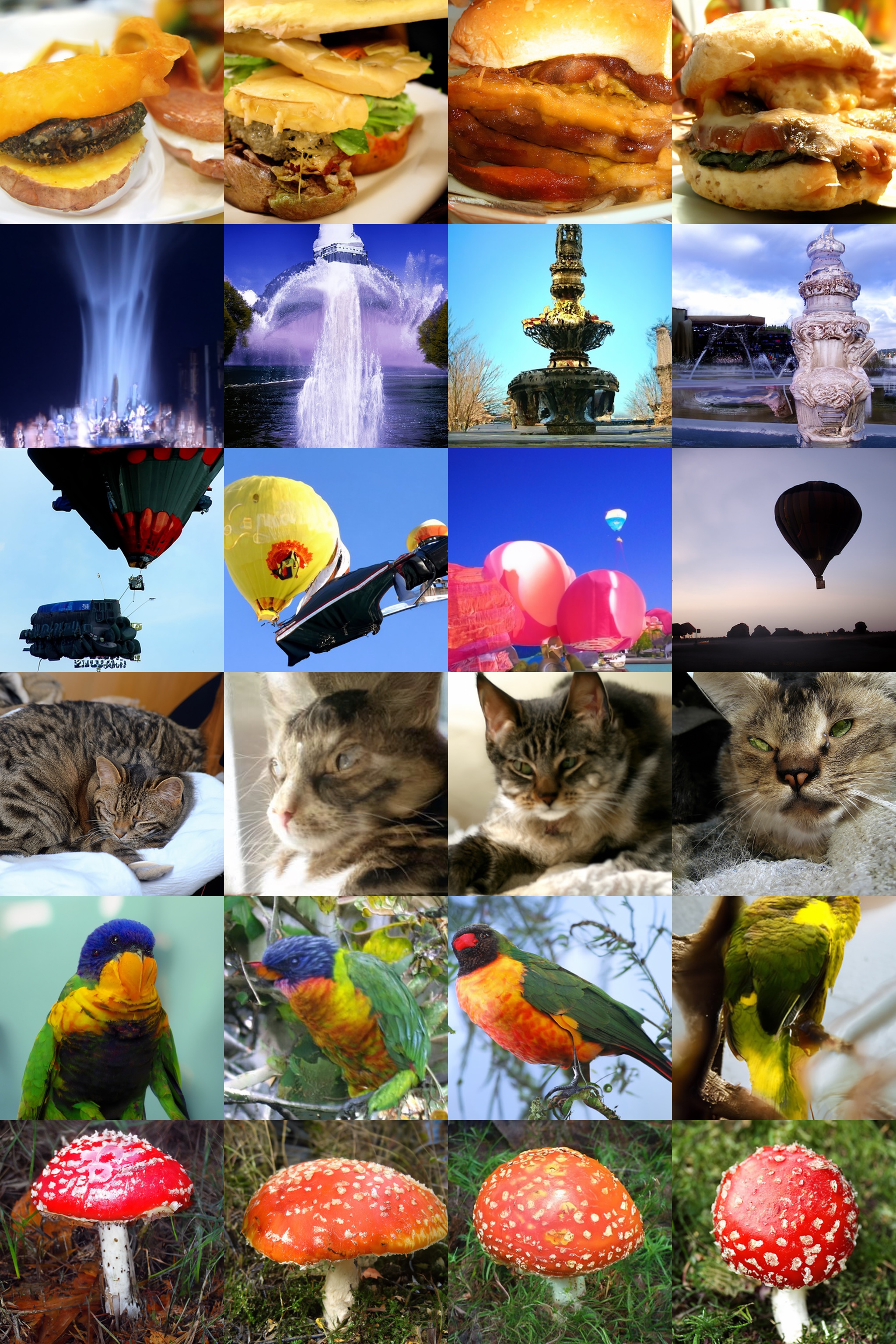}}
    \caption{Samples from our guided 512$\times$512 model using 250 steps with classifier scale 4.0 (FID 7.72). Classes are 933: cheeseburger, 562: fountain, 417: balloon, 281: tabby cat, 90: lorikeet, 992: agaric.}
\end{figure}

\clearpage

\begin{figure}[tb]
    \centerline{\includegraphics[width=\textwidth]{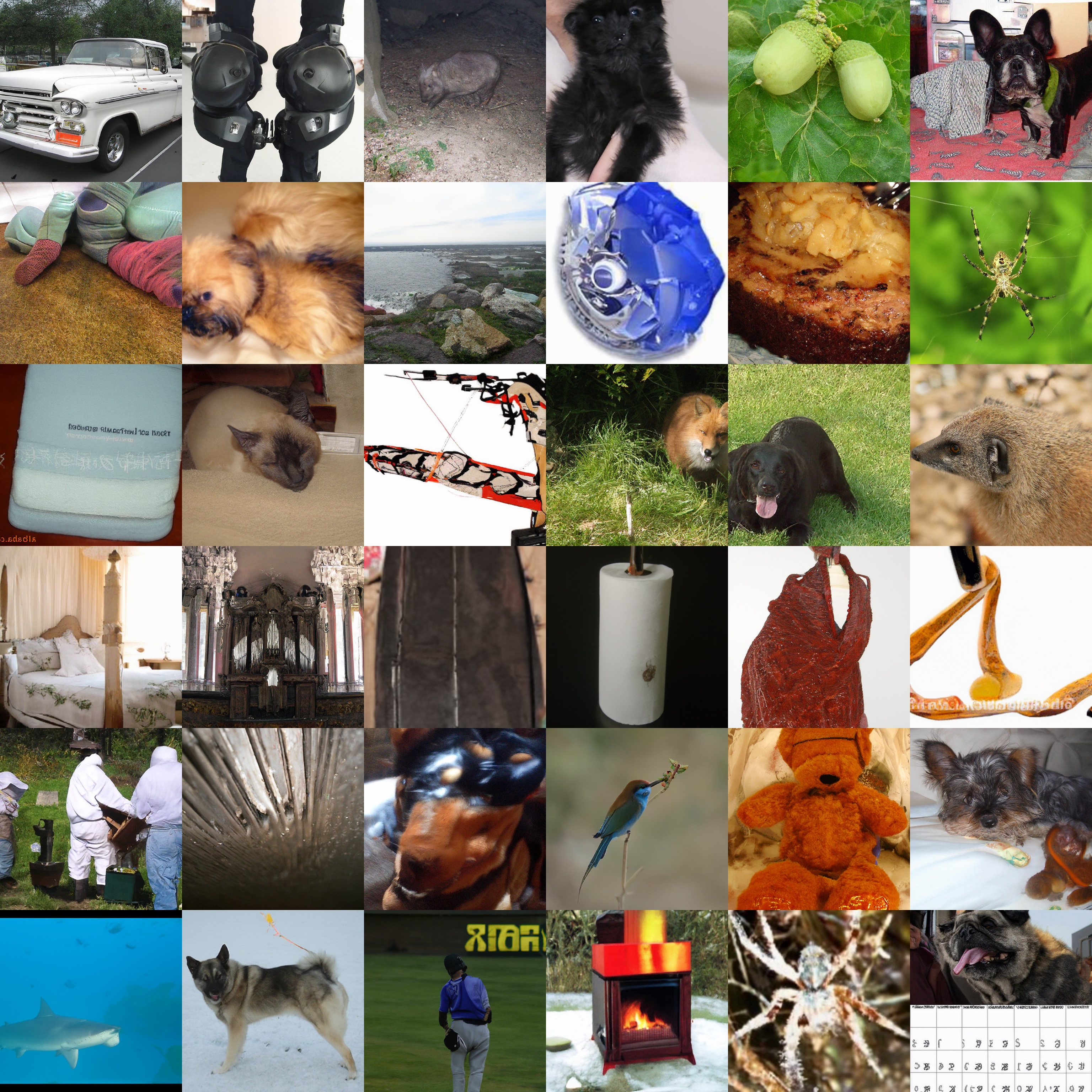}}
    \caption{Random samples from our best ImageNet 512$\times$512 model (FID 3.85).}
    \vskip -1in
\end{figure}

\begin{figure}[tb]
    \centerline{\includegraphics[width=\textwidth]{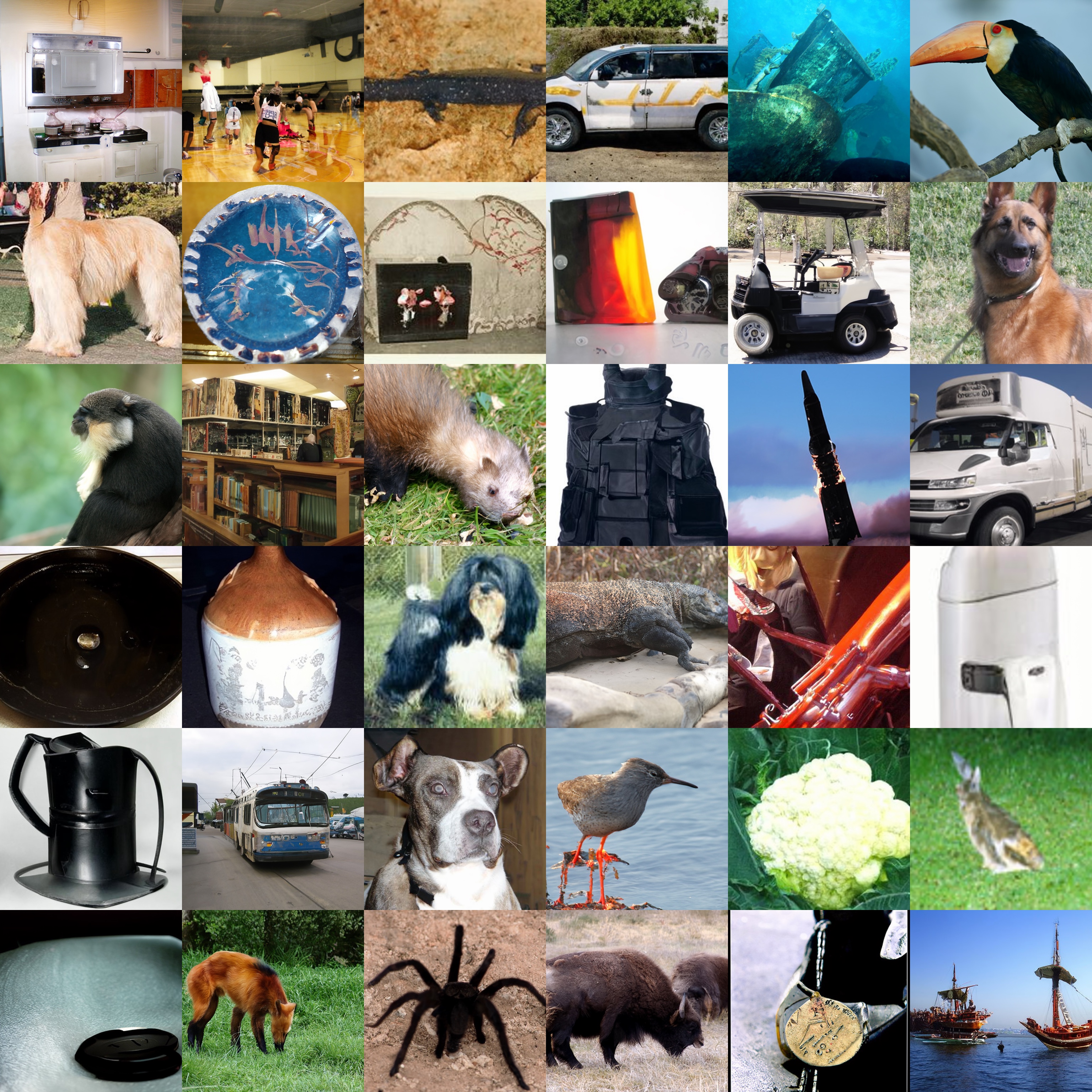}}
    \caption{Random samples from our guided 512$\times$512 model using 250 steps with classifier scale 4.0 (FID 7.72).}
    \vskip -1in
\end{figure}

\clearpage
\section{Samples from ImageNet \texorpdfstring{256$\times$256}{256x256}}
\begin{figure}[h]
    \centerline{\includegraphics[width=\textwidth]{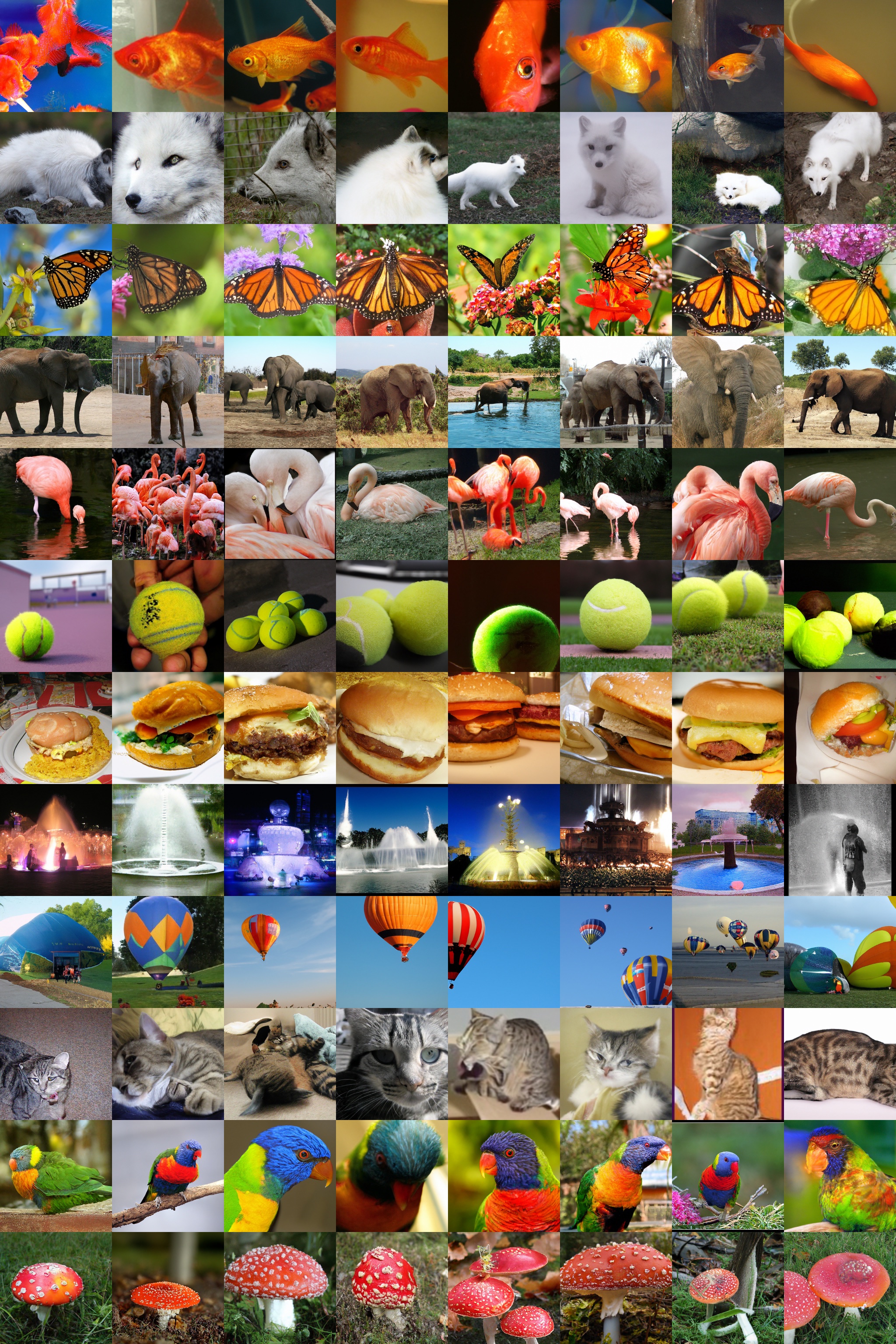}}
    \caption{Samples using our best 256$\times$256 model (FID 3.94). Classes are 1: goldfish, 279: arctic fox, 323: monarch butterfly, 386: african elephant, 130: flamingo, 852: tennis ball, 933: cheeseburger, 562: fountain, 417: balloon, 281: tabby cat, 90: lorikeet, 992: agaric}
    \vskip -1in
\end{figure}
\clearpage
\begin{figure}[h]
    \centerline{\includegraphics[width=\textwidth]{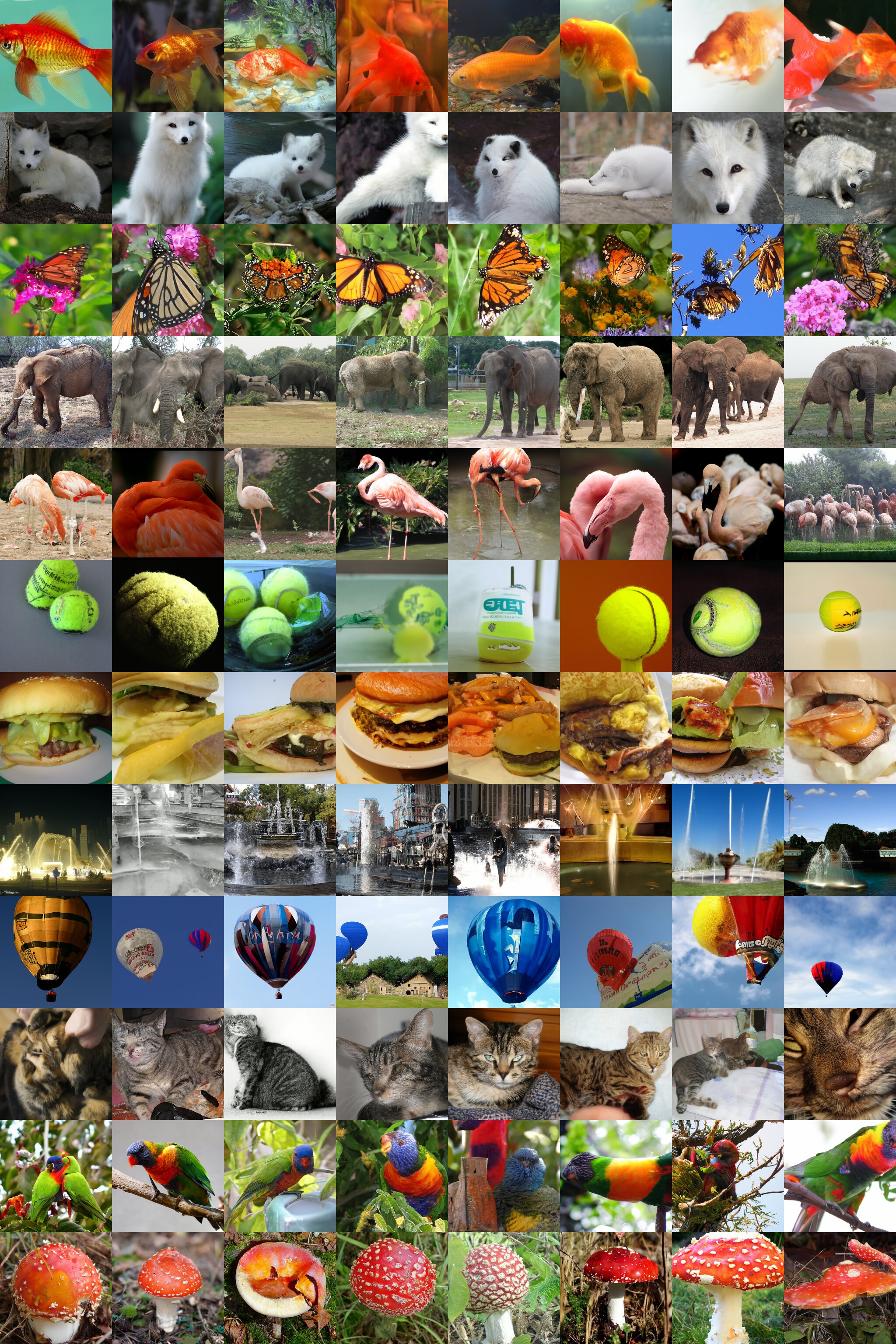}}
    \caption{Samples from our guided 256$\times$256 model using 250 steps with classifier scale 1.0 (FID 4.59). Classes are 1: goldfish, 279: arctic fox, 323: monarch butterfly, 386: african elephant, 130: flamingo, 852: tennis ball, 933: cheeseburger, 562: fountain, 417: balloon, 281: tabby cat, 90: lorikeet, 992: agaric}
\end{figure}
\clearpage
\begin{figure}[h]
    \centerline{\includegraphics[width=\textwidth]{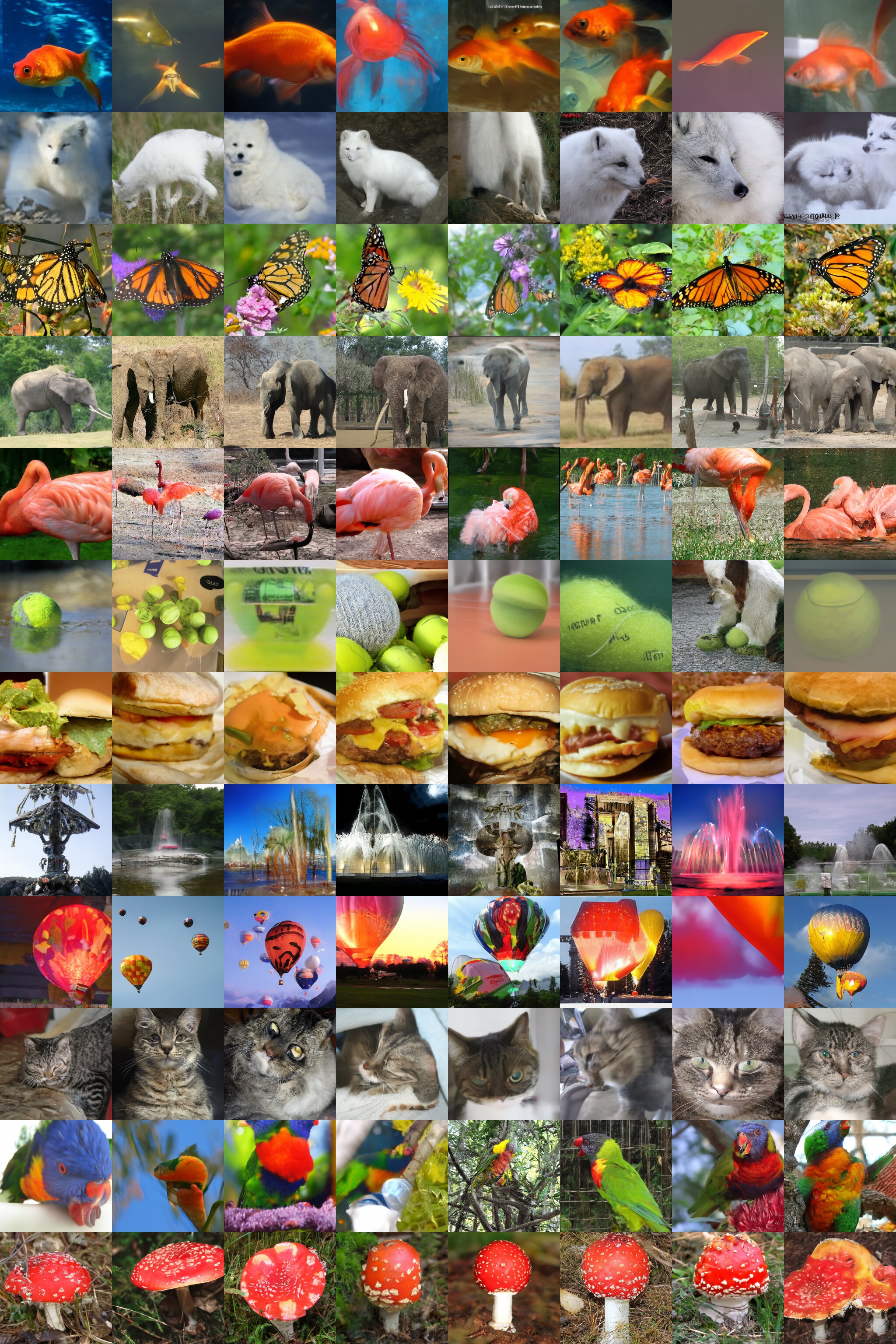}}
    \caption{Samples from our guided 256$\times$256 model using 25 DDIM steps with classifier scale 2.5 (FID 5.44). Classes are 1: goldfish, 279: arctic fox, 323: monarch butterfly, 386: african elephant, 130: flamingo, 852: tennis ball, 933: cheeseburger, 562: fountain, 417: balloon, 281: tabby cat, 90: lorikeet, 992: agaric}
\end{figure}

\begin{figure}[h]
    \centerline{\includegraphics[width=\textwidth]{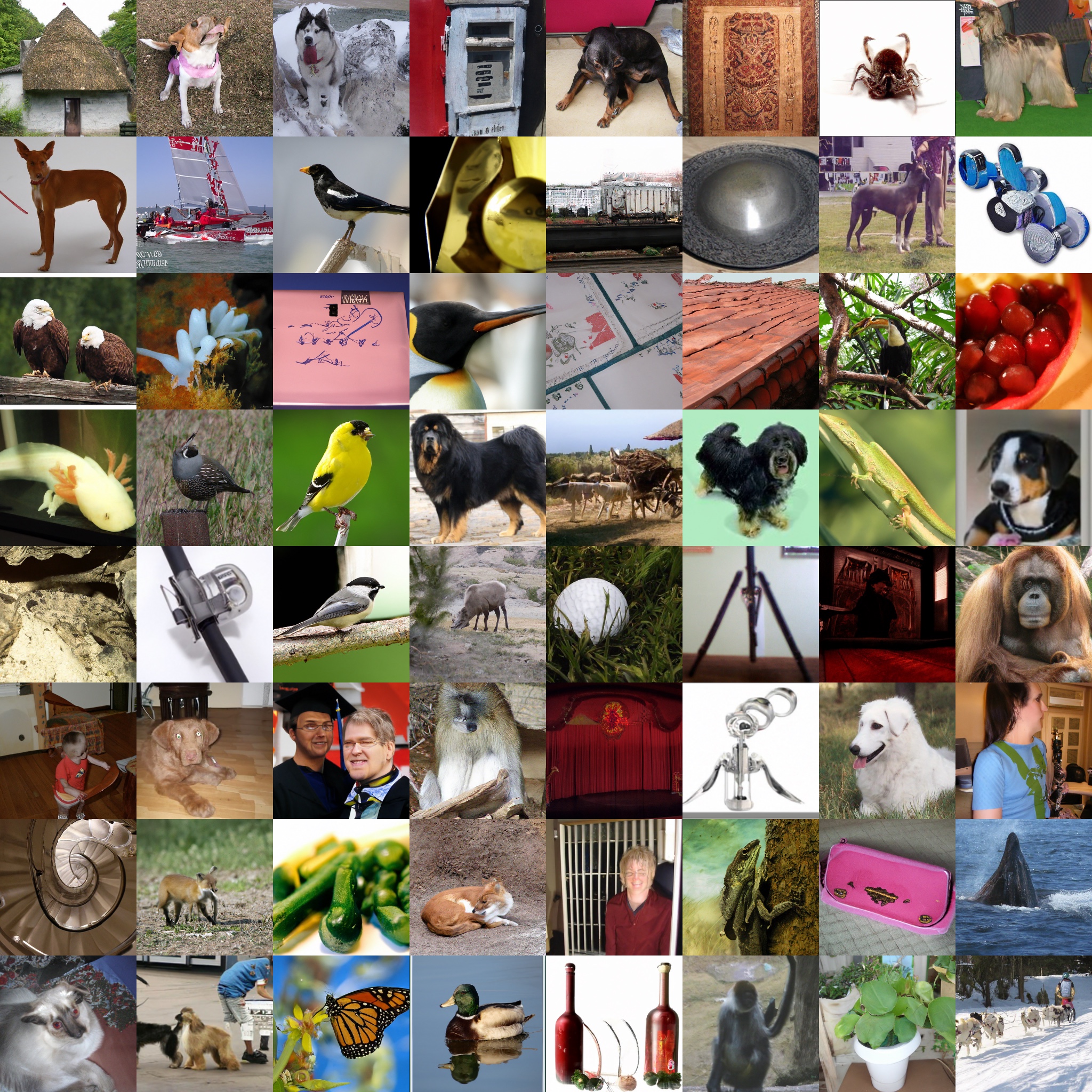}}
    \caption{Random samples from our best 256$\times$256 model (FID 3.94).}
    \vskip -1in
\end{figure}

\begin{figure}[h]
    \centerline{\includegraphics[width=\textwidth]{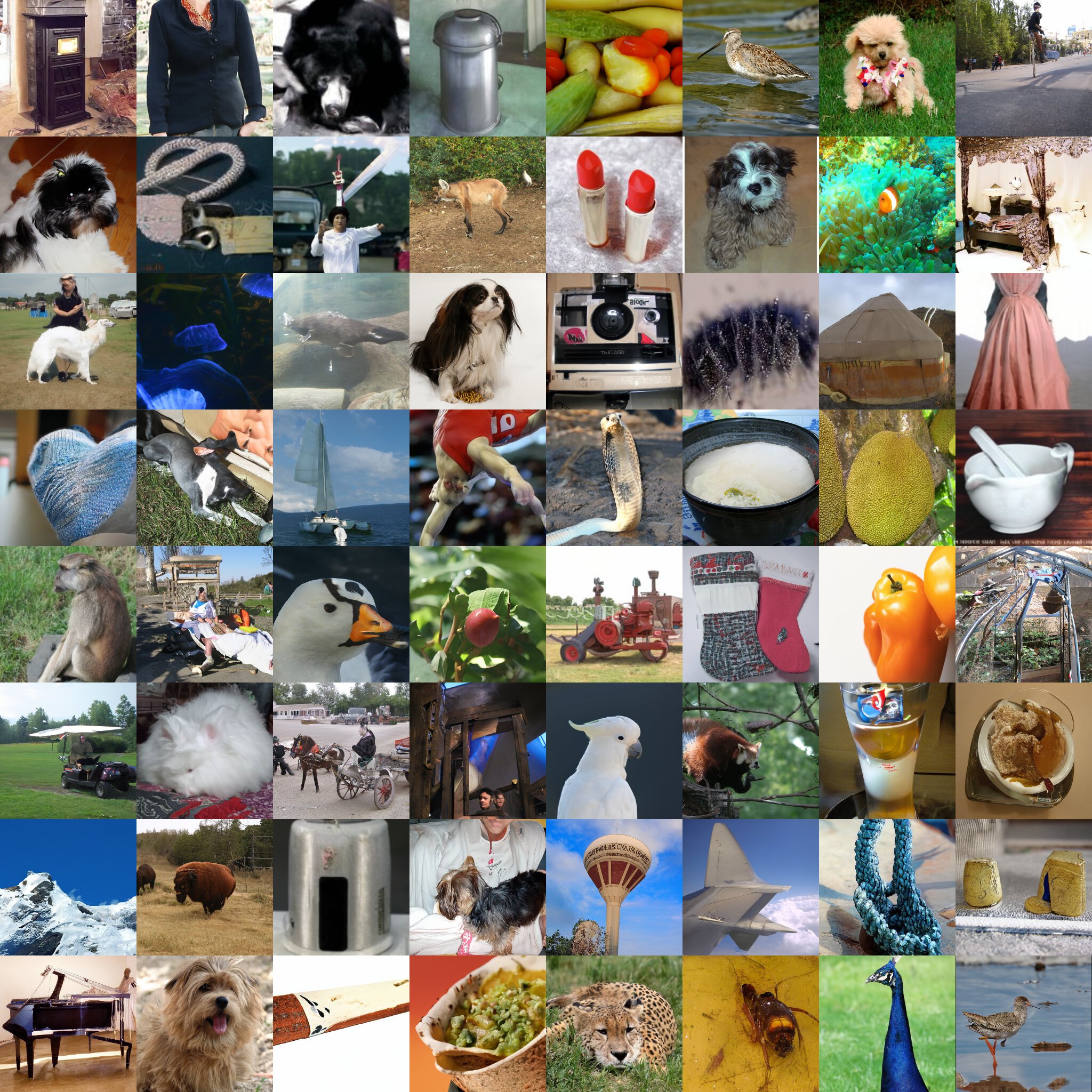}}
    \caption{Random samples from our guided 256$\times$256 model using 250 steps with classifier scale 1.0 (FID 4.59).}
    \vskip -1in
\end{figure}

\clearpage
\section{Samples from LSUN}
\label{app:endfullsamples}
\begin{figure}[h]
    \centerline{\includegraphics[width=\textwidth]{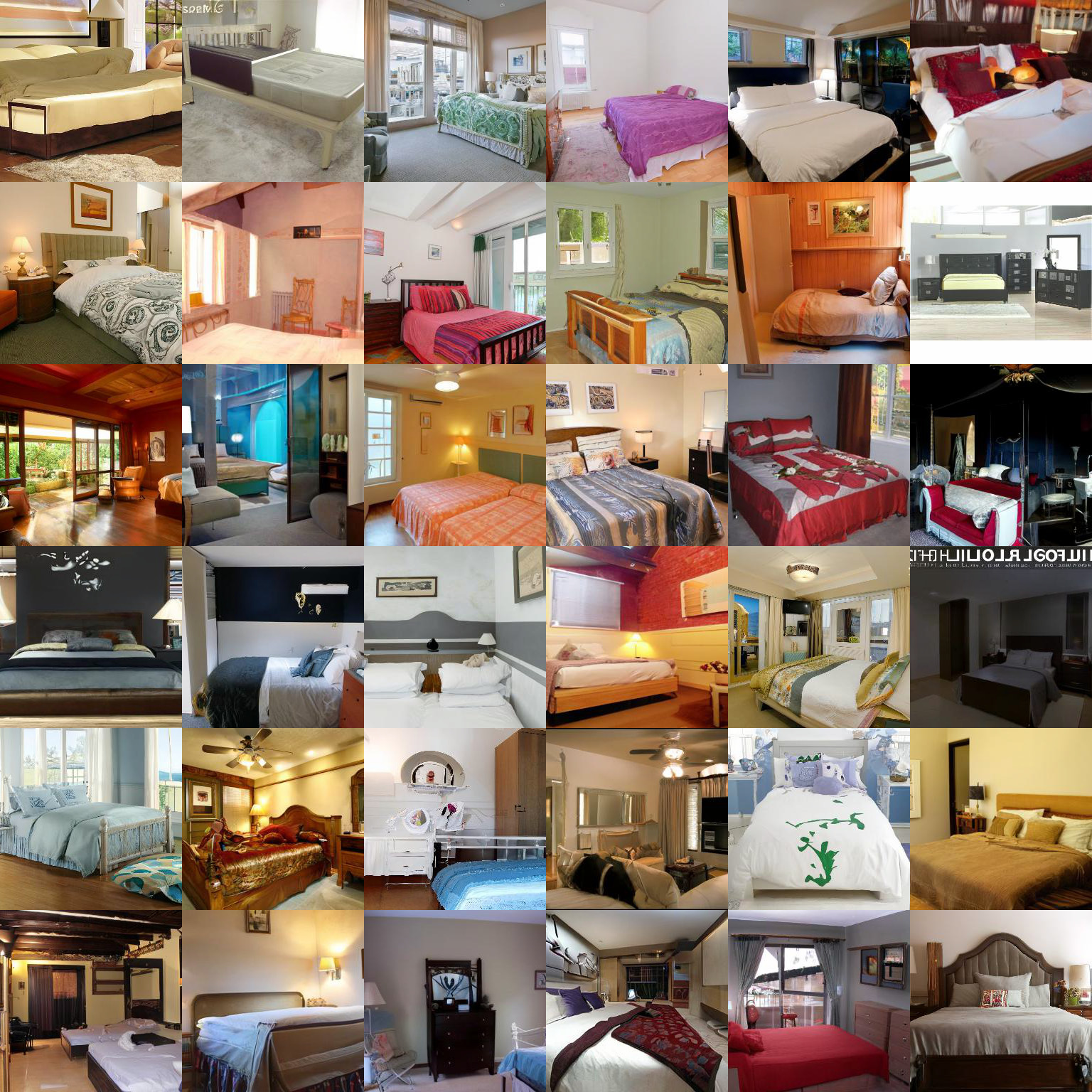}}
    \caption{Random samples from our LSUN bedroom model using 1000 sampling steps. (FID 1.90)}
    \vskip -1in
\end{figure}

\begin{figure}[h]
    \centerline{\includegraphics[width=\textwidth]{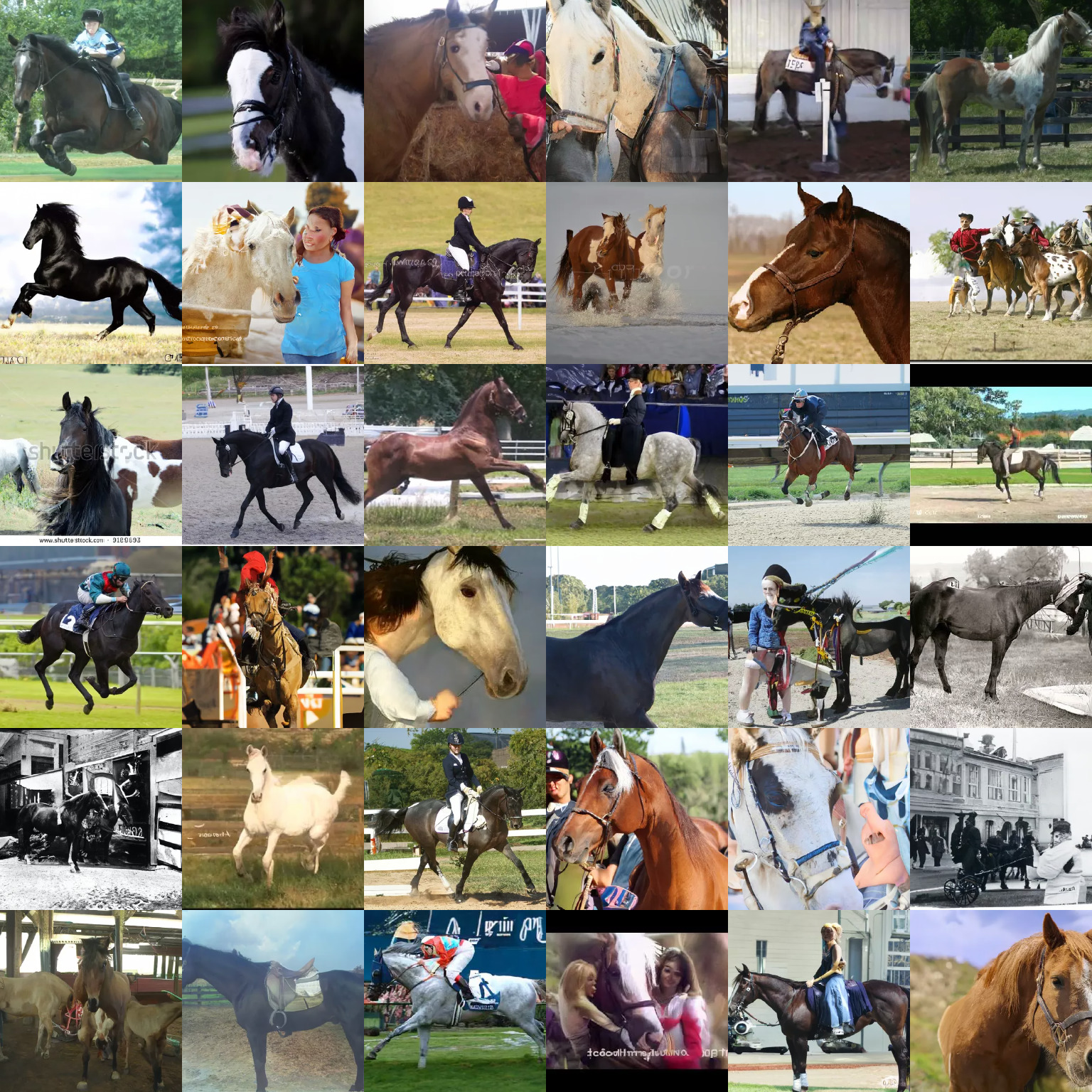}}
    \caption{Random samples from our LSUN horse model using 1000 sampling steps. (FID 2.57)}
    \vskip -1in
\end{figure}

\begin{figure}[h]
    \centerline{\includegraphics[width=\textwidth]{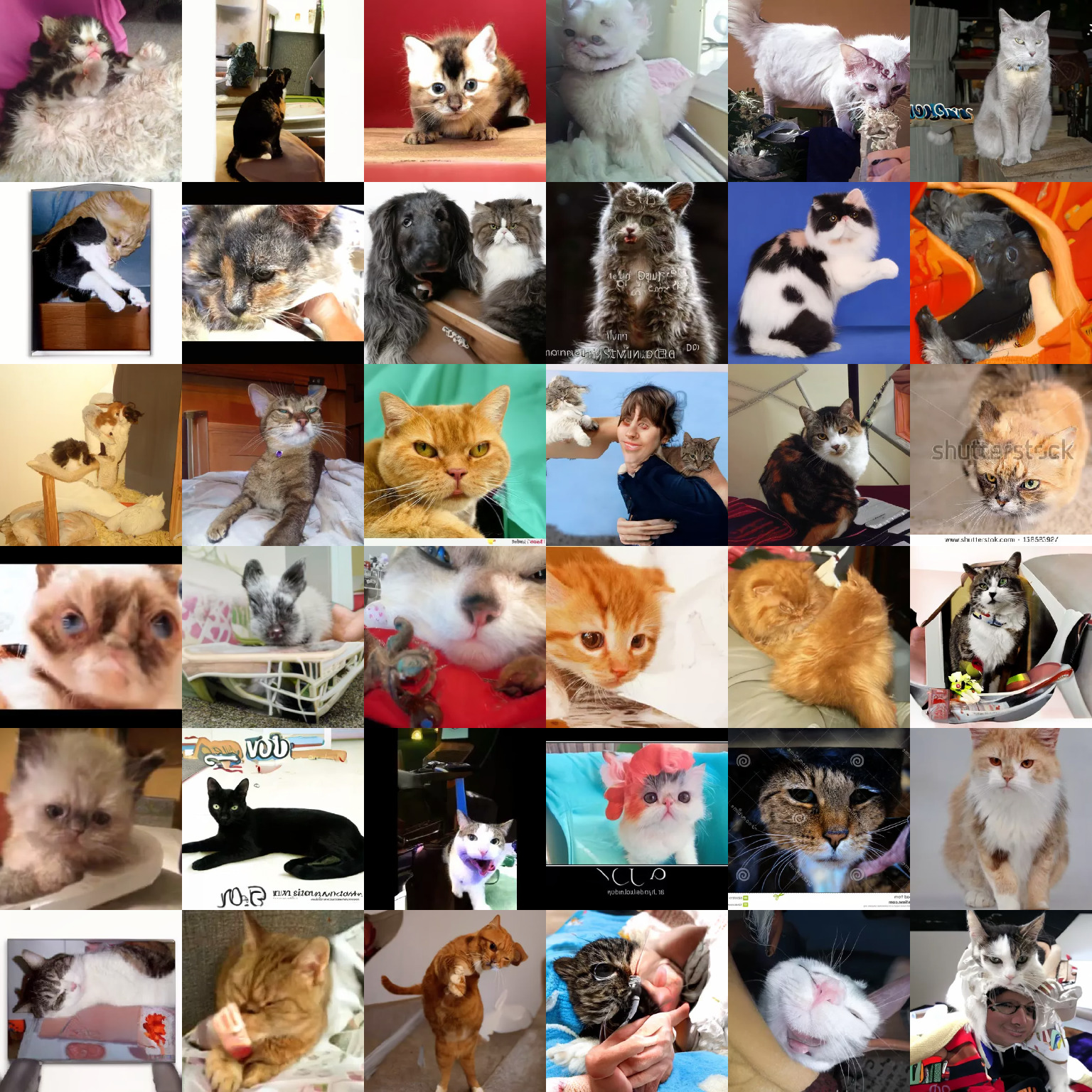}}
    \caption{Random samples from our LSUN cat model using 1000 sampling steps. (FID 5.57)}
    \vskip -1in
\end{figure}

\end{document}